\documentclass[10pt,twocolumn,letterpaper]{article}

\usepackage{iccv}
\usepackage{times}
\usepackage{epsfig}
\usepackage{graphicx}
\usepackage{amsmath}
\usepackage{amssymb}
\usepackage{booktabs}
\usepackage{xcolor}
\usepackage{array}
\newcolumntype{P}[1]{>{\centering\arraybackslash}p{#1}}

\newcommand{\firstone}[1]{\colorbox{red!15}{#1}}
\newcommand{\secondone}[1]{\colorbox{blue!15}{#1}}

\usepackage{bm}
\usepackage{bbm}
\usepackage{multirow}
\usepackage{wrapfig}
\usepackage[accsupp]{axessibility}

\usepackage[pagebackref=true,breaklinks=true,letterpaper=true,colorlinks,bookmarks=false]{hyperref}

\iccvfinalcopy 

\def\iccvPaperID{5315} 

\ificcvfinal\fi

\begin{document}

\title{Multiscale Structure Guided Diffusion for Image Deblurring}
\author{Mengwei Ren$^\dag{^\ddag}$\footnotemark[1]
\quad
Mauricio Delbracio$^\ddag$
\quad
Hossein Talebi$^\ddag$
\quad
Guido Gerig$^\dag$
\quad
Peyman Milanfar$^\ddag$ \\ [0.2em]
$^\dag$New York University \qquad \qquad $^\ddag$Google Research
}

\maketitle
\ificcvfinal\thispagestyle{empty}\fi
\footnotetext[1]{Work done during an internship at Google Research.}
\renewcommand*{\thefootnote}{\arabic{footnote}}
\setcounter{footnote}{0}
\begin{abstract}
Diffusion Probabilistic Models (DPMs) have recently been employed for image deblurring, formulated as an image-conditioned generation process that maps Gaussian noise to the high-quality image, conditioned on the blurry input. Image-conditioned DPMs (icDPMs) have shown more realistic results than regression-based methods when trained on pairwise in-domain data. However, their robustness in restoring images is unclear when presented with out-of-domain images as they do not impose specific degradation models or intermediate constraints. To this end, we introduce a simple yet effective multiscale structure guidance as an implicit bias that informs the icDPM about the coarse structure of the sharp image at the intermediate layers. This guided formulation leads to a significant improvement of the deblurring results, particularly on unseen domain. The guidance is extracted from the latent space of a regression network trained to predict the clean-sharp target at multiple lower resolutions, thus maintaining the most salient sharp structures. With both the blurry input and multiscale guidance, the icDPM model can better understand the blur and recover the clean image. We evaluate a single-dataset trained model on diverse datasets and demonstrate more robust deblurring results with fewer artifacts on unseen data. Our method outperforms existing baselines, achieving state-of-the-art perceptual quality while keeping competitive distortion metrics.
\end{abstract}

\section{Introduction}
\label{sec:intro}

Image deblurring is a fundamentally ill-posed inverse problem that aims to estimate one (or several) high-quality image(s) given a blurry observation. Deep networks allow for end-to-end image deblurring with pairwise supervised learning. 
\begin{figure}[ht]
    \centering
    \includegraphics[width=\linewidth]{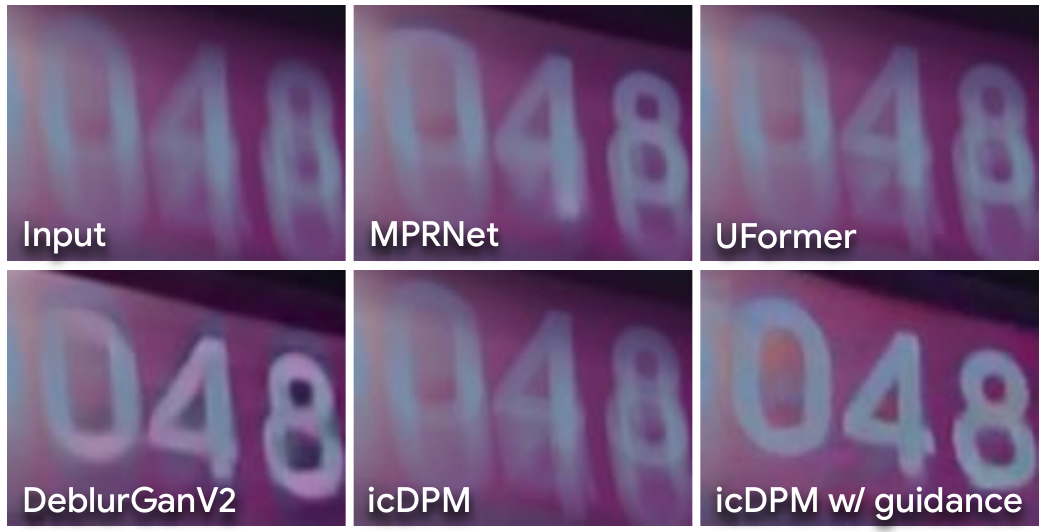}
    \caption{Deblurring example on Realblur-J dataset~\cite{rim2020real} with models \textit{only} trained on synthetic GoPro data~\cite{nah2017deep}, from recent regression-based~\cite{Zamir2021MPRNet,Wang_2022_CVPR_uformer} (MPRNet, UFormer), GAN-based~\cite{Kupyn_2019_ICCV_Deblurganv2} (DeblurGANV2) and image-conditioned diffusion probabilistic methods (icDPM). We introduce a guidance module onto the icDPM formulation, and improves its robustness on unseen image.}
    \label{fig:teaser}
\end{figure}
While deep regression-based methods~\cite{wang2018recovering,zhang2018dynamic,zhou2019spatio,tu2022maxim,Chen_2021_CVPR_HINet,Zamir2021MPRNet,cho2021rethinking_mimounet,whang2022deblurring,tsai2022stripformer,li2022learning,rim2022realistic,nah2022clean} optimize distortion metrics such as PSNR, they often produce over-smoothed outputs that lack visual fidelity~\cite{ledig2017photo,bruna2016super,delbracio2021projected,blau2018perception}.
Therefore, perceptual-driven methods~\cite{li2021perceptual_simplenet,johnson2016perceptual} aim to produce sharp and visually pleasing images that are still faithful to the sharp reference image, typically with a slight compromise on distortion performance, i.e., a less than 3dB drop on PSNR~\cite{blau2018perception,ohayon2021high} allows for significantly better visual quality while still being close to the target image.
GANs~\cite{goodfellow2014generative} are leveraged for improved deblurring perception~\cite{DeblurGAN,Kupyn_2019_ICCV_Deblurganv2}. However, GAN training suffers from instability, mode-collapse and artifacts~\cite{mescheder2018training}, which may hamper the plausibility of the generated images.

Recently, DPMs~\cite{ho2020denoising} further improved the photo-realism in a variety of imaging inverse problems~\cite{saharia2022image,li2022srdiff,whang2022deblurring,saharia2022palette,delbracio2023inversion}, formulated as an image-conditioned generation process, where the DPM takes the degraded estimation as an auxiliary input. Image-conditioned DPMs (icDPMs) do not estimate the degradation kernel nor impose any intermediate constraints. These models are trained using a standard denoising loss~\cite{ho2020denoising} with pairwise training data in a supervised fashion.
In image restoration, such pairwise training dataset is typically artificially curated by applying known degradation models on a group of clean images, which inevitably introduces a domain gap between the synthetic training dataset and real-world blurry images. 
When presented with unseen data, the robustness of icDPMs are rather unclear as the intermediate restoration process is intractable. E.g., we observe a noticeable performance drop when we apply the synthetically trained icDPM to out-of-domain data, including failure to deblur the input (Fig.~\ref{fig:teaser}) and injection of artifacts (Fig.~\ref{fig:regression} `icDPM' and Fig.~\ref{fig:realblur_results} `DvSR’).
We empirically established a connection between domain sensitivity and image-conditioning in the existing deblurring icDPMs~\cite{saharia2022palette, saharia2022image, whang2022deblurring}, where the observed poor generalization is attributed to the naive input-level concatenation and the lack of intermediate constraints during the deblurring process. When optimized on the synthetic training set, overfitting or memorization ~\cite{somepalli2022diffusion} may occur, making the model vulnerable to shift of the input distribution.
Currently, conditioning DPM on blurred or corrupted images is under-explored~\cite{rombach2022high}, and we hypothesize that more effective image conditioning for icDPM is crucial to make the model more constrained and robust towards unseen domain. 

Inspired by traditional blind deblurring algorithms where optimization is made using explicit structural priors (e.g., containing image saliency~\cite{pan2014deblurring,xu2013unnatural}), we enhance the icDPM backbone (UNet~\cite{ronneberger2015u}) with a multiscale structure guidance at intermediate layers. 
These guidance features are obtained through a regression network trained to predict salient sharp features from the input. 
The guidance, in conjunction with the blurry image,  provide more informative cues to the model regarding the specific degradation in the image. As a result, the model can more accurately recover the clean image and generalize more effectively. 
Our contributions are threefold: (1) we investigate and analyze the domain generalization of conditional diffusion models in motion deblurring task, and empirically find a relationship between model robustness and image-conditioning;
(2) we propose an intuitive but effective guidance module that projects the input image to a multiscale structure representation, which is then incorporated as an auxiliary prior to make the diffusion model more robust; (3) Compared with existing benchmarks, our single-dataset trained model shows more robust results across different test sets by producing more plausible deblurring and fewer artifacts, quantified by the state-of-the-art perceptual quality and on par distortion metrics. 

\section{Related Works}
\noindent \textbf{Single image deblurring} is the inverse process of recovering one or multiple high-quality, sharp images from the blurry observation.
Typically, classic deblurring approaches involve variational optimization~\cite{fergus2006removing,krishnan2011blind,levin2011efficient,michaeli2014blind,pan2016blind,xu2013unnatural,anger2019efficient,jin2018normalized}, with prior assumptions on blur kernels, images or both, to alleviate the ill-posedness of the inverse problem. Handcrafted structural priors, such as edges and shapes, have been used successfully in many algorithms to guide the deblurring process towards preserving important features in the image while removing blur~\cite{pan2014deblurring,pan2016blind,xu2013unnatural}. Our design principle is inspired by these approaches and involves a learned guidance as implicit structural bias.
With the emergent of deep learning, deblurring can be cast as a particular image-to-image translation problem where a deep model takes the blurry image as its input, and predicts a high quality counterpart, supervised by pixel wise losses between the recovered image and the target~\cite{wang2018recovering,zhang2018dynamic,zhou2019spatio,tu2022maxim,Chen_2021_CVPR_HINet,Zamir2021MPRNet,cho2021rethinking_mimounet,whang2022deblurring,tsai2022stripformer,li2022learning,rim2022realistic,nah2022clean,ji2022xydeblur}.
Pixel-wise losses, such as $L_1$ and $L_2$, are known to result in over-smoothed images~\cite{ledig2017photo,bruna2016super,delbracio2021projected} given their `regression to the mean' nature. To this end, perception-driven losses including perceptual~\cite{johnson2016perceptual,zhang2018unreasonable, mechrez2018contextual,mechrez2018maintaining,zhang2019zoom,delbracio2021projected} and adversarial losses~\cite{DeblurGAN,Kupyn_2019_ICCV_Deblurganv2} are added on top of the pixel-wise constraints, to improve the visual fidelity of the deblurred image, with a compromising drop in distortion scores~\cite{blau2018perception,ohayon2021high}. Tangentially, recent works seek to improve the architectural design by exploring attention mechanisms~\cite{niu2020single,zamir2020learning,Wang_2022_CVPR_uformer,Zamir2021Restormer,tsai2022stripformer,tu2022maxim}, multi-scale paradigms~\cite{nah2017deep,cho2021rethinking_mimounet} and multi-stage frameworks~\cite{zhang2019deep,Chen_2021_CVPR_HINet,Zamir2021MPRNet}.

\noindent \textbf{Diffusion Probablistic Models (DPM)}~\cite{sohl_thermodynamics,ho2020denoising,ddim,dhariwal2021diffusion}, Score-based models~\cite{ncsn, ncsnv2, ncsnv3} and their recent exploratory generalizations~\cite{bansal2022cold,hoogeboom2022blurring,daras2022soft} achieved remarkable results in a varied range of applications\cite{diffusion_survey}, from image and video synthesis\cite{saharia2022photorealistic,ramesh2022hierarchical,latent_diffusion,ho2021classifier,kawar2022enhancing,ho2022video}, to solving general imaging inverse problems\cite{daras_dagan_2022score,kadkhodaie2021stochastic,kawar2021snips,kawar2022ddrm,laumont2022bayesian,chung2022improving}. DPMs are characterized for having stable training~\cite{ho2020denoising,dhariwal2021diffusion,karras2022elucidating}, diverse mode coverage~\cite{song2021maximum,kingma2021variational}, and high perception~\cite{saharia2022photorealistic,dhariwal2021diffusion,ramesh2022hierarchical}.
 DPM formulation involves a fixed forward process of gradually adding Gaussian noise to the image, and a learnable reverse process to denoise and recover the clean image, operated with a Markov chain structure. 
Conditional DPMs aim to perform image synthesis with an additional input (class~\cite{dhariwal2021diffusion}, text~\cite{saharia2022photorealistic,ramesh2022hierarchical}, source image).

\noindent \textbf{Image-conditioned DPMs (icDPMs)} have been successfully re-purposed for image restoration tasks such as super-resolution~\cite{saharia2022image,li2022srdiff}, deblurring~\cite{whang2022deblurring}, JPEG restoration~\cite{saharia2022palette,kawar2022jpeg}. This is achieved by concatenating the corrupted observation at input level. They do not require task-specific losses or architectural designs, and have been adopted due to high sample perceptual quality. ControlNet~\cite{zhang2023adding} further enables the task-specific image conditioning for pretrained text-to-image diffusion models. InDI~\cite{delbracio2023inversion} presents an alternative, and more intuitive, diffusion process where the low-quality input is directly restored into a high-quality image in small steps.
Nevertheless, generalization of DPMs to unseen shifts in domain, and their low-quality/corrupted image conditioning remains unexplored.

\noindent \textbf{Generalization to unseen domain}
As mentioned above, deep restoration models for deblurring rely on synthetic pairwise training data. 
However, any well-trained deep restoration model may fail to produce comparable results on out-of-domain data (Fig.~\ref{fig:teaser}). 
To address this issue, researchers have pursued two main directions for improving model generalization: enhancing the representativeness and realism of the training data, or improving the model's domain generalization ability.
Our method focuses on the latter, but it is not mutually exclusive with the former direction and can be combined to further improve the results.
To tackle the data limitations, previous works focus on acquiring or combining more representative training data~\cite{Nah_2019_CVPR_Workshops_REDS,rim2020real,rim2022realistic,zhou2022lednet}, and/or generate realistic degraded images using generative approaches~\cite{zhang2020deblurring,wolf2021deflow}.
Other prior works focus on explicit domain adaptation, leveraging transfer learning techniques to reduce domain gaps. These approaches include unpaired image translation~\cite{hoffman2018cycada,ren2021segmentation} and domain adaptation~\cite{wang_unsupervised_sr,shao2020domain,lu2019unsupervised,nah2022clean}, which typically involve an adversarial formulation and joint training between two specified domains. However, these methods may require retraining when a new dataset is introduced. 
In contrast, our method does not involve explicit adaptation between specified domains. Instead, we focus on introducing more effective image conditioning mechanism that naturally makes the model more robust towards distribution shift. 

\begin{figure}[!t]
    \centering
    \includegraphics[width=0.98\linewidth]{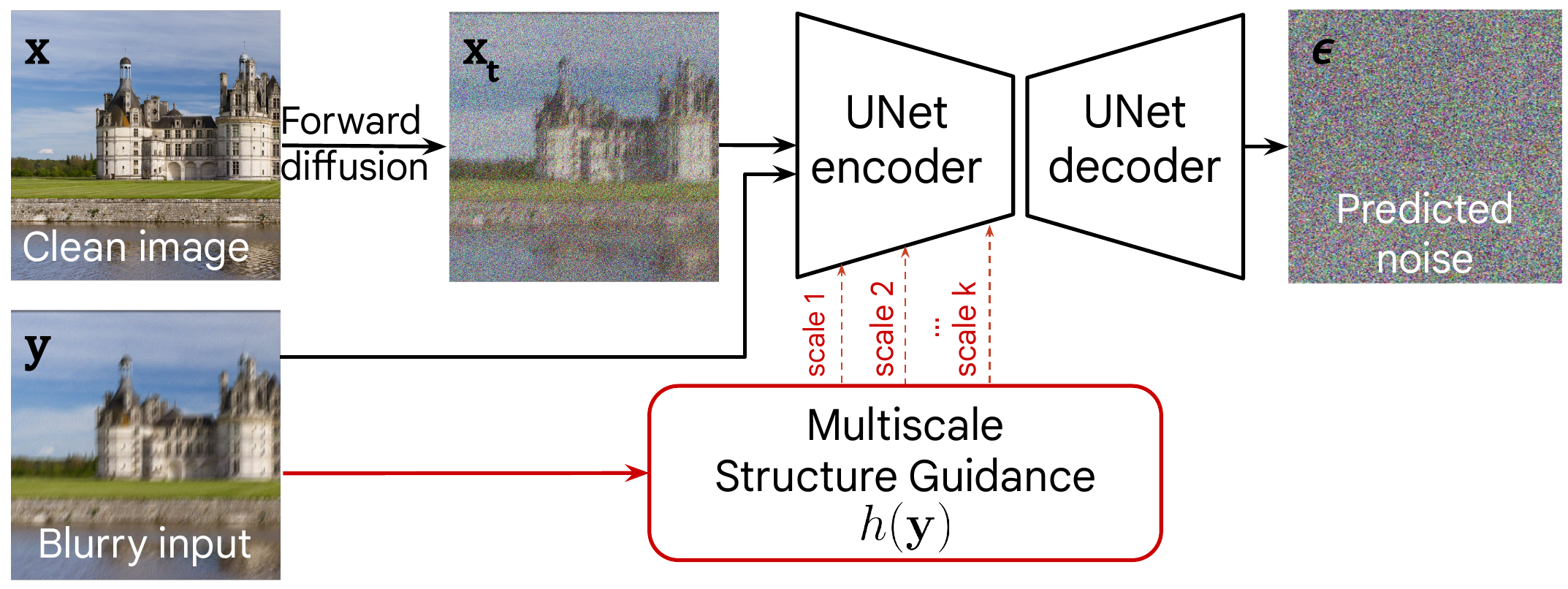}
    \caption{The \textit{training} process of the proposed deblurring method. The backbone model is a standard image-conditioned DPM (icDPM) where a UNet learns to perform denoising towards the clean image conditioned on the blurry input. We equip the icDPM with a structure guidance module (detailed in Fig.~\ref{fig:representation}) to better inform the model of the coarse sharp structure at multiple scales.
    }
    \label{fig:pipeline}
\end{figure}
\section{Method}
\subsection{Overview}
We assume access to a paired dataset with samples $(\bm{x},\bm{y}) \sim p_\text{train}(\bm{x},\bm{y})$, where $\bm{x}$ represents the high-quality sharp image, and $\bm{y}$ is the respective low-quality blurry observation (denoted in Fig.~\ref{fig:pipeline}). Such paired dataset is typically generated by simulating degraded images from the high-quality ones adopting a specific degradation model. 

The goal is to reconstruct one or multiple clean, sharp images $\bm{x}$ from the low-quality observation $\hat{\bm{y}} \sim p_\text{real}(\hat{\bm{y}})$. 
Generally, the distribution of the training set $p_\text{train}$ differs from that of unseen images $p_\text{real}$. It is thus crucial that a model not only performs well on $p_\text{train}$ but also generalizes to $p_\text{real}$.  

\noindent \textbf{DPMs} We consider a general-purpose DPM for our formulation given its superior performance in high-quality image restoration~\cite{saharia2022image,whang2022deblurring}. 
In what follows, we briefly describe the training and sampling of a DPM to contextualize our work. 
Unconditional DPMs aim to sample from the data distribution $p(\bm{x})$ by iteratively denoising samples from a Gaussian distribution and converting them into samples from the target data distribution. To train such model, a forward diffusion process and a reverse process are involved. As illustrated in Fig.~\ref{fig:pipeline}, at a diffusion step $t$, a noisy version $\bm{x}_t$ of the target image $\bm{x}$ is generated by $\bm{x}_t = \sqrt{\alpha_t} \bm{x} + \sqrt{(1- \alpha_t)} \bm{\epsilon}$, $\bm{\epsilon} \sim \mathcal{N}(0,\bm{I_d})$,  where $\bm{\epsilon}$ is sampled from a standard Gaussian distribution $\mathcal{N}(0,\bm{I}_d)$, and $ \alpha_t$ controls the amount of noise added at each step $t$. In the reverse process, an image-to-image network (i.e. UNet) $\mathcal{G}_\theta (\bm{x}_t, t)$ parameterized by $\theta$ learns to estimate the clean image from the partially noisy input $\bm{x}_t$. In practice, a reparameterization of the model to predict the noise instead of the clean image leads to better sample quality~\cite{ho2020denoising}. Once trained, it samples a clean image by iteratively running for $T$ steps starting from a pure Gaussian noise $\bm{x}_T \sim \mathcal{N}(0,\bm{I_d})$.
\noindent \textbf{Image-conditioned DPMs} further inject an input image $\bm{y}$ so as to generate high-quality samples that are paired with the low-quality observation. This involves generating samples from the conditional distribution of $p(\bm{x}|\bm{y})$ (posterior). 
A conditional DPM $\mathcal{G}_\theta ([\bm{x}_t, \bm{y}], t)$ is used, where the image conditioning is typically implemented via concatenation of $\bm{y}$ and $\bm{x}_t$ at input-level~\cite{saharia2022image,whang2022deblurring,saharia2022palette}.
However, we found that this formulation is sensitive to domain shift in input images, and leads to poor generalization (`DPM' in Fig.~\ref{fig:teaser}). Moreover, in many cases it introduces visual artifacts (`DvSR' in Fig.~\ref{fig:realblur_results}). 
We speculate that this is due to the naive image-conditioning (input-level concatenation), which lacks constraints in the intermediate process. 
Therefore, we integrate a multiscale structure guidance $h(\bm{y})$ into the latent space of the icDPM backbone, to inform the model about salient image features, such as significant coarse structures that are essential for reconstructing a high-quality image, while disentangling irrelevant information, such as the footprint of blur kernels and color information. 
To obtain such guidance with the aforementioned characteristics, we propose an auxiliary regression network and leverage its learned features as the realization of the guidance, described below in  Sec.~\ref{sec:domain_invariant_representation}.

\subsection{Multiscale structure guidance}
\label{sec:domain_invariant_representation}
\begin{figure}[!t]
    \centering
    \includegraphics[width=\linewidth]{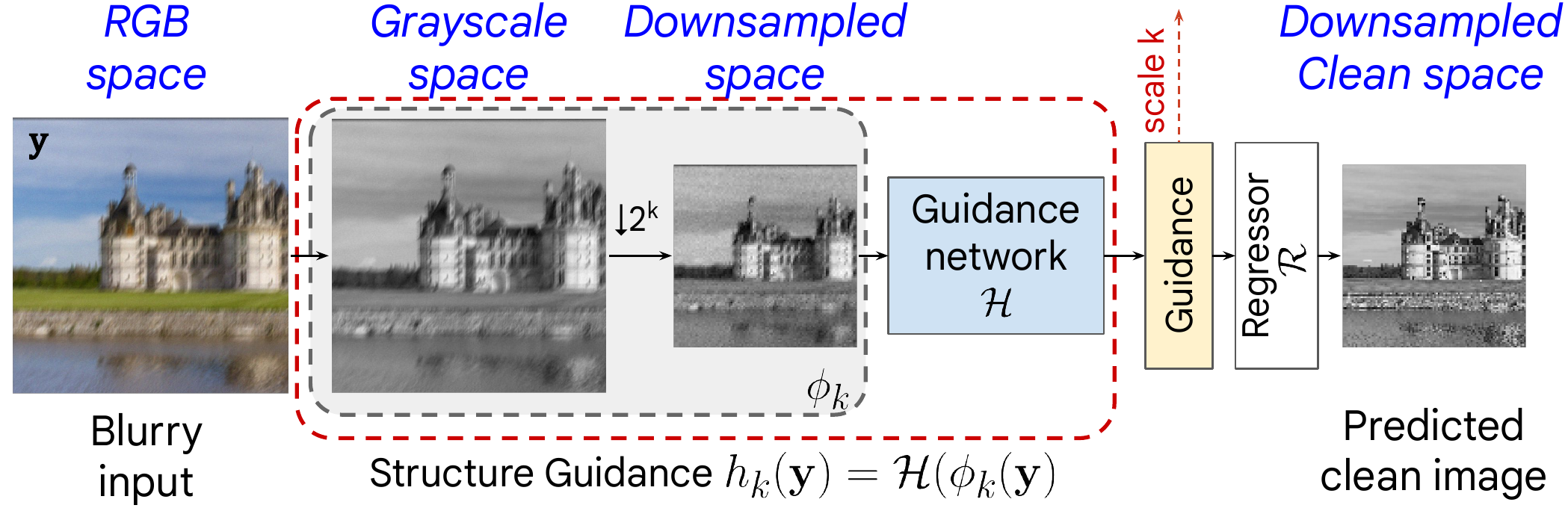}
    \caption{The learned structure guidance (red box in Fig.~\ref{fig:pipeline}) is extracted from the latent features of a regression network trained to predict the luma channel of the sharp target at multiple lower resolutions. In this way, the guidance maintains only structure-relevant information, representing the underlying sharp image.
    }
    \label{fig:representation}
\end{figure}

Fig.~\ref{fig:representation} shows the details of our proposed guidance, denoted as $h(\cdot)$. 
The DPM equipped with such multiscale guidance is better aware of the underlying salient structures of the input, thus it learns to better sample from the target conditional distribution to $p_\text{real}(\bm{x}|\bm{y})$. Moreover, the distribution of $h(\bm{y})$ does not change significantly when the input domain changes, so that it can reliably provide the auxiliary structure guidance even when applied to unseen domain. 
To both ends, we construct the guidance module as $h_k(\cdot)=\mathcal{H}(\phi_k(\cdot))$. At scale $k$, it consists of an image transformation function $\phi_k(\cdot)$,
followed by a regression-driven guidance network $\mathcal{H}$.
Specifically, $\phi_k(\cdot)$ transforms the input image $\bm{y}$ to suppress information not relevant to the coarse sharp image structures (e.g., color, and information about the domain-specific degradation). This ensures that $\mathcal{H}$ operates on a less input-domain-sensitive space.
We first convert $\bm{y}$ to the grayscale space $\bar{\bm{y}}$, and then, $\bar{\bm{y}}$ gets downsampled by a factor of $2^k$, where $k=1,2,3$. This removes fine details (including the footprint of blur to certain amount), while preserving coarse structures at multiple lower resolutions.
Motivated by~\cite{wolf2021deflow}, we also add a small amount of Gaussian noise to mask other domain specific degradations/characteristics, and make the output less sensitive to input domain shift. In recent diffusion models, the addition of noise becomes a common practice~\cite{ruiz2022dreambooth,saharia2022photorealistic} as it enhances robustness when dealing with out-of-domain data.
Thus, 

\begin{align}
\phi_k(\bm{y}) = d_{\downarrow k}(\bm{\bar y}) + \bm{n}, \quad \bm{n} \sim \mathcal{N}(0, \sigma^2 \bm{I}).
\end{align}
Then, the guidance network $\mathcal{H}_\varphi$ extracts the guidance feature by mapping $\phi_k(\bm{y})$ onto the representation/latent space as $h_k(\bm{y})=\mathcal{H}_\varphi(\phi_k(\bm{y}))$. To make sure it obtains salient structure features and further filters out insignificant information, we apply a regression task $\mathcal{R}_\varphi$ on top of $h_k(\bm{y})$, and constraints the output to be closer to its sharp target $\phi_k(\bm{x})$. In this way, the guidance $h_k(\bm{y})$ at scale $k$ is enforced to maintain information that is relevant to a sharp image, and suppress other signals that are input-specific (e.g., trace of blurs).

Finally, we incorporate the multiscale guidance $\{h_k(\bm{y})\}$ to the original diffusion UNet by adding the extracted representation to the feature map at the respective scale on the diffusion encoder (Fig~\ref{fig:pipeline}) as an extra bias. To compensate for the difference in depth, at each corresponding scale, we apply a convolutional layer that has the same number of features as in the diffusion encoder. Detailed diagram is provided in the appendix.

\subsection{Training loss}
\label{sec:objectives}
Our model is trained end-to-end with both a multiscale regression loss for optimizing the guidance network, and a denoising loss in icDPM. 
The regression loss is the mean squared error at each scale $k$ defined as:
\begin{align}
\mathcal{L}_\text{guidance}^k = \mathbb{E}_{({\bm{x}},{\bm{y}}) \sim p_\text{train}}\|\mathcal{R}_{\varphi}({\mathcal{H}}_{\varphi}(\phi_k({\bm{y}}))) - \phi_k({\bm{x}}) \|_2, 
\end{align}
where $\mathcal{H}_\varphi$ is the guidance feature extractor, and $\mathcal{R}$ is instantiated as a single convolutional layer that projects the guidance feature to the final output towards the clean image (as depicted in Fig.~\ref{fig:representation}). The total regression loss is the average over different scales $\mathcal{L}_\text{guidance}=\sum_k \mathcal{L}_\text{guidance}^k$.
Note that we do not use any additional downsampling/upsampling operation in the guidance network, so the spatial dimension remains the same at each scale. We empirically observe that the best performance is obtained by integrating three different scales with $k=1,2,3$ with details discussed in Sec.~\ref{sec:ablation}.

By aggregating the information from the input image $\bm{y}$, and the multi-scale guidance $\{h_k(\bm{y})\}$, our icDPM $\mathcal{G}$ is trained by minimizing the denoising loss,
\begin{align}
    \mathcal{L}_\text{DPM} &= \mathbb{E}_{({\bm{x}},{\bm{y}}) \sim p_\text{train}}\mathbb{E}_{t \sim \text{Unif}(0,1)} \mathbb{E}_{\bm{\epsilon} \sim \mathcal{N}(0, \bm{I})} \\
   &\| \mathcal{G}_{\theta}(\bm{x}_t,\bm{y}, \{\mathcal{H}_{\varphi}(\phi_k(\bm{y}))\}, \alpha_t) - \bm{\epsilon} \|_1. \nonumber
\end{align}

The denoising model parameterized by $\theta$ predicts the noise $\bm{\epsilon}$, given the noisy corruption ${{\bm{x}}}_t$, the blurry input $\bm{y}$, the noise scheduler $\alpha_t$ as well as the proposed multiscale guidance $\{\mathcal{H}_{\varphi}(\phi_k(\bm{y}))\}$. 
The total training loss $\mathcal{L} = \mathcal{L}_\text{guidance} + \mathcal{L}_{DPM}$, which is used to optimize the guidance network $\mathcal{H}$, the regression layer ${\mathcal{R}}$, and icDPM $\mathcal{G}$ in an end-to-end manner. During the inference, the model starts with a Gaussian noise, and iteratively recovers the clean image, conditioned on both the blurry input and the multiscale guidance at each denoising step.
\section{Experiments}
\subsection{Setup and metrics}
As motivated above, we are particularly interested in the model generalization of DPMs to unseen blurry data. Therefore, we set up our experiments under the scenario that the model will be only trained with synthetic paired dataset, and will be evaluated on a few unseen testing sets where the images may present different content and distortions than the in-domain data. To benchmark, we use the widely adopted motion deblurring dataset {GoPro~\cite{nah2017deep}} as our training data, and assume {Realblur-J}~\cite{rim2020real}, {REDS}~\cite{Nah_2019_CVPR_Workshops_REDS} and {HIDE}~\cite{shen2019human} are representatives of unseen test
sets. 

In \textbf{GoPro}~\cite{shen2019human}, 3214 pairs of blurry/clean training examples are provided for training, and 1111 images are held-out for evaluation. 
\textbf{Realblur-J}~\cite{rim2020real} is a recent realistic dataset mainly consisting of low-light scenes with motion blur with 980 test images provided. We consider it to present the largest domain gap with GoPro. 
\textbf{REDS}~\cite{Nah_2019_CVPR_Workshops_REDS} presents a complimentary video deblurring dataset with more realistic motion blur. We follow~\cite{Nah_2019_CVPR_Workshops_REDS,Chen_2021_CVPR_HINet} and extract 300 validation images for the motion deblur test.
\textbf{HIDE}~\cite{shen2019human} is the most commonly adopted dataset to test the model generalization ability trained from GoPro with 2025 test images. 
\subsection{Implementation details}
Our framework is implemented in TensorFlow 2.0 and trained on 32 TPU v3 cores. We warm start the training with only regression loss, and linearly increase the weight of the denoising loss to 1 within the first 60k iterations. 
Adam optimizer~\cite{kingma2014adam} is used during the training ($\beta_1= 0.5, \beta_2=0.999$), with batch size 256 on $128\times 128$ random crops. We use linear increasing learning rate within the first $20$k iterations, then with a constant learning rate $1\times 10^{-4}$.
We use a fully-convolutional UNet architecture~\cite{whang2022deblurring} for icDPM to ensure the model can be used at arbitrary image resolutions. During the inference,
we follow~\cite{whang2022deblurring} and perform a sequence of sampling under different parameters. More details are included in the appendix.

\subsection{Effectiveness of the guidance}
We first validate the effectiveness of the proposed guidance module by qualitatively comparing with our baseline setup, which is a standard image-conditioned DPM (abbrev as `icDPM'), on top of which we will introduce the guidance module (abbrev as `icDPM w/ Guide').
\begin{table}[h]
\footnotesize
\centering
\caption{Inception distances analysis (domain-shift) between two different domains (GoPro v.s. Realblur-J) at different scales within different image space. At each scale, the guidance network output consistently reduces the gap compared to the downsampled input images, as expected. The distance between grayscale inputs (at original spatial resolution) are also given as a reference.
KID values are scaled by a factor of 100 for readability.
}
\label{table:inception_distance}
\begin{tabular}{lccccc}
\toprule

Space                   & FID $\downarrow$ & KID $\downarrow$  \\
\midrule
Input         &  61.115         &  3.07                       \\
\midrule
Input $\times 2$ downsampled
                        &  58.266         &  3.02                 \\
 Guidance $\times 2$ output
                        &  49.437         &  3.00              \\
\midrule
Input $\times 4$ downsampled
                         &  56.313         &  3.60                 \\
Guidance $\times 4$ output         
                        &  47.984         &  3.46                 \\
\midrule
Input $\times 8$ downsampled
                         &  49.684         &  4.91                 \\
Guidance $\times 8$ output        
                        &  44.649         &  4.70                 \\
\bottomrule
\vspace{-1em}
\end{tabular}
\end{table}
\begin{figure}[t]
    \centering
    \includegraphics[width=0.75\linewidth]{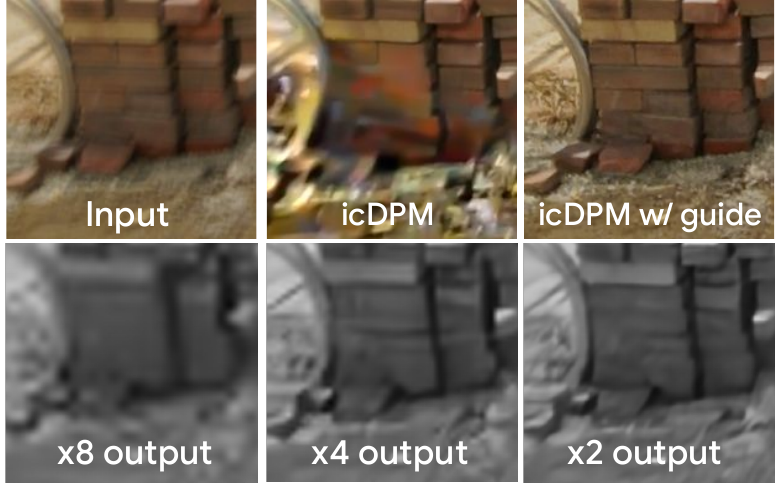}
    \caption{Top row: We compare the visual deblurring results on an out-of-domain image between a standard icDPM and icDPM with the proposed guidance module. While icDPM is prone to producing artifacts, our method that incorporates the guidance is more robust.
    Bottom row: We also visualize our multiscale regression outputs from scales of $\times 8, \times 4, \times 2$, indicating the prediction at spatial resolutions of $1/8, 1/4, 1/2$ the input image.}
    \label{fig:regression}
\end{figure}

\noindent \textbf{Guidance and domain gap.}
As the guidance is designed to improve the robustness towards domain shift, we perform an analysis on the Inception distances from different intermediate `image space' to verify whether the guidance module is progressively reducing the gap between inputs from different sources (i.e. different blurry images in our scenario). In Table~\ref{table:inception_distance}, we start by calculating the per-scale Inception distance between GoPro (in-domain) and Realblur-J (out-of-domain) images. 
At each scale of $\times2$, $\times4$ and $\times8$ downsampled space, we observe a consistent reduction of FID and KID on the guidance network outputs, compared with the downsampled grayscale inputs. This demonstrates that the introduction of the learned guidance may provide more domain-agnostic information and benefit generalization of the model on unseen domains. 
In Fig.~\ref{fig:regression}, we display the multiscale regression outputs at different scales on an out-of-domain input. The results align with our expectations, as the grayscale prediction at each scale progressively approached a clean image. Further, we noticed pronounced sampling artifacts from icDPM in this example, which are effectively eliminated with the proposed guidance.\\
\noindent\textbf{Guidance and model capacity.}
\begin{table*}[t]
\footnotesize
\centering
\caption{The effectiveness of the proposed guidance on top of image-conditioned DPM (icDPM), under different network size (`-S' and `-L' refer to small and large networks respectively). We show both In-Domain (train on GoPro, test on GoPro), and Out-of-Domain (train on GoPro, test on Realblur-J) results. Based on (a)-(b), we observe a larger icDPM boost in-domain performance, while not necessarily lead to better out-of-domain results. With guidance (c)-(d), we observe consistent improvements both in-domain and out-of-domain.}
\label{table:ablation_network_size}
\begin{tabular}{lccccccc}
\toprule
\multirow{2}{*}{} & \multirow{2}{*}{Guidance} & \multirow{2}{*}{Diffusion} & \multirow{2}{*}{\#Params} & \multicolumn{2}{c}{\textbf{In-Domain}}& 
\multicolumn{2}{c}{\textbf{Out-of-domain}}              \\ \cmidrule(lr){5-6}\cmidrule(lr){7-8}
                  &              network & network &                           & best LPIPS $\downarrow$ & best PSNR $\uparrow$ & best LPIPS $\downarrow$ & best PSNR $\uparrow$ \\ \midrule
(a) icDPM-S             & -                             & ch=32                          & 6M                        & 0.077                   & 30.555               & 0.150                   & 28.209               \\
(b) icDPM-L             & -                             & ch=64                          & 27M                       & 0.058                   & 32.105               & 0.156                   & 27.996               \\
\midrule
(c) icDPM-S w/ Guide-S     & ch=32                         & ch=32                          & 10M                       & 0.068                   & 31.298               & 0.145                   & 28.286               \\
(d) icDPM-L w/ Guide-S    & ch=32                         & ch=64                          & 30M                       & 0.058                   & 32.220               & 0.128                   & 28.742               \\
(e) icDPM-L w/ Guide-L             & ch=64                         & ch=64                          & 52M                       & 0.057                   & 32.254               & 0.123                   & 28.711               \\ \bottomrule
\vspace{-1em}
\end{tabular}
\end{table*}
As the guidance network introduces more parameters, we investigate if its performance improvement is solely due to larger models. We perform a joint analysis of varying model size with and without the guidance network, and results are presented in Table~\ref{table:ablation_network_size}. We refer results on the GoPro test set as `In-domain' and on the Realblur-J dataset as `Out-of-domain', using a single GoPro-trained model.
We keep the number of building blocks constant and modulate the network size by changing only the number of convolutional filters. `-S' and `-L' indicate a smaller and larger models, respectively.
\begin{figure}[ht]
    \centering
    \includegraphics[width=0.8\linewidth]{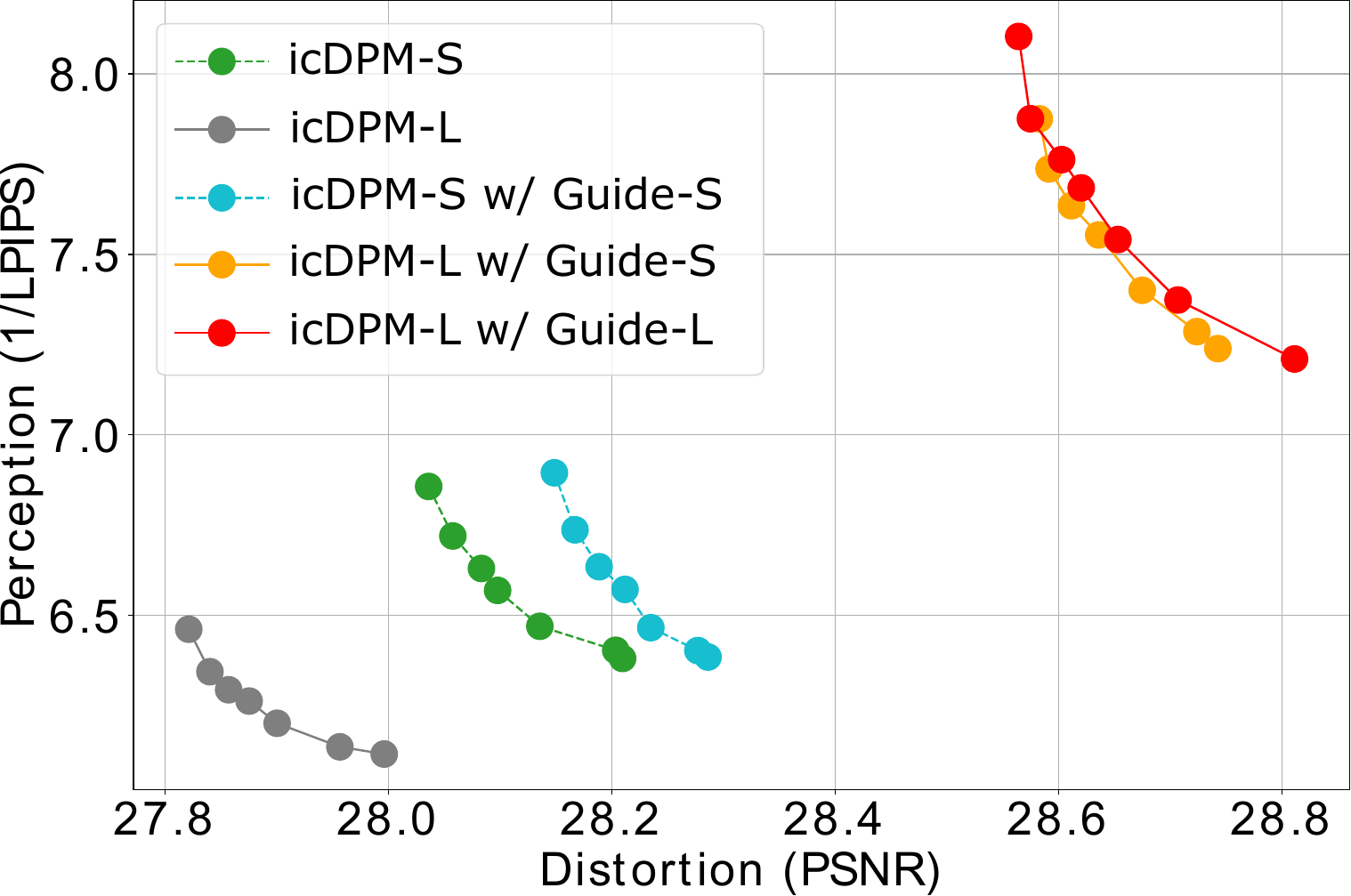}
    \caption{Perception-distortion plot as supplementary for Table~\ref{table:ablation_network_size}, under varying sampling parameters. Under different network capacities (`-S' and `-L' refer to small and large respectively), the guidance mechanism allows for consistently better perceptual quality and lower distortions compared to icDPM. Plots for other datasets are in supplementary, where we observe similar trends.
    }
    \label{fig:pd_curve}
\end{figure}
We start with the image-conditioned DPM (icDPM) without the proposed guidance network under different network sizes. In Table~\ref{table:ablation_network_size} row $(a)$ and $(b)$, we observe a significant improvement of the in-domain deblurring performance by increasing the UNet capacity, in terms of both perception and distortion qualities.
However, the out-of-domain testing results become much worse with a larger network, suggesting potential overfitting during the training. Through visual inspection, we also found that the larger DPM is prone to artifacts when presented with unseen data, as shown in Fig.~\ref{fig:teaser},~\ref{fig:regression} and ~\ref{fig:red_results}. 
By introducing the guidance network, we observe both in-domain and out-of-domain performance gains. To isolate the effects of introducing guidance module, we also compare the performance between the icDPM and icDPM with guidance under similar amount of parameters, where we reduce the parameters of the model by using a smaller guidance module (from (e) to (d)). We observe comparable results of (d) icDPM-L w/ Guide-S with (e) icDPM-L w/ Guide-L in both in-domain and out-of-domain performance, and both of them outperformed icDPM-L.

We additionally present a distortion-perception plot in Fig.~\ref{fig:pd_curve}, with samples acquired from varying sampling parameters (i.e. number of steps and the standard deviation of noise). Similar to~\cite{whang2022deblurring}, we found a general trade-off between perceptual quality and distortion metrics. Also, we observed that all guided models consistently outperform the baseline DPMs under varying sampling parameters.
We provide additional results on other datasets in appendix and observe similar effects and benefits of using the guidance module over the baseline icDPM.

\begin{table}[t]
\footnotesize
\setlength{\tabcolsep}{1pt}
\centering
\caption{Average deblurring results across GoPro~\cite{nah2017deep}, HIDE \cite{shen2019human} Realblur-J~\cite{rim2020real}, REDS~\cite{Nah_2019_CVPR_Workshops_REDS} dataset with GoPro~\cite{nah2017deep}-only trained model, indicating the model robustness on various unseen data. 
}
\label{table:average_results}
\begin{tabular}{lcccccc}
\toprule

& \multicolumn{4}{c}{\textbf{Perceptual}}
& \multicolumn{2}{c}{\textbf{Distortion}} \\

\cmidrule(lr){2-5} \cmidrule(lr){6-7}

& LPIPS $\downarrow$ & NIQE $\downarrow$ & FID $\downarrow$ & KID $\downarrow$ & PSNR $\uparrow$ & SSIM $\uparrow$ \\

\midrule

DeblurGAN-v2 \cite{Kupyn_2019_ICCV_Deblurganv2}
& 0.149   & 3.42     & 14.57    & 5.28   & 28.09    & 0.871 \\

MPRNet \cite{Zamir2021MPRNet}
& 0.140   & 3.70      & 20.22   & 8.49   &  29.78    & 0.897 \\
UFormer~\cite{Wang_2022_CVPR_uformer}   
& 0.133   & 3.65      & 18.99   & 8.13   & \firstone{30.06}    & \firstone{0.903} \\

Restormer~\cite{Zamir2021Restormer} 
& 0.139   & 3.69      & 19.90   & 8.36   & 30.00    & 0.895 \\
Ours-SA
& 0.124  & 3.64       & 14.36    &6.79 & 29.98  & 0.902 \\
Ours   
& \firstone{0.104}   & \firstone{2.94}  & \firstone{8.41}    & \firstone{2.39} & 28.81  &  0.881 \\
\bottomrule
\vspace{-2em}
\end{tabular}
\end{table}
\begin{figure*}[ht]
    \centering
    \includegraphics[width=0.96\linewidth]{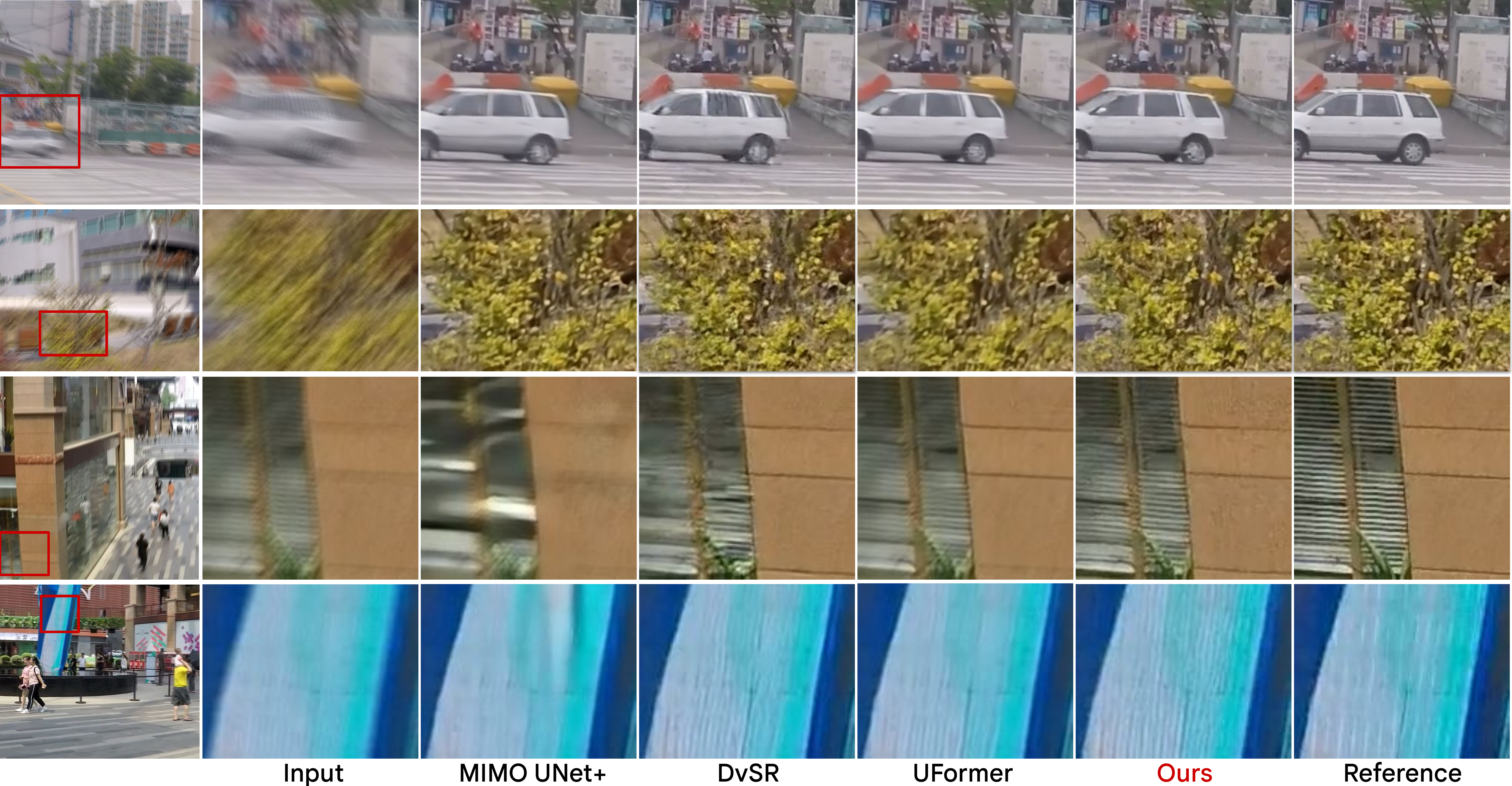}
    \caption{Deblurring results on GoPro~\cite{nah2017deep} (top two rows) and HIDE~\cite{shen2019human} (bottom two rows) test set from MIMO UNet+~\cite{cho2021rethinking_mimounet}, DvSR~\cite{whang2022deblurring},  UFormer~\cite{Wang_2022_CVPR_uformer}, and Ours. All models are trained only on GoPro~\cite{nah2017deep} training set. Our method generates perceptually much sharper images, and reduce artifacts when applied to unseen images (HIDE). Enlarged results are provided in the appendix.}
    \label{fig:gopro_hide_results}
\end{figure*}
\begin{figure*}[ht]
    \centering
    \includegraphics[width=0.96\linewidth]{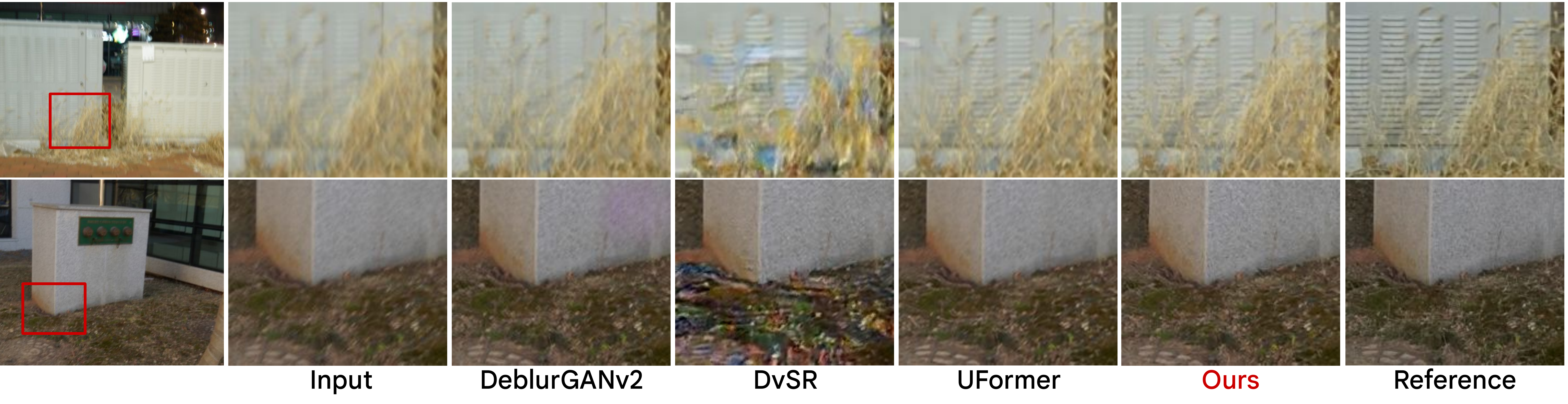}
    \caption{Test examples on Realblur-J~\cite{rim2020real}, with all models trained only on GoPro dataset~\cite{nah2017deep}. Best viewed electronically. We empirically found DeblurGANv2~\cite{Kupyn_2019_ICCV_Deblurganv2} and DvSR~\cite{whang2022deblurring} are prone to artifacts, and regression-based UFormer~\cite{Wang_2022_CVPR_uformer} produces over-smoothing images. Our methods alleviate artifacts, and produce high-fidelity deblurring on unseen data, even when the domain gap is large.}
    \label{fig:realblur_results}
\end{figure*}
\begin{figure*}[t]
    \centering
    \includegraphics[width=0.96\linewidth]{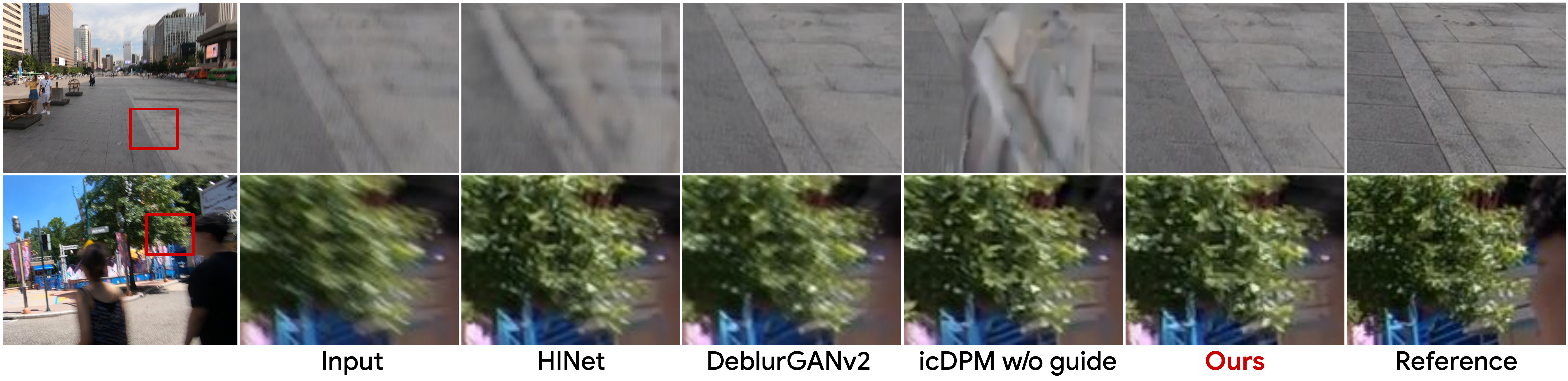}
    \caption{\textbf{REDS}~\cite{Nah_2019_CVPR_Workshops_REDS} deblurring examples from HINet~\cite{Chen_2021_CVPR_HINet}, DeblurGAN-v2~\cite{Kupyn_2019_ICCV_Deblurganv2}, icDPM without guidance and Ours. When trained only on GoPro~\cite{nah2017deep}, our method better removes the blur trace from the input image, while eliminating artifacts when applied on unseen data.}
    \label{fig:red_results}
\end{figure*}
\subsection{Deblurring results}
\label{sec:benchmark}
We compare our deblurring results with the state-of-the-art methods, loosely categorized into distortion-driven models~\cite{Chen_2021_CVPR_HINet,Zamir2021MPRNet,cho2021rethinking_mimounet}, perception-driven GAN based methods~\cite{DeblurGAN,Kupyn_2019_ICCV_Deblurganv2}, as well as recent diffusion-based method~\cite{whang2022deblurring}. In this work, we place particular emphasis on evaluating (1) the generalization ability on unseen data, and (2) the perceptual quality of the output, willing to compromise a slight drop on average distortion scores towards a better trade-off between distortion and perception~\cite{blau2018perception}. For benchmarking, we mainly consider perceptual quality of the results quantified by standard metrics for generative models including LPIPS~\cite{zhang2018perceptual}, NIQE \cite{mittal2012making}, FID (Fréchet Inception Distance)~\cite{heusel2017gans}, and KID (Kernel Inception Distance)~\cite{binkowski2018demystifying}.
We also present distortion metrics including PSNR and SSIM for completeness. 
However, we note that they are less correlated with human perception~\cite{nilsson2020understanding} and maximizing PSNR/SSIM results in a compromise of visual perception~\cite{blau2018perception}. Our method is based on generative models and performs stochastic posterior sampling. The reference image provided in the training dataset is only one of the possible restoration result among other possibilities (due to the ill-posed nature of inverse problems). Thus, similar to~\cite{blau2018perception,ohayon2021high}, our results compromise a certain amount of pixelwise average distortion while still being faithful to the target.
We highlight \firstone{best} and \secondone{second-best} values for each metric. KID values are scaled by a factor of 1000 for readability.
\begin{table}[bht]
\footnotesize
\setlength{\tabcolsep}{1pt}
\centering
\caption{Image deblurring results on GoPro~\cite{nah2017deep} dataset. 
}
\label{table:gopro_results}
\begin{tabular}{lcccccc}
\toprule

& \multicolumn{4}{c}{\textbf{Perceptual}}
& \multicolumn{2}{c}{\textbf{Distortion}} \\

\cmidrule(lr){2-5} \cmidrule(lr){6-7}

& LPIPS $\downarrow$ & NIQE $\downarrow$ & FID $\downarrow$ & KID $\downarrow$ & PSNR $\uparrow$ & SSIM $\uparrow$ \\

\midrule

HINet \cite{Chen_2021_CVPR_HINet}
& 0.088   & 4.01     & 17.91    & 8.15
& 32.77   & 0.960 \\

MPRNet \cite{Zamir2021MPRNet}
& 0.089   & 4.09     & 20.18    & 9.10
& 32.66    & 0.959 \\

MIMO-UNet+ \cite{cho2021rethinking_mimounet}
& 0.091   & 4.03     & 18.05    & 8.17 
& 32.44    & 0.957 \\

SAPHNet \cite{suin2020spatially}
& 0.101   & 3.99     & 19.06    & 8.48
& 31.89   & 0.953 \\

SimpleNet \cite{li2021perceptual_simplenet} 
& 0.108   & -     & -     & -
& 31.52     & 0.950 \\

DeblurGANv2 \cite{Kupyn_2019_ICCV_Deblurganv2}
& 0.117   & 3.68     & 13.40    & 4.41
& 29.08    & 0.918 \\

DvSR~\cite{whang2022deblurring}
& \secondone{0.059}  & \secondone{3.39}
& \secondone{4.04} & \secondone{0.98}
& 31.66    & 0.948 \\

DvSR-SA~\cite{whang2022deblurring}
& 0.078   & 4.07
& 17.46    & 8.03
& \firstone{33.23}   & \secondone{0.963} \\

UFormer~\cite{Wang_2022_CVPR_uformer} &  
0.087 &  4.08
 & 19.66 & 9.09  
& 32.97 & \firstone{0.967} \\

Restormer~\cite{Zamir2021Restormer} & 0.084 &  4.12 & 19.33 & 8.75 & 
32.92 & 0.961 \\
Ours-SA    
& 0.078  & 4.10 & 8.69  & 7.06 & \secondone{33.20} & \secondone{0.963}\\
Ours  & \firstone{0.057} & \firstone{3.27} & \firstone{3.50} & \firstone{0.77}  &  31.19 & 0.943 \\

\bottomrule
\vspace{-2em}
\end{tabular}
\end{table}
\begin{table}[ht]
\footnotesize
\setlength{\tabcolsep}{1pt}
\centering
\caption{\centering Results on Realblur-J~\cite{rim2020real} with GoPro trained models.}
\label{table:realblurj}
\begin{tabular}{lcccccc}
\toprule
& \multicolumn{4}{c}{\textbf{Perceptual}}
& \multicolumn{2}{c}{\textbf{Distortion}} \\

\cmidrule(lr){2-5} \cmidrule(lr){6-7}

                   & LPIPS $\downarrow$ & NIQE $\downarrow$ & FID $\downarrow$ & KID $\downarrow$ & PSNR $\uparrow$ & SSIM $\uparrow$ \\
\midrule
UNet~\cite{ronneberger2015u}    & 0.175   &  3.911     & 22.24  & 8.07
                   & 28.06           & 0.857            
\\
DeblurGAN~\cite{DeblurGAN}          &  -     & -         & -                & -           
                   & 27.97           & 0.834           
\\
DeblurGAN-v2~\cite{Kupyn_2019_ICCV_Deblurganv2}
                    &  \secondone{0.139} & 3.870         & \secondone{14.40} & \secondone{4.64}
                   & 28.70           & 0.866           
\\
MPRNet~\cite{Zamir2021MPRNet}
                  &  0.153  & 3.967    & 20.25 & 7.57    
                   & 28.70           & 0.873           
\\
DvSR~\cite{whang2022deblurring}
                   & 0.153   & \secondone{3.277}        & 18.73   & 6.00        
                   & 28.02           & 0.851            
\\
DvSR-SA~\cite{whang2022deblurring}
                   & 0.156  & 3.783         & 20.09   & 7.43
                   & 28.46           & 0.863            
\\
Restormer~\cite{Zamir2021Restormer}          
                    & 0.149  &3.916         & 19.55   & 7.12           
                   & \secondone{28.96}           & \secondone{0.879}            
\\
UFormer-B~\cite{Wang_2022_CVPR_uformer} 
            & 0.140  & 3.857        & 18.56  &  7.02
                   & \firstone{29.06}           & \firstone{0.884}           
\\
Ours-SA        & 0.139  & 3.809  & 16.84  & 6.25
                   & 28.81           & 0.872
\\
Ours     & \firstone{0.123} & \firstone{2.976 } & \firstone{12.95} & \firstone{3.58}        
                  & 28.56           & 0.862
      \\
\bottomrule
\end{tabular}
\end{table}
\begin{table}[ht]
\footnotesize
\setlength{\tabcolsep}{1pt}
\centering
\caption{\centering Results on HIDE \cite{shen2019human} with GoPro~\cite{nah2017deep} trained models. 
}
\label{table:hide_results}
\begin{tabular}{lcccccc}
\toprule

& \multicolumn{4}{c}{\textbf{Perceptual}}
& \multicolumn{2}{c}{\textbf{Distortion}} \\

\cmidrule(lr){2-5} \cmidrule(lr){6-7}

& LPIPS $\downarrow$ & NIQE $\downarrow$ & FID $\downarrow$ & KID $\downarrow$ & PSNR $\uparrow$ & SSIM $\uparrow$ \\

\midrule

HINet \cite{Chen_2021_CVPR_HINet}
& 0.120   & 3.20     & 15.17    & 7.33   & 30.33    & 0.932 \\

MIMO-UNet+ \cite{cho2021rethinking_mimounet}
& 0.124   & 3.24     & 16.01    & 7.91   & 29.99    & 0.930 \\

MPRNet \cite{Zamir2021MPRNet}
& 0.114   & 3.46     & 16.58    & 8.35   & \secondone{30.96}   & \firstone{0.940} \\

SAPHNet \cite{suin2020spatially}
& 0.128   & 3.21     & 16.78    & 8.39   & 29.99    & 0.930 \\

DeblurGAN-v2 \cite{Kupyn_2019_ICCV_Deblurganv2}
& 0.159   & 2.96     & 15.51    & 6.96   & 27.51    & 0.885 \\

DvSR-SA~\cite{whang2022deblurring}
& 0.105  & 3.29    & 15.34    & 8.00  & 30.94    & \firstone{0.940} \\
DvSR~\cite{whang2022deblurring}
& \secondone{0.089}   & \firstone{2.69}      & \secondone{5.43}    & \firstone{1.61}    & 29.77    & 0.922 \\

UFormer~\cite{Wang_2022_CVPR_uformer}   & 0.113 &  3.40 & 16.27 & 8.51 & 30.89 & 0.920 \\
Restormer~\cite{Zamir2021Restormer} & 0.108 &  3.41 & 15.84 & 8.28 & \firstone{31.22} & 0.923 \\
Ours-SA
&0.104  &   3.40    & 14.62  & 7.62      & \secondone{30.96}  &  \secondone{0.938}
      \\
Ours   
&\firstone{0.088} & \secondone{2.91} &  \firstone{5.28} & \secondone{1.68} & 29.14 &  0.910\\

\bottomrule

\end{tabular}
\end{table}
\begin{table}[t]
\footnotesize
\setlength{\tabcolsep}{1pt}
\centering
\caption{\centering Results on REDS~\cite{Nah_2019_CVPR_Workshops_REDS} with GoPro-trained models.} 
\label{table:reds}
\begin{tabular}{lcccccc}
\toprule

& \multicolumn{4}{c}{\textbf{Perceptual}}
& \multicolumn{2}{c}{\textbf{Distortion}} \\

\cmidrule(lr){2-5} \cmidrule(lr){6-7}

                   & LPIPS $\downarrow$ & NIQE $\downarrow$ & FID $\downarrow$ & KID $\downarrow$ & PSNR $\uparrow$ & SSIM $\uparrow$ \\
\midrule
HINet~\cite{Chen_2021_CVPR_HINet} & 0.195 & 3.223 & 21.48 & 7.91 &
                    26.72        & 0.818 \\
DeblurGAN-v2~\cite{Kupyn_2019_ICCV_Deblurganv2}
                    &  0.181 & \secondone{3.172}      & \secondone{14.98} & \secondone{5.12}
                   & \secondone{27.08}           & 0.814         
\\
MPRNet~\cite{Zamir2021MPRNet}
                  &  0.204  & 3.282    & 23.90 & 8.94    
                   & 26.80           & 0.814        
\\
Restormer~\cite{Zamir2021Restormer}          
                    & 0.213  & 3.326        & 24.86   & 9.30       
                   & 26.91         & 0.818          
\\
UFormer-B~\cite{Wang_2022_CVPR_uformer} 
            & 0.192 &  3.272     & 21.48  &  7.91
                   &\firstone{27.31}          & \firstone{0.842}           
\\
Ours-SA        & \secondone{0.178}  & 3.248 & 17.27  & 6.21
                   & 26.95           & \secondone{0.834}
\\
Ours     & \firstone{0.147} & \firstone{2.610}  & \firstone{11.91} & \firstone{3.51}        
                  & 26.36          & 0.810
      \\ 
\bottomrule
\vspace{-2em}
\end{tabular}
\end{table}

We first present the in-domain GoPro performance in Table~\ref{table:gopro_results}. Our model achieved state-of-the-art perceptual metrics across the board, while maintaining competitive distortion metrics by taking average of multiple samples (`Ours-SA'). Moreover, we are interested in the domain generalization and out-of-domain results on Realblur-J (Table~\ref{table:realblurj}), REDS (Table~\ref{table:reds}) and HIDE (Table~\ref{table:hide_results}). We achieved a significantly better perceptual quality on the unseen Realblur-J and REDS, and competitive results on HIDE. Further, we analyze the robustness the best-performing single-dataset trained models by comparing their average performance across all four test sets summarized in Table~\ref{table:average_results}. Our method significantly improves the perceptual scores, while maintaining highly competitive distortion scores with $<0.08$ dB difference from the best PSNR, and $<0.001$ difference from the best SSIM with averaging samples (`Ours-SA').

Visual deblurring examples are provided in Fig.~\ref{fig:gopro_hide_results} on GoPro~\cite{nah2017deep} and HIDE~\cite{shen2019human}, Fig.~\ref{fig:realblur_results} on RealblurJ~\cite{rim2020real} and Fig.~\ref{fig:red_results} on REDS~\cite{Nah_2019_CVPR_Workshops_REDS}, respectively. On the GoPro (in-domain) test example, we find that all methods are able to produce reasonable artifact-free deblurring, and our method generates sharper and more visual realistic results. On the three out-of-domain datasets, performance degradation starts to occur from baseline methods. For instance, GAN-based model~\cite{Kupyn_2019_ICCV_Deblurganv2} and previous diffusion based model~\cite{whang2022deblurring} tend to produce artifacts on out-of-domain data, and state-of-the-art regression based model~\cite{Wang_2022_CVPR_uformer} produces over-smoothed results. Our formulation performs more consistently better across different datasets, significantly reducing artifacts on unseen data with high perceptual realism. More enlarged visual examples are in the appendix.

\subsection{Perceptual Study}
\label{sec:user_study}
We further conducted a user study with human subjects to verify the perceptual quality of the deblurring performance on unseen data, with all models trained on GoPro and tested on Relblur-J. 
We asked Amazon Mechanical Turk raters to select the best quality image from a given pair. We used 30 unique pairs of size $512\times512$ and averaged the 750 ratings from 25 raters. 
In Table~\ref{table:perceptual_study}, each value represents the fraction of times that the raters preferred the row over the column. As can be seen, our method outperforms the existing solutions.
Also, it is worth pointing out that there is a significant gap in the preference of our method with and without the guidance mechanism (denoted as icDPM).
\begin{table}[h]
\footnotesize
\setlength{\tabcolsep}{2pt}
\renewcommand{\arraystretch}{0.5}
    \centering
    \caption{Perceptual study with human subjects on Realblur-J~\cite{rim2020real} dataset using GoPro~\cite{nah2017deep} trained models. Each value represents the fraction of times that raters preferred the row over the column. }
    \label{table:perceptual_study}
    \begin{tabular}{lP{12mm}P{12mm}P{10mm}P{8mm}P{8mm}P{8mm}} \toprule
             & DGANv2 \cite{Kupyn_2019_ICCV_Deblurganv2} & UFormer	\cite{Wang_2022_CVPR_uformer}& DvSR \cite{whang2022deblurring}& icDPM & Ours-SA & Ours \\\midrule
     DGANv2 & - & 0.28 & 0.43  & 0.56 & 0.26 & 0.17 \\
     UFormer & 0.72 & - & 0.53 & 0.58  & 0.44 & 0.35 \\
     DvSR & 0.57 & 0.47 & - & 0.61  & 0.32 & 0.29 \\
     icDPM & 0.44 & 0.42 & 0.39 & - & 0.28 & 0.21 \\
     Ours-SA & 0.74 & 0.56 &  0.68 & 0.72  & - & 0.38 \\
     Ours & \firstone{0.83} & \firstone{0.65} & \firstone{0.71} & \firstone{0.79}  & \firstone{0.62} & - \\
     \bottomrule
    \end{tabular}
\end{table}

\subsection{Additional modeling choices}
\label{sec:ablation}
\textbf{Guidance network.} We carried out additional ablation studies for the modeling choices of the guidance network on regression target (RGB v.s. grayscale), the number of scales to adopt for the guidance (single v.s. multiscale), and the mechanism of incorporating the guidance (input-level vs latent space). For fair comparison, identical diffusion UNet is used under different configurations during training, and the same sampling parameters are used during inference.
Table~\ref{table:ablation_guidance} (a) indicates our baseline icDPM without any guidance.
We first compare the difference between incorporating the guidance at input-level and at latent space (Table~\ref{table:ablation_guidance} (b) and (c)). In (b), we upscale the regression output to the original input size, and concatenate the result to the diffusion UNet. 
In (c), we incorporate the feature maps before regression output into the UNet latent space via addition operation described above. The results indicate the benefit of latent-space guidance over input-level concatenation. Both (b)(c) improve on (a), showing the overall benefit of introducing the guidance.
 \begin{table}[hb]
\footnotesize
\setlength{\tabcolsep}{3pt}
\centering
\vspace{-.5em}
\caption{Effect of various settings on the domain invariant guidance. The scale column denotes the downsampling factors.}
\label{table:ablation_guidance}
\begin{tabular}{cccccc}
\toprule
 & Regression & Scale(s)       & Guidance      & LPIPS & PSNR \\
\midrule
 (a)  & -         &  -           &  -            & 0.156 & 28.21  \\
 (b)  & RGB       &  $\times$8          &  input        & 0.145 & 28.34  \\
 (c)  & RGB       &  $\times$8          &  latent       & 0.143 & 28.45  \\
 (d)  & RGB       &  $\times$2, $\times$4, $\times$8  &  latent       & 0.141 & 28.45  \\
 (e)  & Grayscale &  $\times$2, $\times$4, $\times$8  &  latent       & 0.137 & 28.63  \\
\bottomrule
\vspace{-2em}
\end{tabular}
\end{table}
We also observe a moderate improvement by using multiscale guidance rather than single scale guidance in row (d) over (c). In row (e), we simplified the regression target from color space to grayscale space, which further improved the results.

\textbf{Guidance operation.} 
We performed ablation studies on the operations that injects the guidance into the icDPM backbone. We tested three different possible guidance operations including addition (our final choice), concatenation, and adaptive group normalization. Table~\ref{tab:guide_op} shows the deblurring PSNR on GoPro, where addition achieves the best performance. 
Empirically, we observe that adaptive norm in our use case tends to introduce artifacts. Compared with concatenation, addition is more memory-efficient, and ensures that the guidance cannot be simply neglected.
\begin{table}[h]
\vspace{-0.5em}
\footnotesize
    \centering
    \caption{Ablation on guidance operations.}
    \begin{tabular}{lc}
    \toprule
      Operation  &  PSNR \\ \midrule
      Addition      &  31.139\\
      Concatenate   &  30.248\\
      Adaptive norm &  29.676\\
      \bottomrule
    \end{tabular}
    \label{tab:guide_op}
    \vspace{-0.5em}
\end{table}

\textbf{Training objectives for guidance feature.}
As we aim to enhance the icDPM backbone with a more robust conditioning mechanism  through the integration of multiscale structure guidance, we introduce a regression loss to jointly train the guidance module and the icDPM. We assume that this loss is essential to ensure that the derived guidance representation retains prominent structural features while filtering out extraneous information. To verify our assumption, we trained our model without the regression loss, and we observed that the PSNR on GoPro test set degraded from 31.19 to 30.56. This indicates that the regression loss is crucial to effectively constraint learning of the guidance, and removing the loss potentially makes the model equivalent to just a larger UNet architecture.

\begin{figure}[h]
    \centering
    \vspace{-1mm}
    \includegraphics[width=0.78\linewidth]{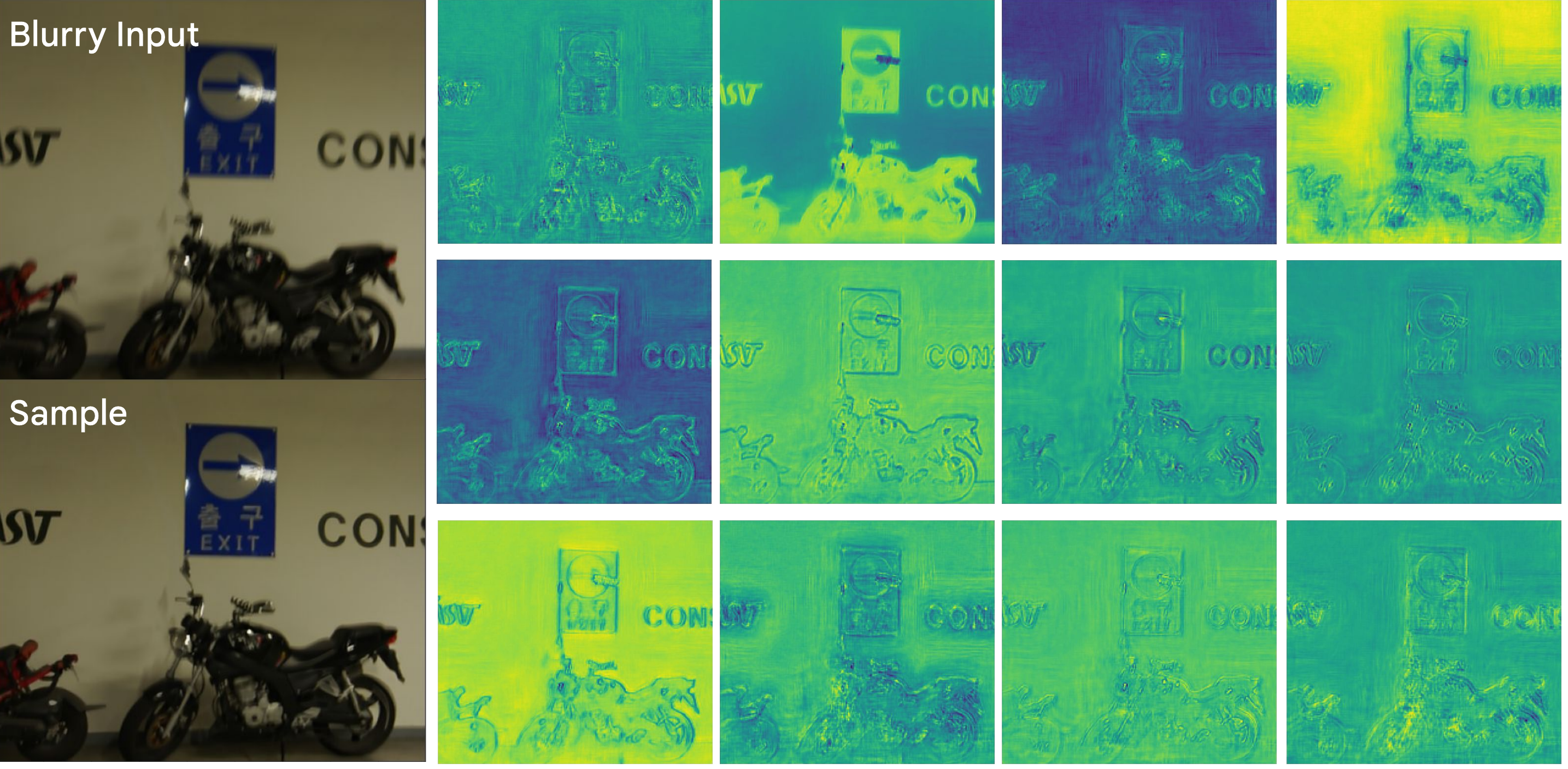}
    \caption{An out-of-domain deblurring result on a test image from Realblur-J~\cite{rim2020real} (left), alongside 12 (out of 64) selected channelwise guidance feature maps at the scale of $k=1$ (right). }
    \label{fig:feature_map}
\end{figure}
Qualitatively, we examine the learned channel-wise guidance feature maps depicted in Fig.~\ref{fig:feature_map}. As expected, these feature maps are highly related to edges and overall structures, which provide as auxiliary information to the icDPM about the coarse structures of its sharp reconstruction, and eventually benefit the robustness of the model. 

\section{Discussion}
We present a learned multiscale structure guidance mechanism for icDPM that acts as an implicit bias which enhances its deblurring robustness. We acknowledge that limitations exist and require further investigation. 

While we focus on improving the model's generalization to unseen data without access to large-scale realistic training data, we recognize that the quality and realism of the training dataset ultimately bounds the capability of the model. 
In our setup, we are restricted to the GoPro training dataset for benchmarking, which does not adequately cover all real-world scenarios, such as saturated regions with poor light conditions. 
Almost all methods fail on deblurring such images and examples are in appendix. We believe our method can further benefit from large-scale diverse sets of training data.
Further, the scope of our work is specifically on improving the robustness of deblurring in the context of icDPM where the deblurring task is cast as a \textit{conditional generation problem}, similar ideas could be explored in different contexts, e.g., making a regression model more robust.
However, as the guidance module is also trained with regression objectives, attaching it to a regression model potentially leads to a single model with increased number of parameters. Such formulation along with other extensions (e.g. plugging the guidance into other SoTA regression backbones like transformers) require further investigations.

\noindent \textbf{Acknowledgments} Co-author Guido Gerig acknowledges support from the New York Center for Advanced Technology in Telecommunications (CATT).
\newpage
{\small
\bibliographystyle{ieee_fullname}
\bibliography{egbib}

\begin{thebibliography}{100}\itemsep=-1pt

\bibitem{anger2019efficient}
J{\'e}r{\'e}my Anger, Mauricio Delbracio, and Gabriele Facciolo.
\newblock Efficient blind deblurring under high noise levels.
\newblock In {\em 2019 11th International Symposium on Image and Signal Processing and Analysis (ISPA)}, pages 123--128. IEEE, 2019.

\bibitem{bansal2022cold}
Arpit Bansal, Eitan Borgnia, Hong-Min Chu, Jie~S Li, Hamid Kazemi, Furong Huang, Micah Goldblum, Jonas Geiping, and Tom Goldstein.
\newblock Cold diffusion: Inverting arbitrary image transforms without noise.
\newblock {\em arXiv preprint arXiv:2208.09392}, 2022.

\bibitem{binkowski2018demystifying}
Mikołaj Bińkowski, Dougal~J. Sutherland, Michael Arbel, and Arthur Gretton.
\newblock Demystifying {MMD} {GAN}s.
\newblock In {\em International Conference on Learning Representations}, 2018.

\bibitem{blau2018perception}
Yochai Blau and Tomer Michaeli.
\newblock The perception-distortion tradeoff.
\newblock In {\em Proceedings of the IEEE conference on computer vision and pattern recognition}, pages 6228--6237, 2018.

\bibitem{bruna2016super}
Joan Bruna, Pablo Sprechmann, and Yann LeCun.
\newblock Super-resolution with deep convolutional sufficient statistics.
\newblock In {\em International Conference on Learning Representations}, 2016.

\bibitem{Chen_2021_CVPR_HINet}
Liangyu Chen, Xin Lu, Jie Zhang, Xiaojie Chu, and Chengpeng Chen.
\newblock Hinet: Half instance normalization network for image restoration.
\newblock In {\em Proceedings of the IEEE/CVF Conference on Computer Vision and Pattern Recognition (CVPR) Workshops}, pages 182--192, June 2021.

\bibitem{cho2021rethinking_mimounet}
Sung-Jin Cho, Seo-Won Ji, Jun-Pyo Hong, Seung-Won Jung, and Sung-Jea Ko.
\newblock Rethinking coarse-to-fine approach in single image deblurring.
\newblock In {\em Proceedings of the IEEE/CVF International Conference on Computer Vision (ICCV)}, pages 4641--4650, October 2021.

\bibitem{chung2022improving}
Hyungjin Chung, Byeongsu Sim, Dohoon Ryu, and Jong~Chul Ye.
\newblock Improving diffusion models for inverse problems using manifold constraints.
\newblock {\em arXiv preprint arXiv:2206.00941}, 2022.

\bibitem{chung2022come}
Hyungjin Chung, Byeongsu Sim, and Jong~Chul Ye.
\newblock Come-closer-diffuse-faster: Accelerating conditional diffusion models for inverse problems through stochastic contraction.
\newblock In {\em Proceedings of the IEEE/CVF Conference on Computer Vision and Pattern Recognition}, pages 12413--12422, 2022.

\bibitem{diffusion_survey}
Florinel-Alin Croitoru, Vlad Hondru, Radu~Tudor Ionescu, and Mubarak Shah.
\newblock {Diffusion Models in Vision: A Survey}.
\newblock {\em arXiv preprint arXiv:2209.04747}, 2022.

\bibitem{daras_dagan_2022score}
Giannis Daras, Yuval Dagan, Alexandros~G Dimakis, and Constantinos Daskalakis.
\newblock Score-guided intermediate layer optimization: Fast langevin mixing for inverse problem.
\newblock {\em arXiv preprint arXiv:2206.09104}, 2022.

\bibitem{daras2022soft}
Giannis Daras, Mauricio Delbracio, Hossein Talebi, Alexandros~G Dimakis, and Peyman Milanfar.
\newblock Soft diffusion: Score matching for general corruptions.
\newblock {\em arXiv preprint arXiv:2209.05442}, 2022.

\bibitem{delbracio2023inversion}
Mauricio Delbracio and Peyman Milanfar.
\newblock Inversion by direct iteration: An alternative to denoising diffusion for image restoration.
\newblock {\em Transactions on Machine Learning Research}, 2023.
\newblock Featured Certification.

\bibitem{delbracio2021projected}
Mauricio Delbracio, Hossein Talebei, and Pevman Milanfar.
\newblock Projected distribution loss for image enhancement.
\newblock In {\em 2021 IEEE International Conference on Computational Photography (ICCP)}, pages 1--12. IEEE, 2021.

\bibitem{dhariwal2021diffusion}
Prafulla Dhariwal and Alexander Nichol.
\newblock Diffusion models beat gans on image synthesis.
\newblock {\em Advances in Neural Information Processing Systems}, 34:8780--8794, 2021.

\bibitem{dockhorn2022genie}
Tim Dockhorn, Arash Vahdat, and Karsten Kreis.
\newblock Genie: Higher-order denoising diffusion solvers.
\newblock {\em arXiv preprint arXiv:2210.05475}, 2022.

\bibitem{fergus2006removing}
Rob Fergus, Barun Singh, Aaron Hertzmann, Sam~T Roweis, and William~T Freeman.
\newblock Removing camera shake from a single photograph.
\newblock In {\em Acm Siggraph 2006 Papers}, pages 787--794. 2006.

\bibitem{goodfellow2014generative}
Ian Goodfellow, Jean Pouget-Abadie, Mehdi Mirza, Bing Xu, David Warde-Farley, Sherjil Ozair, Aaron Courville, and Yoshua Bengio.
\newblock Generative adversarial nets.
\newblock {\em Advances in neural information processing systems}, 27, 2014.

\bibitem{heusel2017gans}
Martin Heusel, Hubert Ramsauer, Thomas Unterthiner, Bernhard Nessler, and Sepp Hochreiter.
\newblock Gans trained by a two time-scale update rule converge to a local nash equilibrium.
\newblock In I. Guyon, U.~V. Luxburg, S. Bengio, H. Wallach, R. Fergus, S. Vishwanathan, and R. Garnett, editors, {\em Advances in Neural Information Processing Systems}, volume~30. Curran Associates, Inc., 2017.

\bibitem{ho2020denoising}
Jonathan Ho, Ajay Jain, and Pieter Abbeel.
\newblock Denoising diffusion probabilistic models.
\newblock {\em Advances in Neural Information Processing Systems}, 33:6840--6851, 2020.

\bibitem{ho2021classifier}
Jonathan Ho and Tim Salimans.
\newblock Classifier-free diffusion guidance.
\newblock In {\em NeurIPS 2021 Workshop on Deep Generative Models and Downstream Applications}, 2021.

\bibitem{ho2022video}
Jonathan Ho, Tim Salimans, Alexey Gritsenko, William Chan, Mohammad Norouzi, and David~J Fleet.
\newblock Video diffusion models.
\newblock {\em arXiv:2204.03458}, 2022.

\bibitem{hoffman2018cycada}
Judy Hoffman, Eric Tzeng, Taesung Park, Jun-Yan Zhu, Phillip Isola, Kate Saenko, Alexei Efros, and Trevor Darrell.
\newblock Cycada: Cycle-consistent adversarial domain adaptation.
\newblock In {\em International conference on machine learning}, pages 1989--1998. Pmlr, 2018.

\bibitem{hoogeboom2022blurring}
Emiel Hoogeboom and Tim Salimans.
\newblock Blurring diffusion models.
\newblock {\em arXiv preprint arXiv:2209.05557}, 2022.

\bibitem{ji2022xydeblur}
Seo-Won Ji, Jeongmin Lee, Seung-Wook Kim, Jun-Pyo Hong, Seung-Jin Baek, Seung-Won Jung, and Sung-Jea Ko.
\newblock Xydeblur: Divide and conquer for single image deblurring.
\newblock In {\em Proceedings of the IEEE/CVF Conference on Computer Vision and Pattern Recognition}, pages 17421--17430, 2022.

\bibitem{jin2018normalized}
Meiguang Jin, Stefan Roth, and Paolo Favaro.
\newblock Normalized blind deconvolution.
\newblock In {\em Proceedings of the European Conference on Computer Vision (ECCV)}, pages 668--684, 2018.

\bibitem{johnson2016perceptual}
Justin Johnson, Alexandre Alahi, and Li Fei-Fei.
\newblock Perceptual losses for real-time style transfer and super-resolution.
\newblock In {\em European conference on computer vision}, pages 694--711. Springer, 2016.

\bibitem{kadkhodaie2021stochastic}
Zahra Kadkhodaie and Eero Simoncelli.
\newblock Stochastic solutions for linear inverse problems using the prior implicit in a denoiser.
\newblock {\em Advances in Neural Information Processing Systems}, 34:13242--13254, 2021.

\bibitem{karras2022elucidating}
Tero Karras, Miika Aittala, Timo Aila, and Samuli Laine.
\newblock Elucidating the design space of diffusion-based generative models.
\newblock {\em arXiv preprint arXiv:2206.00364}, 2022.

\bibitem{kawar2022ddrm}
Bahjat Kawar, Michael Elad, Stefano Ermon, and Jiaming Song.
\newblock Denoising diffusion restoration models.
\newblock In {\em Advances in Neural Information Processing Systems}, 2022.

\bibitem{kawar2022enhancing}
Bahjat Kawar, Roy Ganz, and Michael Elad.
\newblock Enhancing diffusion-based image synthesis with robust classifier guidance.
\newblock {\em arXiv preprint arXiv:2208.08664}, 2022.

\bibitem{kawar2022jpeg}
Bahjat Kawar, Jiaming Song, Stefano Ermon, and Michael Elad.
\newblock Jpeg artifact correction using denoising diffusion restoration models.
\newblock {\em arXiv preprint arXiv:2209.11888}, 2022.

\bibitem{kawar2021snips}
Bahjat Kawar, Gregory Vaksman, and Michael Elad.
\newblock Snips: Solving noisy inverse problems stochastically.
\newblock {\em Advances in Neural Information Processing Systems}, 34:21757--21769, 2021.

\bibitem{kingma2021variational}
Diederik Kingma, Tim Salimans, Ben Poole, and Jonathan Ho.
\newblock Variational diffusion models.
\newblock {\em Advances in neural information processing systems}, 34:21696--21707, 2021.

\bibitem{kingma2014adam}
Diederik~P Kingma and Jimmy Ba.
\newblock Adam: A method for stochastic optimization.
\newblock {\em arXiv preprint arXiv:1412.6980}, 2014.

\bibitem{krishnan2011blind}
Dilip Krishnan, Terence Tay, and Rob Fergus.
\newblock Blind deconvolution using a normalized sparsity measure.
\newblock In {\em CVPR 2011}, pages 233--240. IEEE, 2011.

\bibitem{DeblurGAN}
Orest Kupyn, Volodymyr Budzan, Mykola Mykhailych, Dmytro Mishkin, and Jiri Matas.
\newblock Deblurgan: Blind motion deblurring using conditional adversarial networks.
\newblock In {\em 2018 IEEE/CVF Conference on Computer Vision and Pattern Recognition}, pages 8183--8192. IEEE, 2018.

\bibitem{Kupyn_2019_ICCV_Deblurganv2}
Orest Kupyn, Tetiana Martyniuk, Junru Wu, and Zhangyang Wang.
\newblock Deblurgan-v2: Deblurring (orders-of-magnitude) faster and better.
\newblock In {\em The IEEE International Conference on Computer Vision (ICCV)}, Oct 2019.

\bibitem{laumont2022bayesian}
R{\'e}mi Laumont, Valentin~De Bortoli, Andr{\'e}s Almansa, Julie Delon, Alain Durmus, and Marcelo Pereyra.
\newblock Bayesian imaging using plug \& play priors: when langevin meets tweedie.
\newblock {\em SIAM Journal on Imaging Sciences}, 15(2):701--737, 2022.

\bibitem{ledig2017photo}
Christian Ledig, Lucas Theis, Ferenc Husz{\'a}r, Jose Caballero, Andrew Cunningham, Alejandro Acosta, Andrew Aitken, Alykhan Tejani, Johannes Totz, Zehan Wang, et~al.
\newblock Photo-realistic single image super-resolution using a generative adversarial network.
\newblock In {\em Proceedings of the IEEE conference on computer vision and pattern recognition}, pages 4681--4690, 2017.

\bibitem{levin2011efficient}
Anat Levin, Yair Weiss, Fredo Durand, and William~T Freeman.
\newblock Efficient marginal likelihood optimization in blind deconvolution.
\newblock In {\em CVPR 2011}, pages 2657--2664. IEEE, 2011.

\bibitem{li2022learning}
Dasong Li, Yi Zhang, Ka~Chun Cheung, Xiaogang Wang, Hongwei Qin, and Hongsheng Li.
\newblock Learning degradation representations for image deblurring.
\newblock In {\em European Conference on Computer Vision}, pages 736--753. Springer, 2022.

\bibitem{li2022srdiff}
Haoying Li, Yifan Yang, Meng Chang, Shiqi Chen, Huajun Feng, Zhihai Xu, Qi Li, and Yueting Chen.
\newblock Srdiff: Single image super-resolution with diffusion probabilistic models.
\newblock {\em Neurocomputing}, 479:47--59, 2022.

\bibitem{li2021perceptual_simplenet}
Jichun Li, Weimin Tan, and Bo Yan.
\newblock Perceptual variousness motion deblurring with light global context refinement.
\newblock In {\em Proceedings of the IEEE/CVF International Conference on Computer Vision (ICCV)}, pages 4116--4125, October 2021.

\bibitem{li2022efficient}
Muyang Li, Ji Lin, Chenlin Meng, Stefano Ermon, Song Han, and Jun-Yan Zhu.
\newblock Efficient spatially sparse inference for conditional gans and diffusion models.
\newblock {\em arXiv preprint arXiv:2211.02048}, 2022.

\bibitem{liu2021pseudo}
Luping Liu, Yi Ren, Zhijie Lin, and Zhou Zhao.
\newblock Pseudo numerical methods for diffusion models on manifolds.
\newblock In {\em International Conference on Learning Representations}, 2021.

\bibitem{lu2019unsupervised}
Boyu Lu, Jun-Cheng Chen, and Rama Chellappa.
\newblock Unsupervised domain-specific deblurring via disentangled representations.
\newblock In {\em Proceedings of the IEEE/CVF Conference on Computer Vision and Pattern Recognition}, pages 10225--10234, 2019.

\bibitem{lu2022dpm}
Cheng Lu, Yuhao Zhou, Fan Bao, Jianfei Chen, Chongxuan Li, and Jun Zhu.
\newblock Dpm-solver: A fast ode solver for diffusion probabilistic model sampling in around 10 steps.
\newblock {\em arXiv preprint arXiv:2206.00927}, 2022.

\bibitem{ma2022accelerating}
Hengyuan Ma, Li Zhang, Xiatian Zhu, and Jianfeng Feng.
\newblock Accelerating score-based generative models with preconditioned diffusion sampling.
\newblock In {\em European Conference on Computer Vision}, pages 1--16. Springer, 2022.

\bibitem{mechrez2018maintaining}
Roey Mechrez, Itamar Talmi, Firas Shama, and Lihi Zelnik-Manor.
\newblock Maintaining natural image statistics with the contextual loss.
\newblock In {\em Asian Conference on Computer Vision}, pages 427--443. Springer, 2018.

\bibitem{mechrez2018contextual}
Roey Mechrez, Itamar Talmi, and Lihi Zelnik-Manor.
\newblock The contextual loss for image transformation with non-aligned data.
\newblock In {\em European Conference on Computer Vision (ECCV)}, pages 768--783, 2018.

\bibitem{meng2022distillation}
Chenlin Meng, Ruiqi Gao, Diederik~P Kingma, Stefano Ermon, Jonathan Ho, and Tim Salimans.
\newblock On distillation of guided diffusion models.
\newblock {\em arXiv preprint arXiv:2210.03142}, 2022.

\bibitem{mescheder2018training}
Lars Mescheder, Andreas Geiger, and Sebastian Nowozin.
\newblock Which training methods for gans do actually converge?
\newblock In {\em International conference on machine learning}, pages 3481--3490. PMLR, 2018.

\bibitem{michaeli2014blind}
Tomer Michaeli and Michal Irani.
\newblock Blind deblurring using internal patch recurrence.
\newblock In {\em Computer Vision--ECCV 2014: 13th European Conference, Zurich, Switzerland, September 6-12, 2014, Proceedings, Part III 13}, pages 783--798. Springer, 2014.

\bibitem{mittal2012making}
Anish Mittal, Rajiv Soundararajan, and Alan~C Bovik.
\newblock Making a “completely blind” image quality analyzer.
\newblock {\em IEEE Signal processing letters}, 20(3):209--212, 2012.

\bibitem{Nah_2019_CVPR_Workshops_REDS}
Seungjun Nah, Sungyong Baik, Seokil Hong, Gyeongsik Moon, Sanghyun Son, Radu Timofte, and Kyoung~Mu Lee.
\newblock Ntire 2019 challenge on video deblurring and super-resolution: Dataset and study.
\newblock In {\em CVPR Workshops}, June 2019.

\bibitem{nah2017deep}
Seungjun Nah, Tae~Hyun Kim, and Kyoung~Mu Lee.
\newblock Deep multi-scale convolutional neural network for dynamic scene deblurring.
\newblock In {\em The IEEE Conference on Computer Vision and Pattern Recognition (CVPR)}, July 2017.

\bibitem{nah2022clean}
Seungjun Nah, Sanghyun Son, Jaerin Lee, and Kyoung~Mu Lee.
\newblock Clean images are hard to reblur: Exploiting the ill-posed inverse task for dynamic scene deblurring.
\newblock In {\em International Conference on Learning Representations}, 2022.

\bibitem{nilsson2020understanding}
Jim Nilsson and Tomas Akenine-M{\"o}ller.
\newblock Understanding ssim.
\newblock {\em arXiv preprint arXiv:2006.13846}, 2020.

\bibitem{niu2020single}
Ben Niu, Weilei Wen, Wenqi Ren, Xiangde Zhang, Lianping Yang, Shuzhen Wang, Kaihao Zhang, Xiaochun Cao, and Haifeng Shen.
\newblock Single image super-resolution via a holistic attention network.
\newblock In {\em European conference on computer vision}, pages 191--207. Springer, 2020.

\bibitem{ohayon2021high}
Guy Ohayon, Theo Adrai, Gregory Vaksman, Michael Elad, and Peyman Milanfar.
\newblock High perceptual quality image denoising with a posterior sampling cgan.
\newblock In {\em Proceedings of the IEEE/CVF International Conference on Computer Vision}, pages 1805--1813, 2021.

\bibitem{pan2014deblurring}
Jinshan Pan, Zhe Hu, Zhixun Su, and Ming-Hsuan Yang.
\newblock Deblurring text images via l0-regularized intensity and gradient prior.
\newblock In {\em Proceedings of the IEEE Conference on Computer Vision and Pattern Recognition}, pages 2901--2908, 2014.

\bibitem{pan2016blind}
Jinshan Pan, Deqing Sun, Hanspeter Pfister, and Ming-Hsuan Yang.
\newblock Blind image deblurring using dark channel prior.
\newblock In {\em Proceedings of the IEEE conference on computer vision and pattern recognition}, pages 1628--1636, 2016.

\bibitem{ramesh2022hierarchical}
Aditya Ramesh, Prafulla Dhariwal, Alex Nichol, Casey Chu, and Mark Chen.
\newblock Hierarchical text-conditional image generation with clip latents.
\newblock {\em arXiv preprint arXiv:2204.06125}, 2022.

\bibitem{ren2021segmentation}
Mengwei Ren, Neel Dey, James Fishbaugh, and Guido Gerig.
\newblock Segmentation-renormalized deep feature modulation for unpaired image harmonization.
\newblock {\em IEEE Transactions on Medical Imaging}, 40(6):1519--1530, 2021.

\bibitem{rim2022realistic}
Jaesung Rim, Geonung Kim, Jungeon Kim, Junyong Lee, Seungyong Lee, and Sunghyun Cho.
\newblock Realistic blur synthesis for learning image deblurring.
\newblock In {\em Proceedings of the European Conference on Computer Vision (ECCV)}, 2022.

\bibitem{rim2020real}
Jaesung Rim, Haeyun Lee, Jucheol Won, and Sunghyun Cho.
\newblock Real-world blur dataset for learning and benchmarking deblurring algorithms.
\newblock In {\em European Conference on Computer Vision}, pages 184--201. Springer, 2020.

\bibitem{rombach2022high}
Robin Rombach, Andreas Blattmann, Dominik Lorenz, Patrick Esser, and Bj{\"o}rn Ommer.
\newblock High-resolution image synthesis with latent diffusion models.
\newblock In {\em Proceedings of the IEEE/CVF Conference on Computer Vision and Pattern Recognition}, pages 10684--10695, 2022.

\bibitem{latent_diffusion}
Robin Rombach, Andreas Blattmann, Dominik Lorenz, Patrick Esser, and Bj{\"o}rn Ommer.
\newblock High-resolution image synthesis with latent diffusion models.
\newblock In {\em Proceedings of the IEEE/CVF Conference on Computer Vision and Pattern Recognition}, pages 10684--10695, 2022.

\bibitem{ronneberger2015u}
Olaf Ronneberger, Philipp Fischer, and Thomas Brox.
\newblock U-net: Convolutional networks for biomedical image segmentation.
\newblock In {\em International Conference on Medical image computing and computer-assisted intervention}, pages 234--241. Springer, 2015.

\bibitem{ruiz2022dreambooth}
Nataniel Ruiz, Yuanzhen Li, Varun Jampani, Yael Pritch, Michael Rubinstein, and Kfir Aberman.
\newblock Dreambooth: Fine tuning text-to-image diffusion models for subject-driven generation.
\newblock 2022.

\bibitem{saharia2022palette}
Chitwan Saharia, William Chan, Huiwen Chang, Chris Lee, Jonathan Ho, Tim Salimans, David Fleet, and Mohammad Norouzi.
\newblock Palette: Image-to-image diffusion models.
\newblock In {\em ACM SIGGRAPH 2022 Conference Proceedings}, pages 1--10, 2022.

\bibitem{saharia2022photorealistic}
Chitwan Saharia, William Chan, Saurabh Saxena, Lala Li, Jay Whang, Emily Denton, Seyed Kamyar~Seyed Ghasemipour, Burcu~Karagol Ayan, S~Sara Mahdavi, Rapha~Gontijo Lopes, et~al.
\newblock Photorealistic text-to-image diffusion models with deep language understanding.
\newblock {\em arXiv preprint arXiv:2205.11487}, 2022.

\bibitem{saharia2022image}
Chitwan Saharia, Jonathan Ho, William Chan, Tim Salimans, David~J Fleet, and Mohammad Norouzi.
\newblock Image super-resolution via iterative refinement.
\newblock {\em IEEE Transactions on Pattern Analysis and Machine Intelligence}, 2022.

\bibitem{shao2020domain}
Yuanjie Shao, Lerenhan Li, Wenqi Ren, Changxin Gao, and Nong Sang.
\newblock Domain adaptation for image dehazing.
\newblock In {\em Proceedings of the IEEE/CVF conference on computer vision and pattern recognition}, pages 2808--2817, 2020.

\bibitem{shen2019human}
Ziyi Shen, Wenguan Wang, Jianbing Shen, Haibin Ling, Tingfa Xu, and Ling Shao.
\newblock Human-aware motion deblurring.
\newblock In {\em IEEE International Conference on Computer Vision}, 2019.

\bibitem{sohl_thermodynamics}
Jascha Sohl-Dickstein, Eric Weiss, Niru Maheswaranathan, and Surya Ganguli.
\newblock Deep unsupervised learning using nonequilibrium thermodynamics.
\newblock In {\em International Conference on Machine Learning}, pages 2256--2265. PMLR, 2015.

\bibitem{somepalli2022diffusion}
Gowthami Somepalli, Vasu Singla, Micah Goldblum, Jonas Geiping, and Tom Goldstein.
\newblock Diffusion art or digital forgery? investigating data replication in diffusion models.
\newblock {\em arXiv preprint arXiv:2212.03860}, 2022.

\bibitem{ddim}
Jiaming Song, Chenlin Meng, and Stefano Ermon.
\newblock Denoising diffusion implicit models.
\newblock In {\em International Conference on Learning Representations}, 2021.

\bibitem{song2021maximum}
Yang Song, Conor Durkan, Iain Murray, and Stefano Ermon.
\newblock Maximum likelihood training of score-based diffusion models.
\newblock {\em Advances in Neural Information Processing Systems}, 34:1415--1428, 2021.

\bibitem{ncsn}
Yang Song and Stefano Ermon.
\newblock Generative modeling by estimating gradients of the data distribution.
\newblock {\em Advances in Neural Information Processing Systems}, 32, 2019.

\bibitem{ncsnv2}
Yang Song and Stefano Ermon.
\newblock Improved techniques for training score-based generative models.
\newblock {\em Advances in neural information processing systems}, 33:12438--12448, 2020.

\bibitem{ncsnv3}
Yang Song, Jascha Sohl-Dickstein, Diederik~P Kingma, Abhishek Kumar, Stefano Ermon, and Ben Poole.
\newblock Score-based generative modeling through stochastic differential equations.
\newblock In {\em International Conference on Learning Representations}, 2021.

\bibitem{suin2020spatially}
Maitreya Suin, Kuldeep Purohit, and A.~N. Rajagopalan.
\newblock Spatially-attentive patch-hierarchical network for adaptive motion deblurring.
\newblock In {\em Proceedings of the IEEE/CVF Conference on Computer Vision and Pattern Recognition (CVPR)}, June 2020.

\bibitem{tsai2022stripformer}
Fu-Jen Tsai, Yan-Tsung Peng, Yen-Yu Lin, Chung-Chi Tsai, and Chia-Wen Lin.
\newblock Stripformer: Strip transformer for fast image deblurring.
\newblock In {\em Proceedings of the European Conference on Computer Vision (ECCV)}, 2022.

\bibitem{tu2022maxim}
Zhengzhong Tu, Hossein Talebi, Han Zhang, Feng Yang, Peyman Milanfar, Alan Bovik, and Yinxiao Li.
\newblock Maxim: Multi-axis mlp for image processing.
\newblock In {\em Proceedings of the IEEE/CVF Conference on Computer Vision and Pattern Recognition}, pages 5769--5780, 2022.

\bibitem{wang_unsupervised_sr}
Wei Wang, Haochen Zhang, Zehuan Yuan, and Changhu Wang.
\newblock Unsupervised real-world super-resolution: A domain adaptation perspective.
\newblock In {\em 2021 IEEE/CVF International Conference on Computer Vision (ICCV)}, pages 4298--4307, 2021.

\bibitem{wang2018recovering}
Xintao Wang, Ke Yu, Chao Dong, and Chen~Change Loy.
\newblock Recovering realistic texture in image super-resolution by deep spatial feature transform.
\newblock In {\em Proceedings of the IEEE conference on computer vision and pattern recognition}, pages 606--615, 2018.

\bibitem{Wang_2022_CVPR_uformer}
Zhendong Wang, Xiaodong Cun, Jianmin Bao, Wengang Zhou, Jianzhuang Liu, and Houqiang Li.
\newblock Uformer: A general u-shaped transformer for image restoration.
\newblock In {\em Proceedings of the IEEE/CVF Conference on Computer Vision and Pattern Recognition (CVPR)}, pages 17683--17693, June 2022.

\bibitem{whang2022deblurring}
Jay Whang, Mauricio Delbracio, Hossein Talebi, Chitwan Saharia, Alexandros~G Dimakis, and Peyman Milanfar.
\newblock Deblurring via stochastic refinement.
\newblock In {\em Proceedings of the IEEE/CVF Conference on Computer Vision and Pattern Recognition}, pages 16293--16303, 2022.

\bibitem{wolf2021deflow}
Valentin Wolf, Andreas Lugmayr, Martin Danelljan, Luc Van~Gool, and Radu Timofte.
\newblock Deflow: Learning complex image degradations from unpaired data with conditional flows.
\newblock In {\em Proceedings of the IEEE/CVF Conference on Computer Vision and Pattern Recognition}, pages 94--103, 2021.

\bibitem{xiao2022DDGAN}
Zhisheng Xiao, Karsten Kreis, and Arash Vahdat.
\newblock Tackling the generative learning trilemma with denoising diffusion {GAN}s.
\newblock In {\em International Conference on Learning Representations (ICLR)}, 2022.

\bibitem{xu2013unnatural}
Li Xu, Shicheng Zheng, and Jiaya Jia.
\newblock Unnatural l0 sparse representation for natural image deblurring.
\newblock In {\em Proceedings of the IEEE conference on computer vision and pattern recognition}, pages 1107--1114, 2013.

\bibitem{Zamir2021Restormer}
Syed~Waqas Zamir, Aditya Arora, Salman Khan, Munawar Hayat, Fahad~Shahbaz Khan, and Ming-Hsuan Yang.
\newblock Restormer: Efficient transformer for high-resolution image restoration.
\newblock In {\em Proceedings of the IEEE/CVF Conference on Computer Vision and Pattern Recognition}, 2022.

\bibitem{zamir2020learning}
Syed~Waqas Zamir, Aditya Arora, Salman Khan, Munawar Hayat, Fahad~Shahbaz Khan, Ming-Hsuan Yang, and Ling Shao.
\newblock Learning enriched features for real image restoration and enhancement.
\newblock In {\em European Conference on Computer Vision}, pages 492--511. Springer, 2020.

\bibitem{Zamir2021MPRNet}
Syed~Waqas Zamir, Aditya Arora, Salman Khan, Munawar Hayat, Fahad~Shahbaz Khan, Ming-Hsuan Yang, and Ling Shao.
\newblock Multi-stage progressive image restoration.
\newblock In {\em Proceedings of the IEEE/CVF Conference on Computer Vision and Pattern Recognition}, 2021.

\bibitem{zhang2019deep}
Hongguang Zhang, Yuchao Dai, Hongdong Li, and Piotr Koniusz.
\newblock Deep stacked hierarchical multi-patch network for image deblurring.
\newblock In {\em Proceedings of the IEEE/CVF Conference on Computer Vision and Pattern Recognition}, pages 5978--5986, 2019.

\bibitem{zhang2018dynamic}
Jiawei Zhang, Jinshan Pan, Jimmy Ren, Yibing Song, Linchao Bao, Rynson~WH Lau, and Ming-Hsuan Yang.
\newblock Dynamic scene deblurring using spatially variant recurrent neural networks.
\newblock In {\em Proceedings of the IEEE Conference on Computer Vision and Pattern Recognition}, pages 2521--2529, 2018.

\bibitem{zhang2020deblurring}
Kaihao Zhang, Wenhan Luo, Yiran Zhong, Lin Ma, Bjorn Stenger, Wei Liu, and Hongdong Li.
\newblock Deblurring by realistic blurring.
\newblock In {\em Proceedings of the IEEE/CVF Conference on Computer Vision and Pattern Recognition}, pages 2737--2746, 2020.

\bibitem{zhang2023adding}
Lvmin Zhang and Maneesh Agrawala.
\newblock Adding conditional control to text-to-image diffusion models.
\newblock {\em arXiv preprint arXiv:2302.05543}, 2023.

\bibitem{zhang2018unreasonable}
Richard Zhang, Phillip Isola, Alexei~A Efros, Eli Shechtman, and Oliver Wang.
\newblock The unreasonable effectiveness of deep features as a perceptual metric.
\newblock In {\em IEEE conference on computer vision and pattern recognition}, pages 586--595, 2018.

\bibitem{zhang2018perceptual}
Richard Zhang, Phillip Isola, Alexei~A. Efros, Eli Shechtman, and Oliver Wang.
\newblock The unreasonable effectiveness of deep features as a perceptual metric.
\newblock In {\em Proceedings of the IEEE Conference on Computer Vision and Pattern Recognition (CVPR)}, June 2018.

\bibitem{zhang2019zoom}
Xuaner Zhang, Qifeng Chen, Ren Ng, and Vladlen Koltun.
\newblock Zoom to learn, learn to zoom.
\newblock In {\em IEEE Conference on Computer Vision and Pattern Recognition}, pages 3762--3770, 2019.

\bibitem{zhou2022lednet}
Shangchen Zhou, Chongyi Li, and Chen~Change Loy.
\newblock Lednet: Joint low-light enhancement and deblurring in the dark.
\newblock {\em arXiv preprint arXiv:2202.03373}, 2022.

\bibitem{zhou2019spatio}
Shangchen Zhou, Jiawei Zhang, Jinshan Pan, Haozhe Xie, Wangmeng Zuo, and Jimmy Ren.
\newblock Spatio-temporal filter adaptive network for video deblurring.
\newblock In {\em Proceedings of the IEEE/CVF International Conference on Computer Vision}, pages 2482--2491, 2019.

\end{thebibliography}
}

\newpage
\appendix

\clearpage
\onecolumn
\begin{center}
{\Large \bf Appendix: Multiscale Structure Guided Diffusion for Image Deblurring}
\end{center}
\section{Additional Results}
\subsection{Effectiveness of the guidance on GoPro, HIDE and REDS}
\begin{figure*}[!h]
    \centering
    \includegraphics[width=\linewidth]{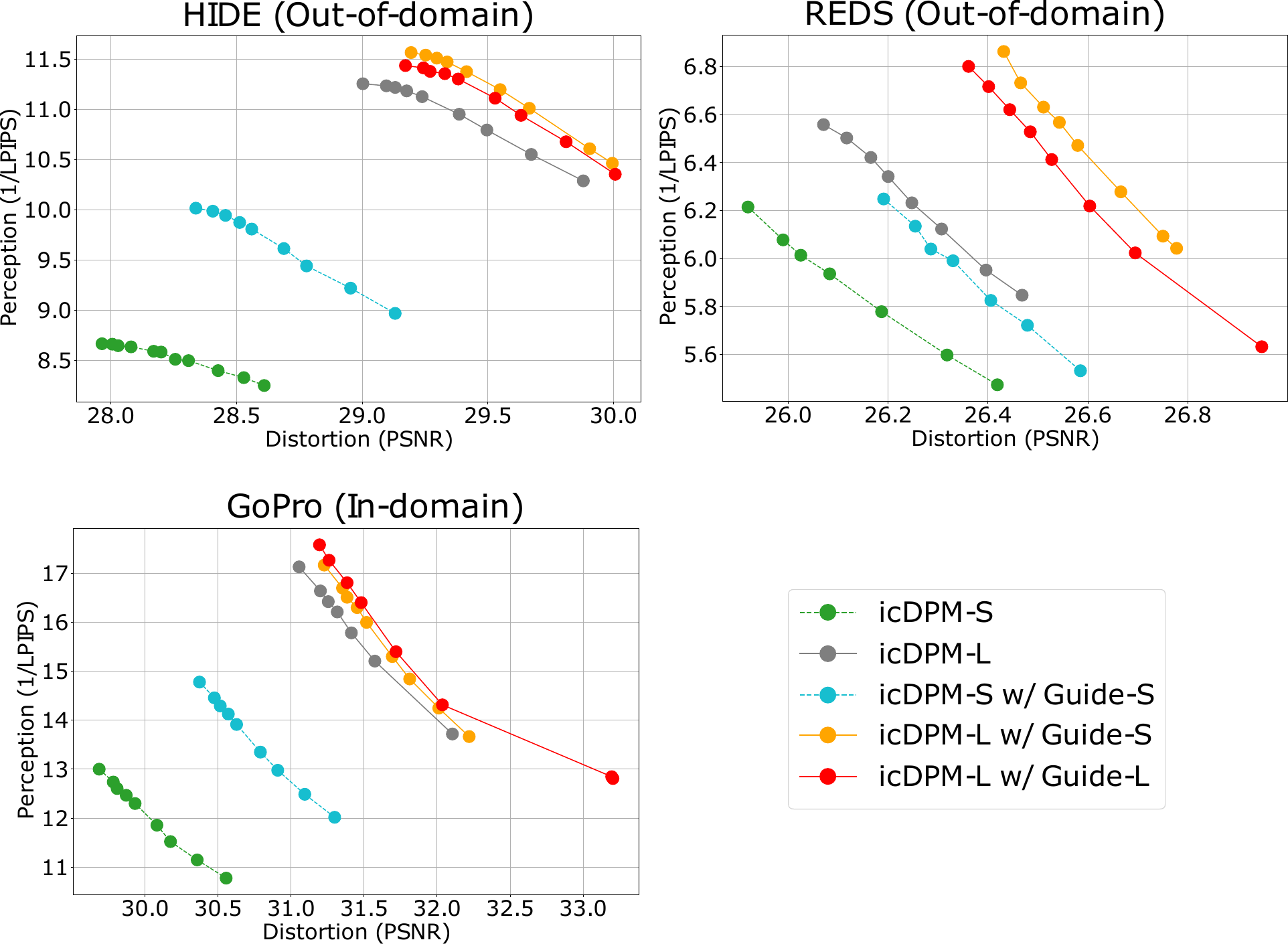}
    \caption{Additional perception-distortion plots as supplementary for Sec. 4.3 in the main paper. All models are trained only on GoPro~\cite{nah2017deep}.
    The guidance mechanism allows for consistent better perceptual qualities and lower distortions compared to image-conditioned diffusion probablistic model (icDPM) under different network capacities ('icDPM-S w/ Guide-S' $>$ `icDPM-S'; `icDPM-L w/ Guide-S/L' $>$ `icDPM-L'), both in-domain (GoPro) and out-of-domain (HIDE, REDS)
    `-S' and `-L' refer to small and large networks respectively. 
    }
    \label{fig:more_pd_curve}
\end{figure*}

We include additional perception-distortion plots for GoPro~\cite{nah2017deep}, HIDE~\cite{shen2019human} and REDS~\cite{Nah_2019_CVPR_Workshops_REDS} datasets in Fig.~\ref{fig:more_pd_curve}, as supplementary for Section 4.3 of the main paper, to verify the effectiveness of the proposed guidance.

\subsection{Additional visual results}
To supplement main paper Fig.6,7,8, we provide additional and enlarged qualitative results for all datasets below.

\textbf{Realblur-J} (out-of-domain) deblurring examples are shown in Fig.~\ref{fig:real1},~\ref{fig:real2},~\ref{fig:real3},~\ref{fig:real4}.

\textbf{REDS} (out-of-domain) deblurring examples are shown in Fig.~\ref{fig:reds1},~\ref{fig:reds2},~\ref{fig:reds3},~\ref{fig:reds4}.

\textbf{HIDE} (out-of-domain) deblurring examples are shown in 
Fig.~\ref{fig:hide1},~\ref{fig:hide2},~\ref{fig:hide3},~\ref{fig:hide4}.

\textbf{GoPro} (in-domain) deblurring examples are shown in Fig.~\ref{fig:gopro1},~\ref{fig:gopro2},~\ref{fig:gopro3},~\ref{fig:gopro4}.

\vspace{2em}
\subsection{Failure cases}
As discussed in the main paper, we acknowledge that the domain generalization of the model is still extensively bounded by the quality of the training set. In our experiments, we only train with GoPro~\cite{nah2017deep} for the sake of benchmarking. However, the data diversity and representativity from GoPro is limited, i.e., it only contains daytime scenes, acquired outdoor under sufficient lighting conditions. Moreover, the synthesis of blur in GoPro by simple averaging of consecutive frames is less realistic~\cite{zhou2022lednet}. Lastly, the ground truth images in GoPro dataset are rather low-quality, which may further hurt the out-of-domain performance.
Therefore, it is expected that it will be extremely hard for the model to perform decent deblurring on scenes significantly different from GoPro, such as low-light images with saturated regions, in Realblur-J~\cite{rim2020real}.

We include a few failure cases on such scenes in Fig.~\ref{fig:failure1},~\ref{fig:failure2},~\ref{fig:failure3}~\ref{fig:failure4}, where \textbf{all} methods fail to remove blur from the night scenes, especially with night streaks. We believe that in practice, more realistic training datasets~\cite{rim2022realistic} will further increase the model generalization.

\vspace{2em}

\section{Additional Ablation}

\textbf{Input concatenation}
During prototyping, we also explored the possibility of removing input-level concatenation, and only rely on the intermediate representations from regression as the condition of the diffusion model, similar as in~\cite{li2022srdiff} for super-resolution. 
Potentially, we expect such setting will further make the model domain-generalizable as it does not directly interact with images from different domains, although it may also risk losing detailed information from the input.

As proof of concept, we use the same multiscale regression networks, and compare the models with or without input concatenation. Further, since the diffusion model now only takes the intermediate representations as input, we reintroduce the RGB information by using our model variants (d) in Table 8. in the main paper (i.e., regression targets are downsampled RGB images instead of grayscale images). From Table~\ref{table:input_concatenate}, we observe that the input concatenation obtained a much better performance both in-domain and out-of-domain than without concatenation.
Therefore, in our final model, we keep the input concatenation and only rely on the guidance features to provide additional information.
\begin{table}[!h]
\centering
\caption{Effects of input-level concatenation. From our model variant (d) in Table 8. of the main paper, we remove the input concatenation, loosely inspired by ~\cite{li2022srdiff} (super-resolution). In the context of deblurring, we observe deteriorate results indicated in row `w/o input concatenation', compared to the setting with additional input concatenation.}
    \label{table:input_concatenate}
\begin{tabular}{lcccc}
\toprule
\multirow{2}{*}{}       & \multicolumn{2}{c}{In-domain} & \multicolumn{2}{c}{Out-of-domain} \\ \cmidrule(lr){2-5} 
                        & PSNR $\uparrow$  & LPIPS $\downarrow$       & PSNR $\uparrow$ & LPIPS $\downarrow$           \\ \midrule
w/o input concatenation & 25.20         & 0.230         & 28.29           & 0.177           \\
w/ concatenation        & 30.65         & 0.090         & 28.45           & 0.141 \\ \bottomrule
\end{tabular}
\end{table}

\begin{figure*}[!h]
    \centering
    \includegraphics[width=0.7\linewidth]{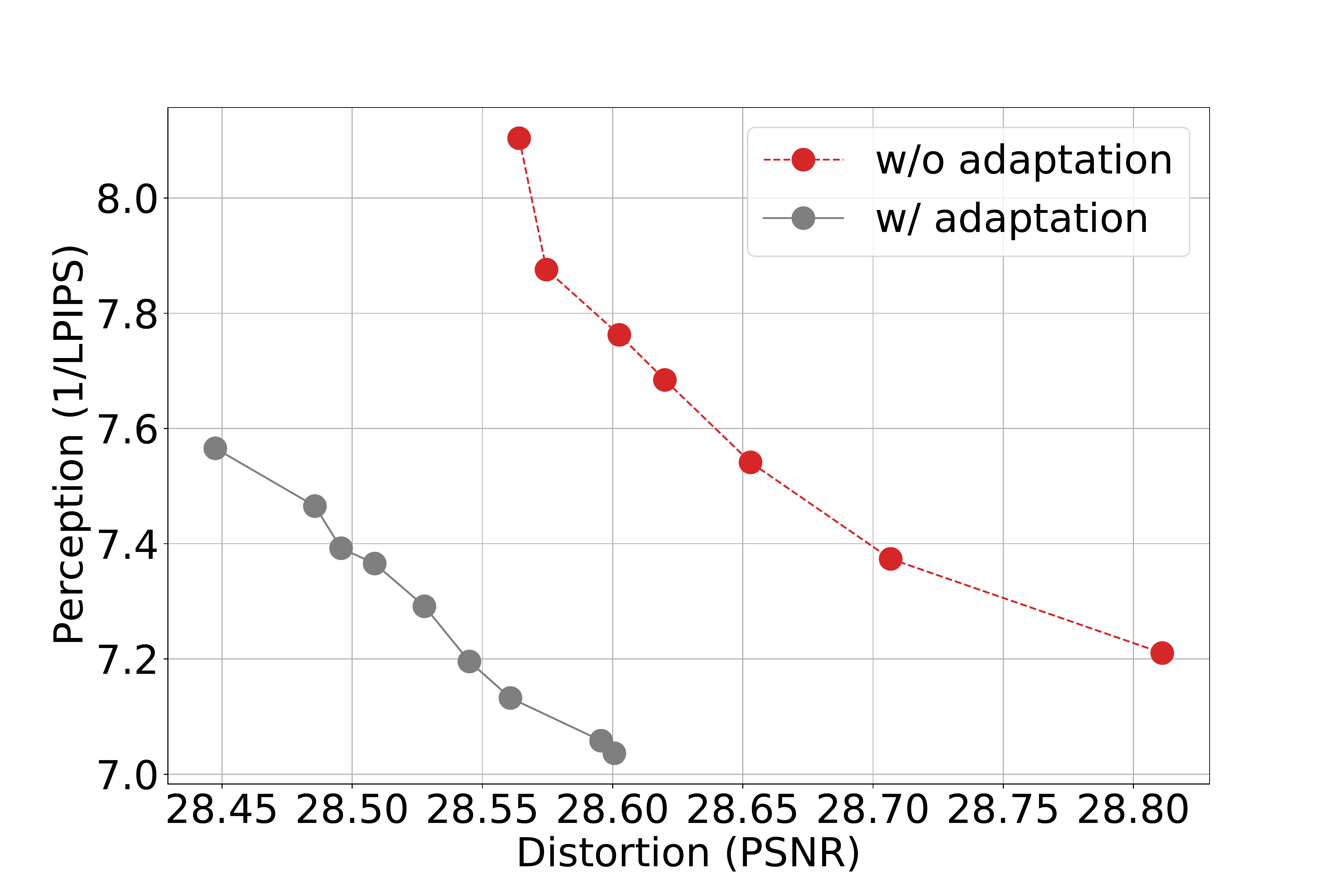}
    \caption{A comparison between our models with or without further domain adaptation with Realblur-J, on a GoPro trained model. Surprisingly, further adversarial domain adaptation on the guidance features between GoPro and Realblur-J hurt the performance. }
    \label{fig:adaptation}
\end{figure*}

\clearpage
\textbf{Further cross-domain alignment.} We also explored the potential effects of finetuning the DPMs with adversarial formulation where we used additional discriminators on the guidance features between different datasets (e.g., GoPro and Realblur-J) so that the features extracted from different domains become indistinguishable, similar to the feature alignment strategy in ~\cite{wang_unsupervised_sr}. However, we do not observe extra benefits, and find that such finetuning may even hurt the performance as shown in Fig.~\ref{fig:adaptation}. We speculate that it could be a result of training instability of GANs, or perhaps the suboptimal formulation under the image-conditioned DPM framework. We will leave this for future investigation.

\clearpage
\section{Additional implementation details}
\subsection{Architectures}
\begin{figure*}[!h]
    \centering
    \includegraphics[width=\linewidth]{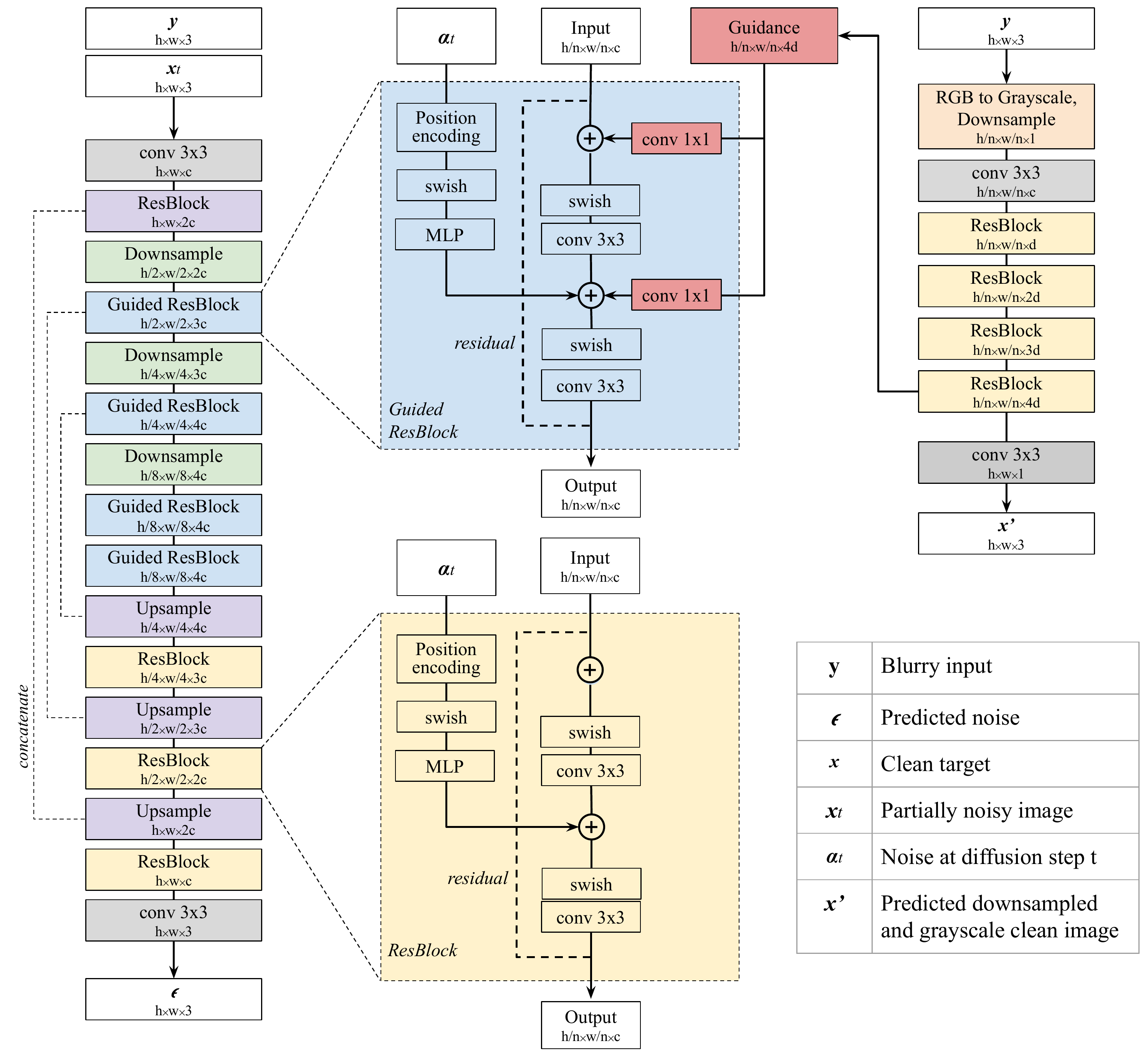}
    \caption{The detailed architecture of the proposed method. \textbf{Left}: the image-conditioned diffusion network based on a fully-convolutional UNet similar to~\cite{whang2022deblurring}, where we replace the residual blocks from the UNet encoder with the proposed guided residual block. \textbf{Middle} column illustrates the difference between a standard residual block and the proposed guided block, where we additionally incorporate  multiscale structure guidance. \textbf{Right}: The proposed guidance network for extracting the coarse structure features from the input at multiple resolutions. At each scale, the blurry image is first converted to grayscale, downsampled, and lastly fed into the network 
    to predict its clean counterpart. The output from the last residual block is leveraged as the guidance feature.
    }
    \label{fig:detailed_arch}
\end{figure*}

\begin{figure*}[t]
    \centering
    \includegraphics[width=0.95\linewidth]{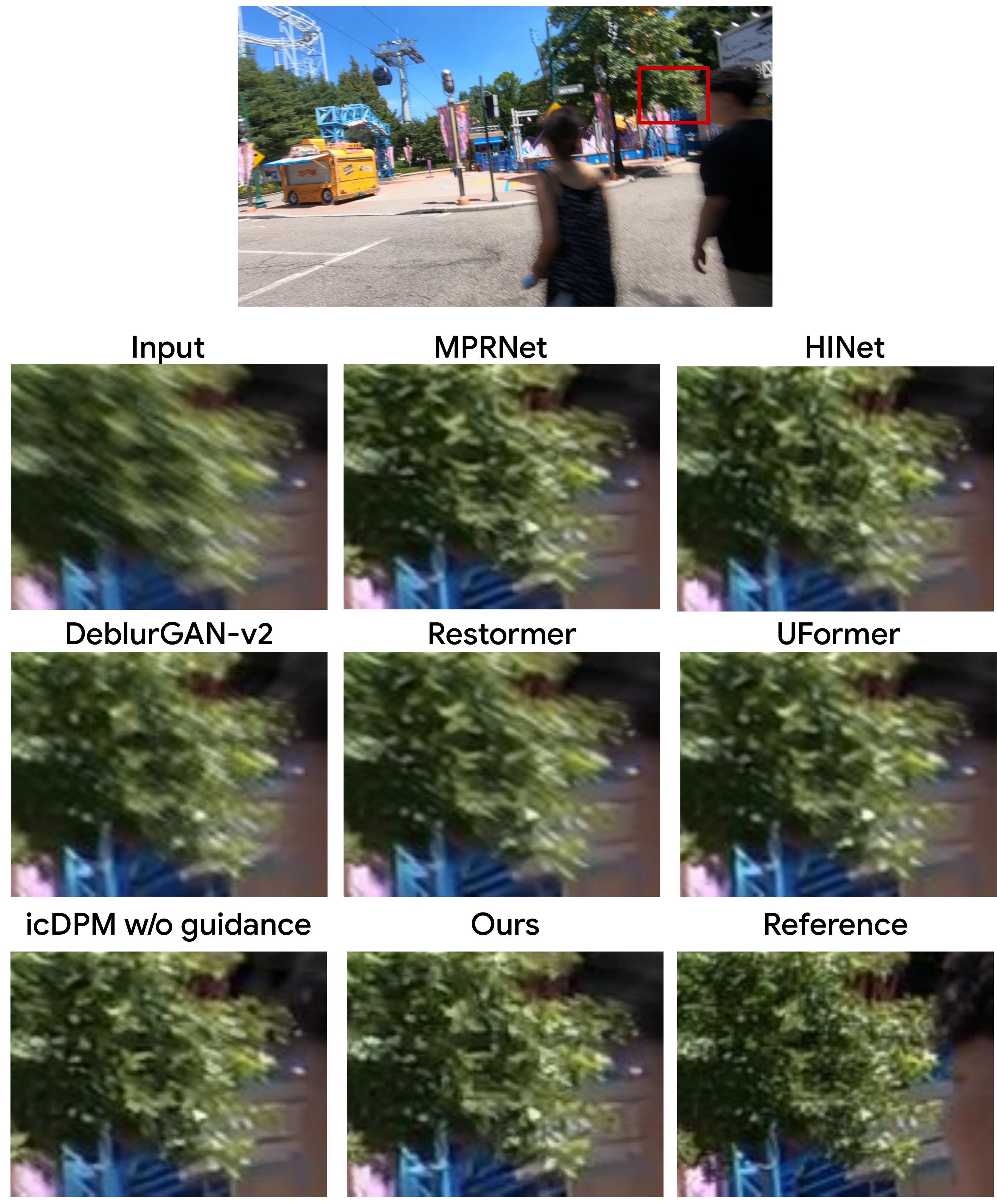}
    \caption{\textbf{REDS}~\cite{Nah_2019_CVPR_Workshops_REDS} deblurring examples from MPRNet~\cite{Zamir2021MPRNet}, HINet~\cite{Chen_2021_CVPR_HINet}, DeblurGAN-v2~\cite{Kupyn_2019_ICCV_Deblurganv2}, Restormer~\cite{Zamir2021Restormer}, UFormer~\cite{Wang_2022_CVPR_uformer}, icDPM without guidance and Ours (icDPM with guidance).}    
    \label{fig:reds1}
\end{figure*}

\begin{figure*}[t]
    \centering
    \includegraphics[width=\linewidth]{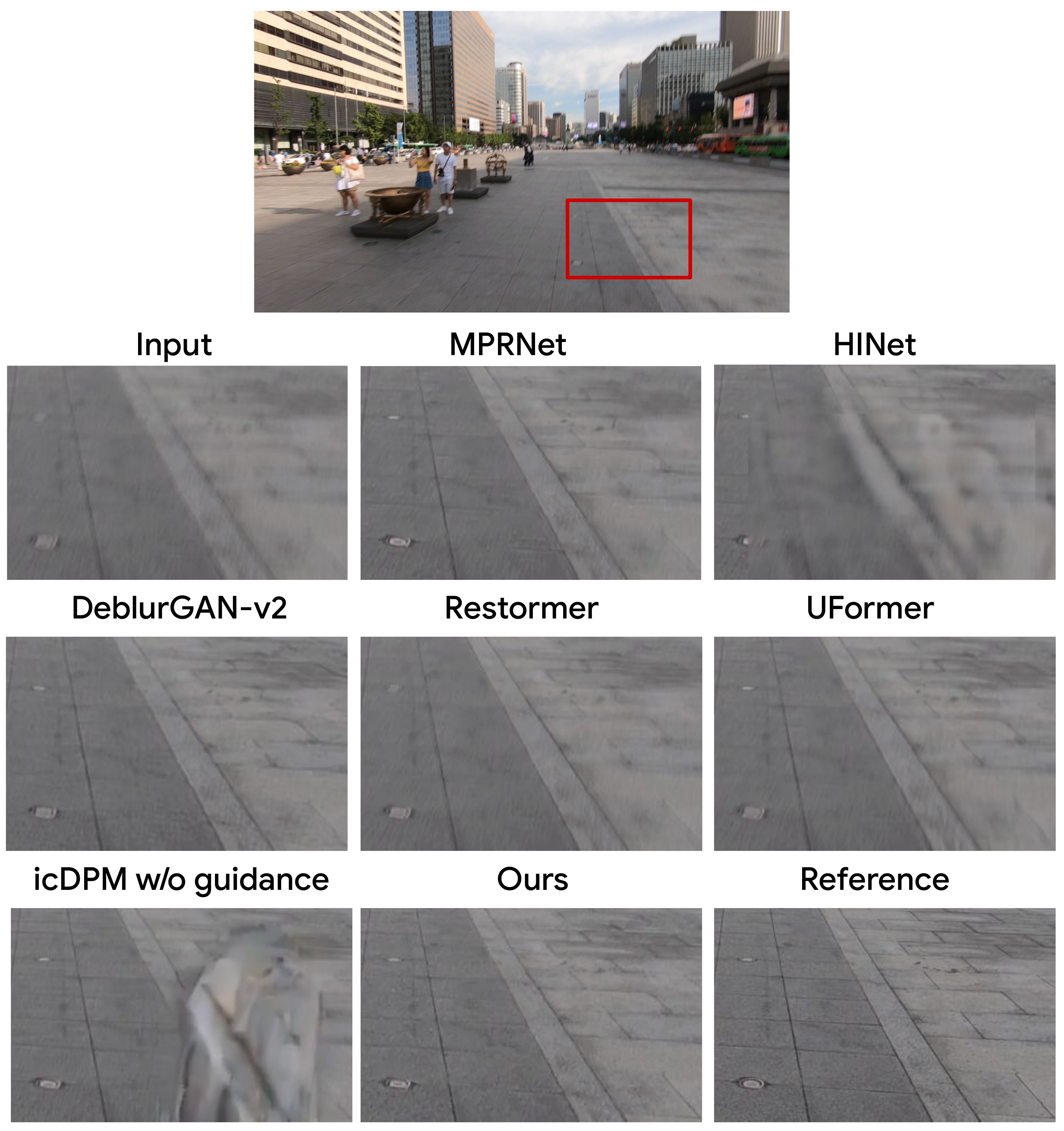}
    \caption{\textbf{REDS}~\cite{Nah_2019_CVPR_Workshops_REDS} deblurring examples from MPRNet~\cite{Zamir2021MPRNet}, HINet~\cite{Chen_2021_CVPR_HINet}, DeblurGAN-v2~\cite{Kupyn_2019_ICCV_Deblurganv2}, Restormer~\cite{Zamir2021Restormer}, UFormer~\cite{Wang_2022_CVPR_uformer}, icDPM without guidance and Ours (icDPM with guidance).}  
    \label{fig:reds2}
\end{figure*}

\begin{figure*}[t]
    \centering
    \includegraphics[width=\linewidth]{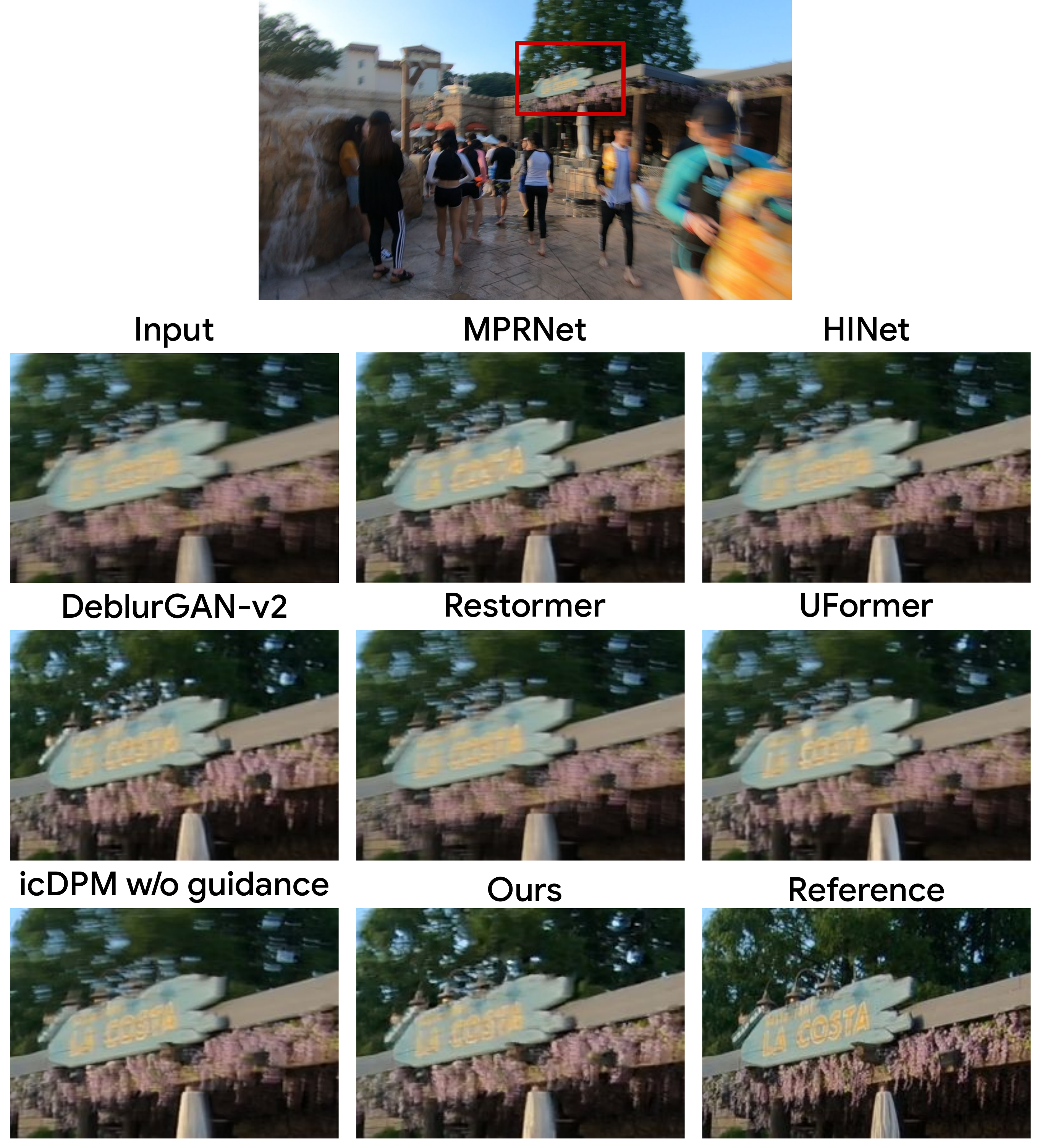}
    \caption{\textbf{REDS}~\cite{Nah_2019_CVPR_Workshops_REDS} deblurring examples from MPRNet~\cite{Zamir2021MPRNet}, HINet~\cite{Chen_2021_CVPR_HINet}, DeblurGAN-v2~\cite{Kupyn_2019_ICCV_Deblurganv2}, Restormer~\cite{Zamir2021Restormer}, UFormer~\cite{Wang_2022_CVPR_uformer}, icDPM without guidance and Ours (icDPM with guidance).}        
    \label{fig:reds3}
\end{figure*}

\begin{figure*}[t]
    \centering
    \includegraphics[width=\linewidth]{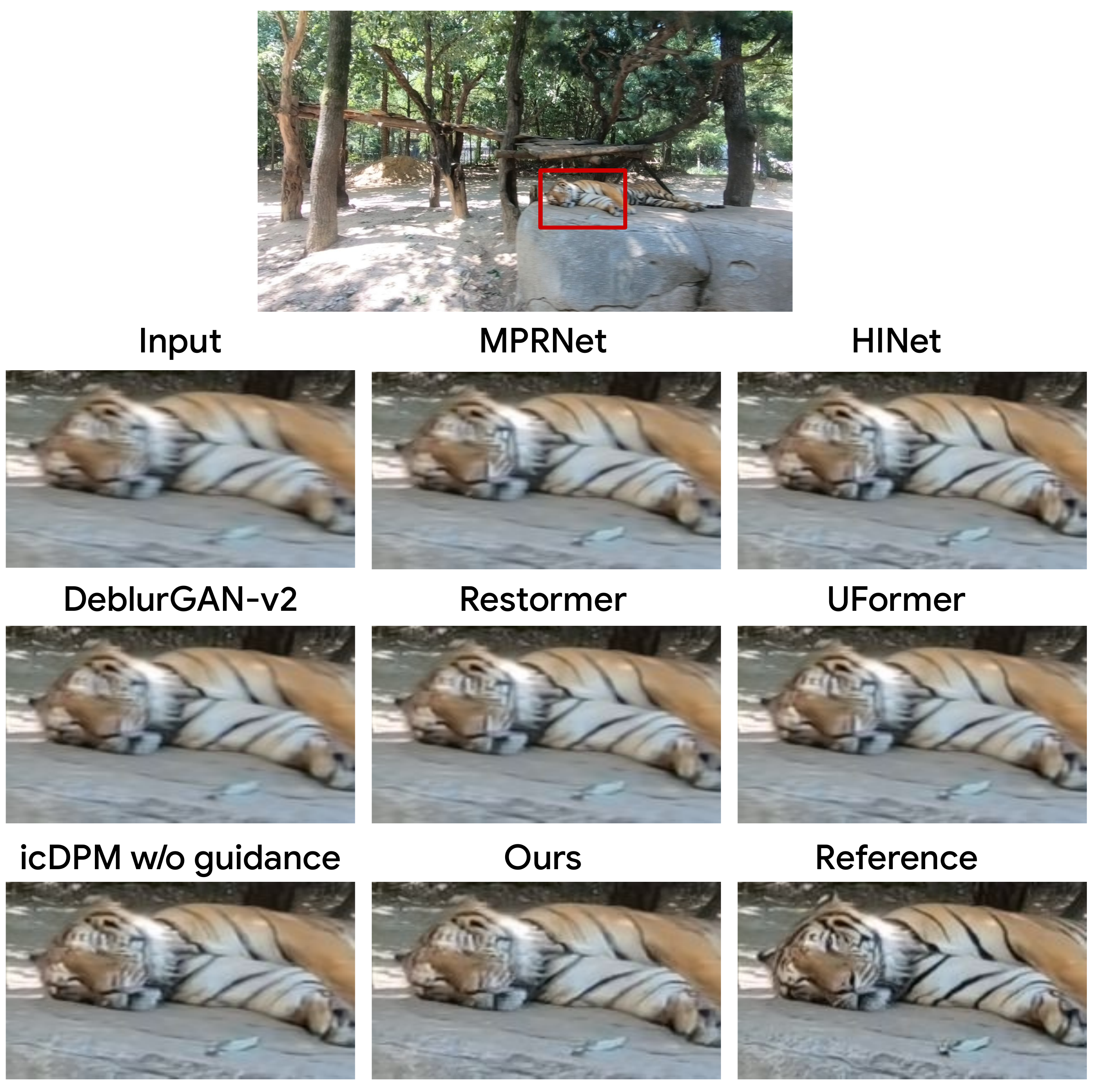}
    \caption{\textbf{REDS}~\cite{Nah_2019_CVPR_Workshops_REDS} deblurring examples from MPRNet~\cite{Zamir2021MPRNet}, HINet~\cite{Chen_2021_CVPR_HINet}, DeblurGAN-v2~\cite{Kupyn_2019_ICCV_Deblurganv2}, Restormer~\cite{Zamir2021Restormer}, UFormer~\cite{Wang_2022_CVPR_uformer}, icDPM without guidance and Ours (icDPM with guidance).}
    \label{fig:reds4}
\end{figure*}

\begin{figure*}[t]
    \centering
    \includegraphics[width=0.88\linewidth]{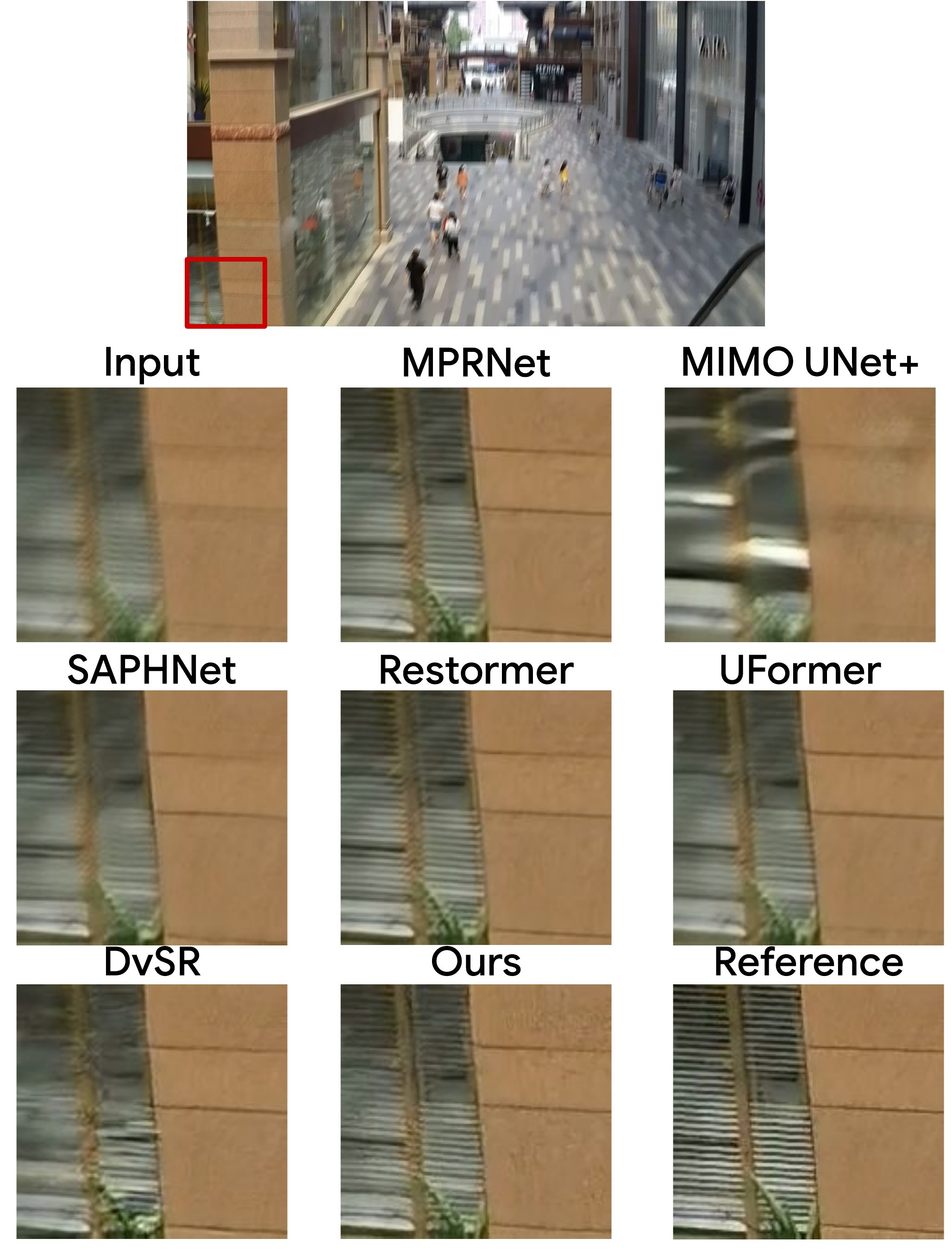}
    \caption{\textbf{HIDE}~\cite{shen2019human} deblurring examples from MPRNet~\cite{Zamir2021MPRNet}, MIMO UNet+~\cite{cho2021rethinking_mimounet}, SAPHNet~\cite{suin2020spatially}, Restormer~\cite{Zamir2021Restormer}, UFormer~\cite{Wang_2022_CVPR_uformer}, DvSR~\cite{whang2022deblurring} and Ours.    
    }
    \label{fig:hide1}
\end{figure*}

\begin{figure*}[t]
    \centering
    \includegraphics[width=0.85\linewidth]{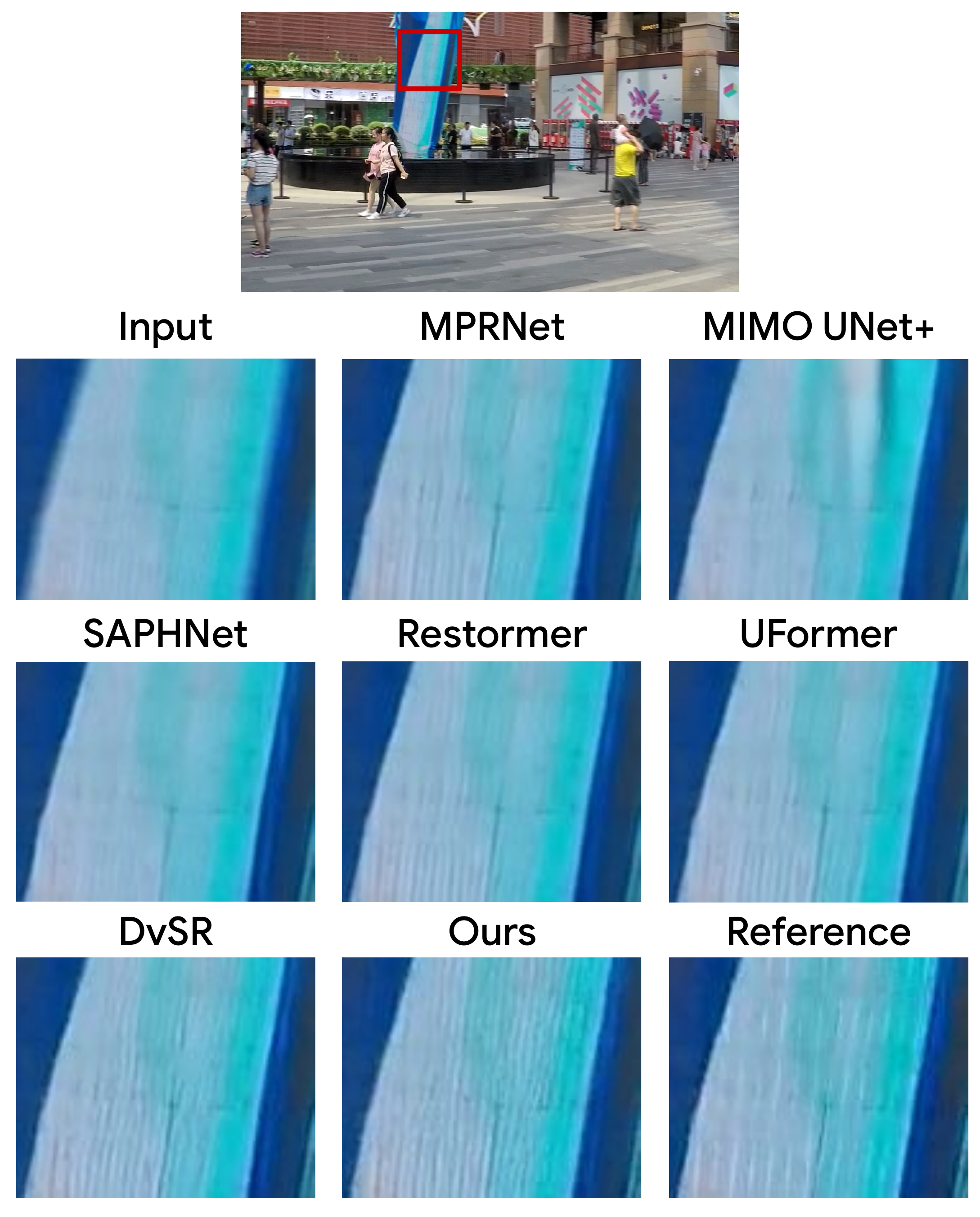}
    \caption{\textbf{HIDE}~\cite{shen2019human} deblurring examples from MPRNet~\cite{Zamir2021MPRNet}, MIMO UNet+~\cite{cho2021rethinking_mimounet}, SAPHNet~\cite{suin2020spatially}, Restormer~\cite{Zamir2021Restormer}, UFormer~\cite{Wang_2022_CVPR_uformer}, DvSR~\cite{whang2022deblurring} and Ours.    
    }    \label{fig:hide2}
\end{figure*}

\begin{figure*}[t]
    \centering
    \includegraphics[width=0.9\linewidth]{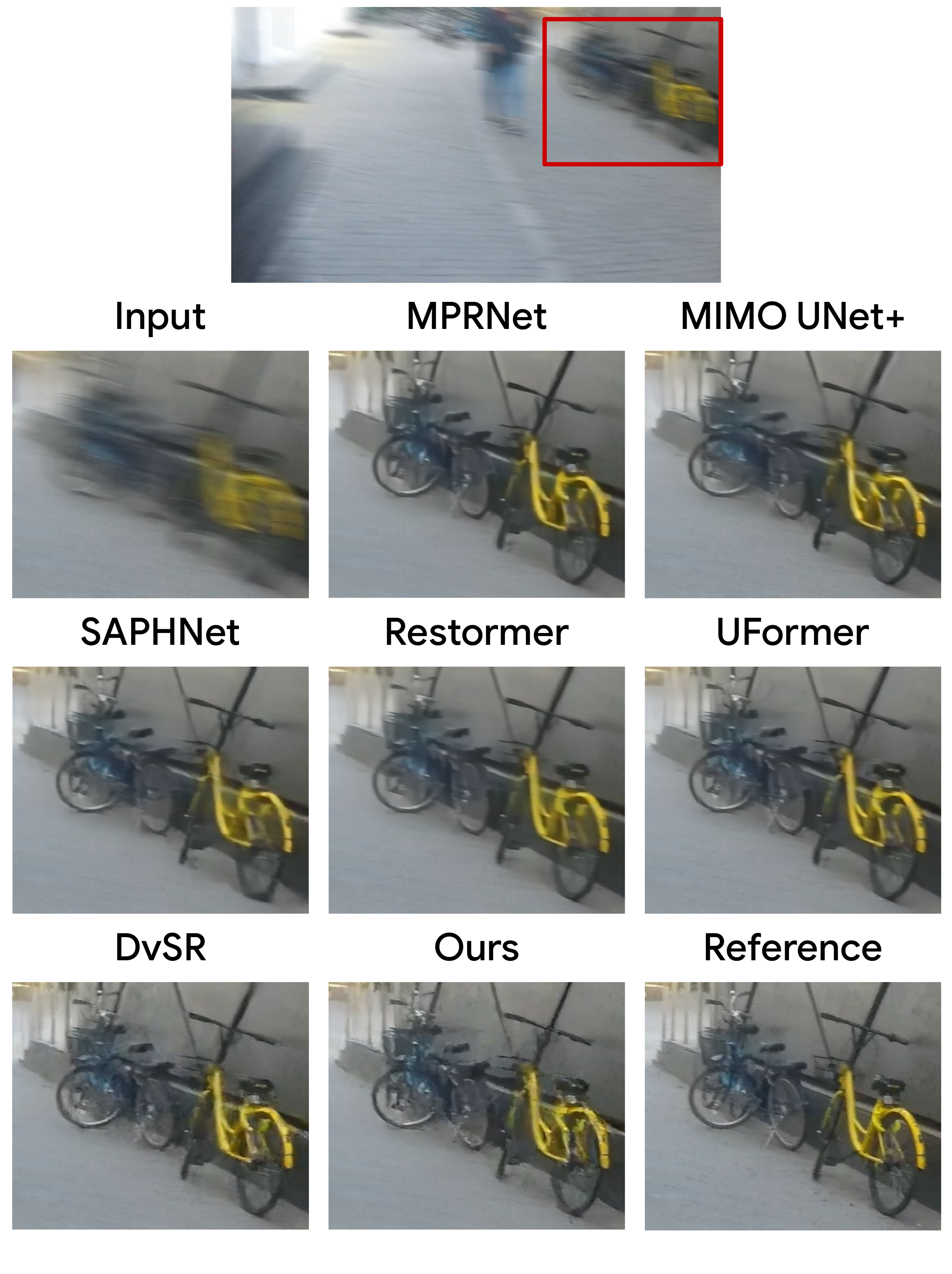}
    \caption{\textbf{HIDE}~\cite{shen2019human} deblurring examples from MPRNet~\cite{Zamir2021MPRNet}, MIMO UNet+~\cite{cho2021rethinking_mimounet}, SAPHNet~\cite{suin2020spatially}, Restormer~\cite{Zamir2021Restormer}, UFormer~\cite{Wang_2022_CVPR_uformer}, DvSR~\cite{whang2022deblurring} and Ours.    
    }
    \label{fig:hide3}
\end{figure*}

\begin{figure*}[t]
    \centering
    \includegraphics[width=0.8\linewidth]{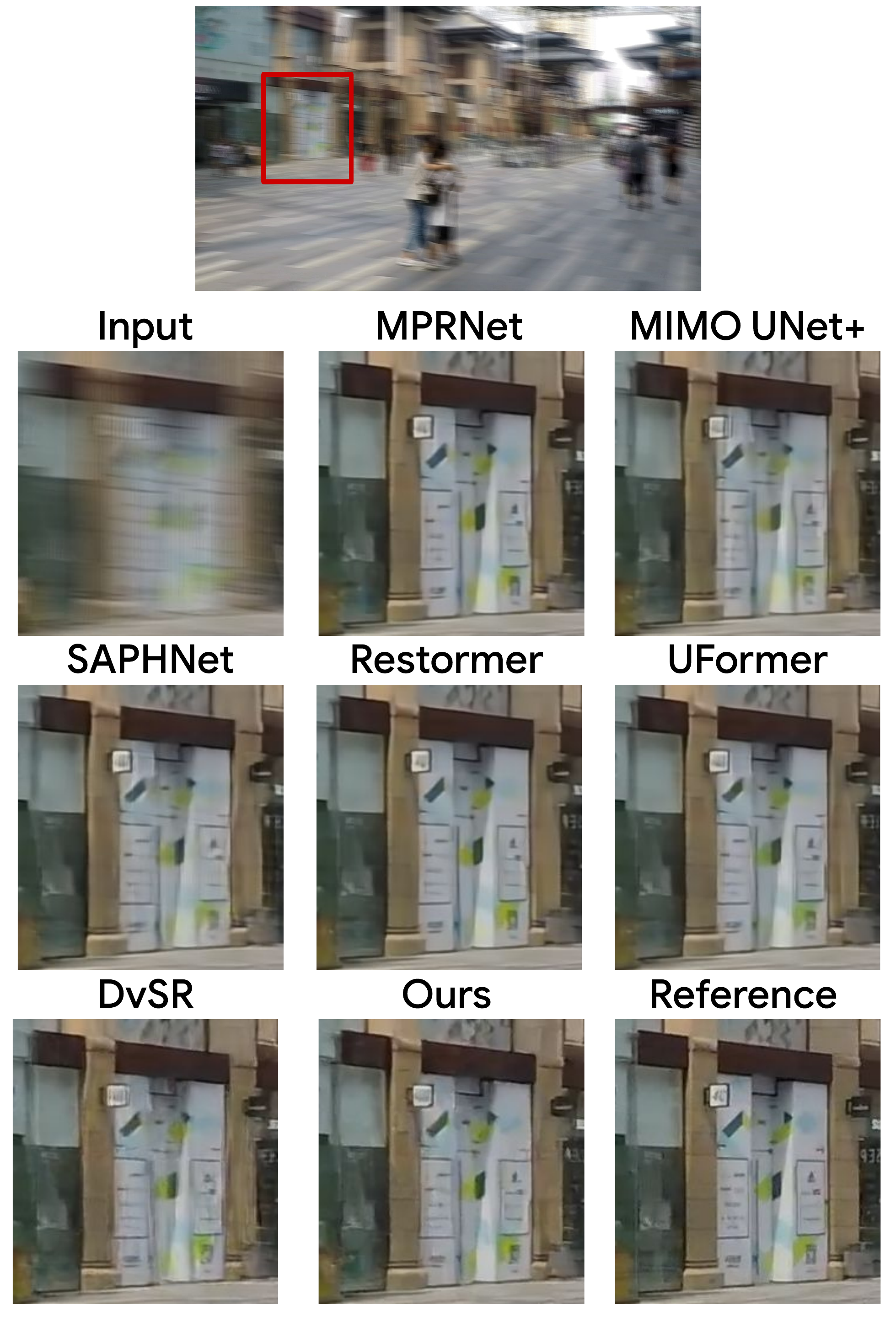}
    \caption{\textbf{HIDE}~\cite{shen2019human} deblurring examples from MPRNet~\cite{Zamir2021MPRNet}, MIMO UNet+~\cite{cho2021rethinking_mimounet}, SAPHNet~\cite{suin2020spatially}, Restormer~\cite{Zamir2021Restormer}, UFormer~\cite{Wang_2022_CVPR_uformer}, DvSR~\cite{whang2022deblurring} and Ours.    
    }
    \label{fig:hide4}
\end{figure*}

\begin{figure*}[h]
    \centering
    \includegraphics[width=0.88\linewidth]{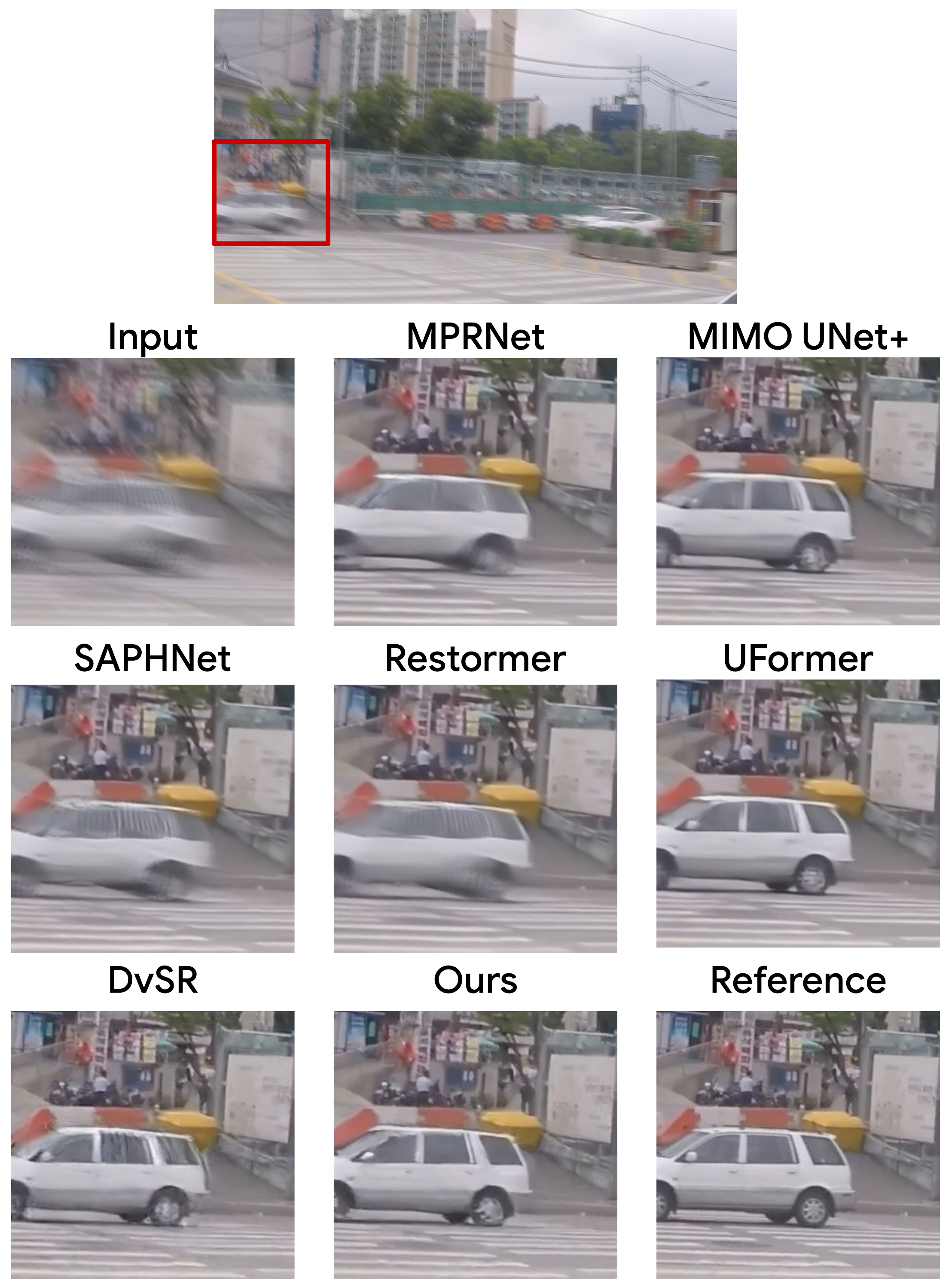}
    \caption{\textbf{GoPro}~\cite{nah2017deep} deblurring examples from MPRNet~\cite{Zamir2021MPRNet}, MIMO UNet+~\cite{cho2021rethinking_mimounet}, SAPHNet~\cite{suin2020spatially}, Restormer~\cite{Zamir2021Restormer}, UFormer~\cite{Wang_2022_CVPR_uformer}, DvSR~\cite{whang2022deblurring} and Ours.}
    \label{fig:gopro1}
\end{figure*}

\begin{figure*}[t]
    \centering
    \includegraphics[width=0.9\linewidth]{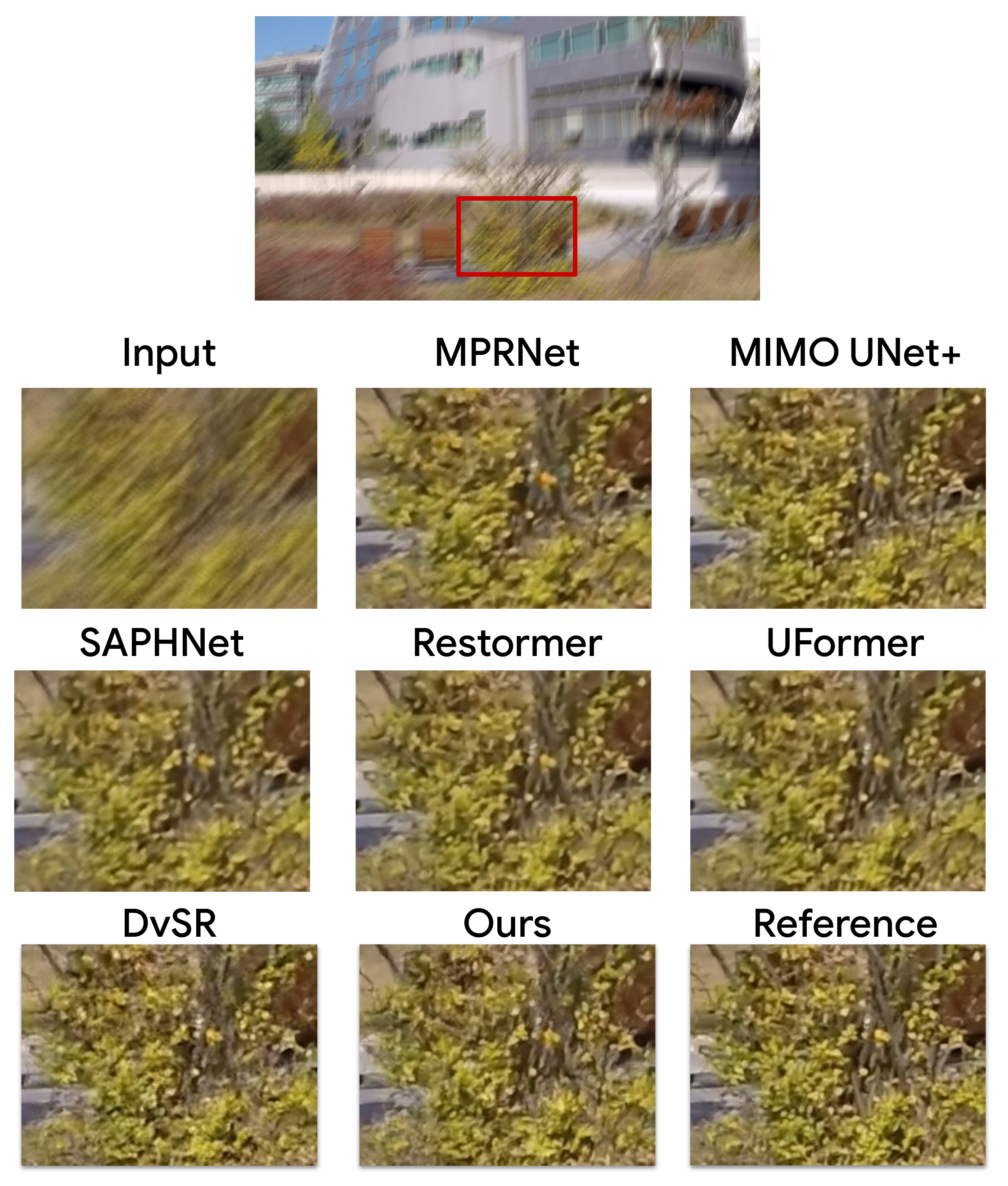}
    \caption{\textbf{GoPro}~\cite{nah2017deep} deblurring examples from MPRNet~\cite{Zamir2021MPRNet}, MIMO UNet+~\cite{cho2021rethinking_mimounet}, SAPHNet~\cite{suin2020spatially}, Restormer~\cite{Zamir2021Restormer}, UFormer~\cite{Wang_2022_CVPR_uformer}, DvSR~\cite{whang2022deblurring} and Ours.}
    \label{fig:gopro2}
\end{figure*}

\begin{figure*}[t]
    \centering
    \includegraphics[width=0.9\linewidth]{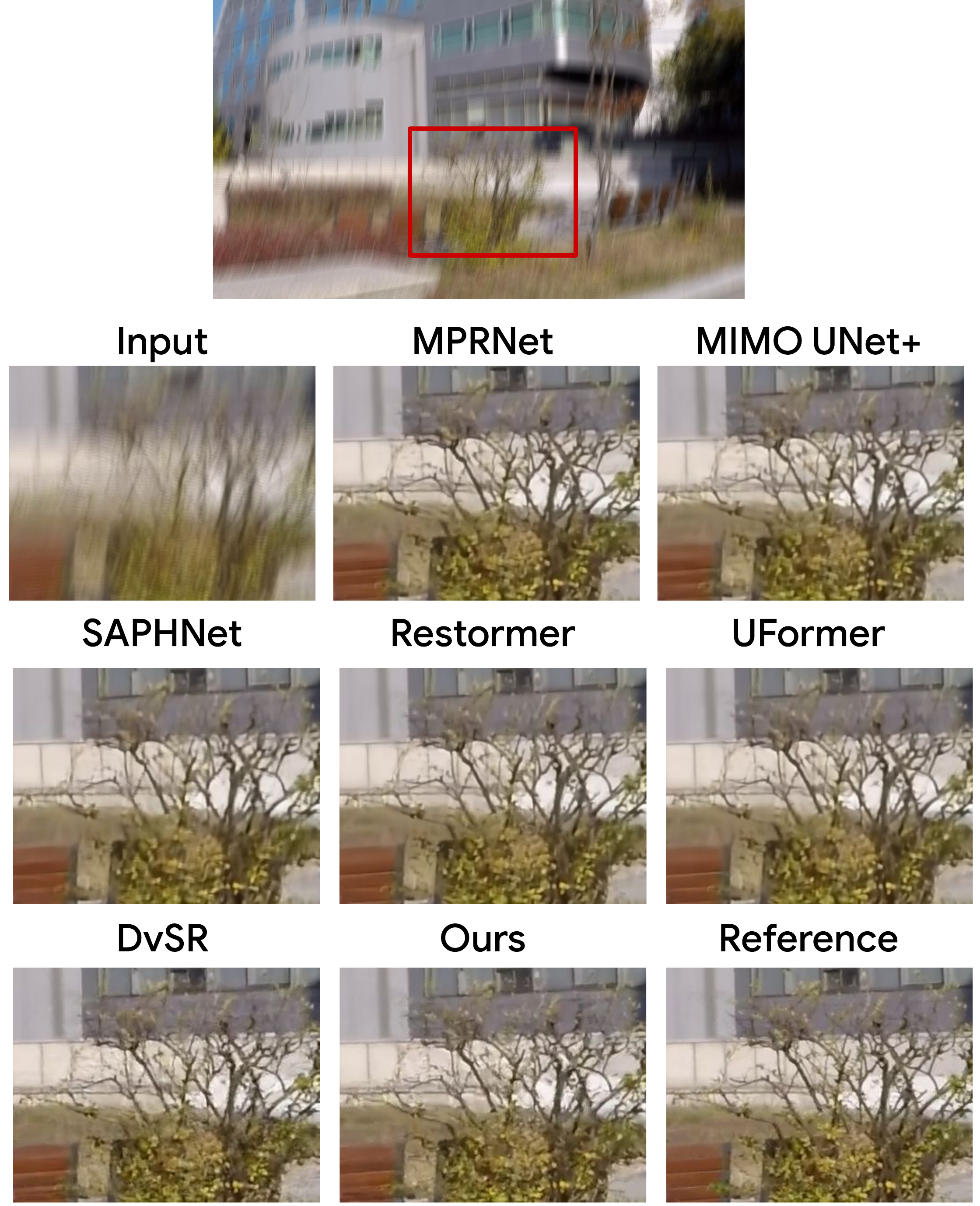}
    \caption{\textbf{GoPro}~\cite{nah2017deep} deblurring examples from MPRNet~\cite{Zamir2021MPRNet}, MIMO UNet+~\cite{cho2021rethinking_mimounet}, SAPHNet~\cite{suin2020spatially}, Restormer~\cite{Zamir2021Restormer}, UFormer~\cite{Wang_2022_CVPR_uformer}, DvSR~\cite{whang2022deblurring} and Ours.}
    \label{fig:gopro3}
\end{figure*}

\begin{figure*}[ht]
    \centering
    \includegraphics[width=0.9\linewidth]{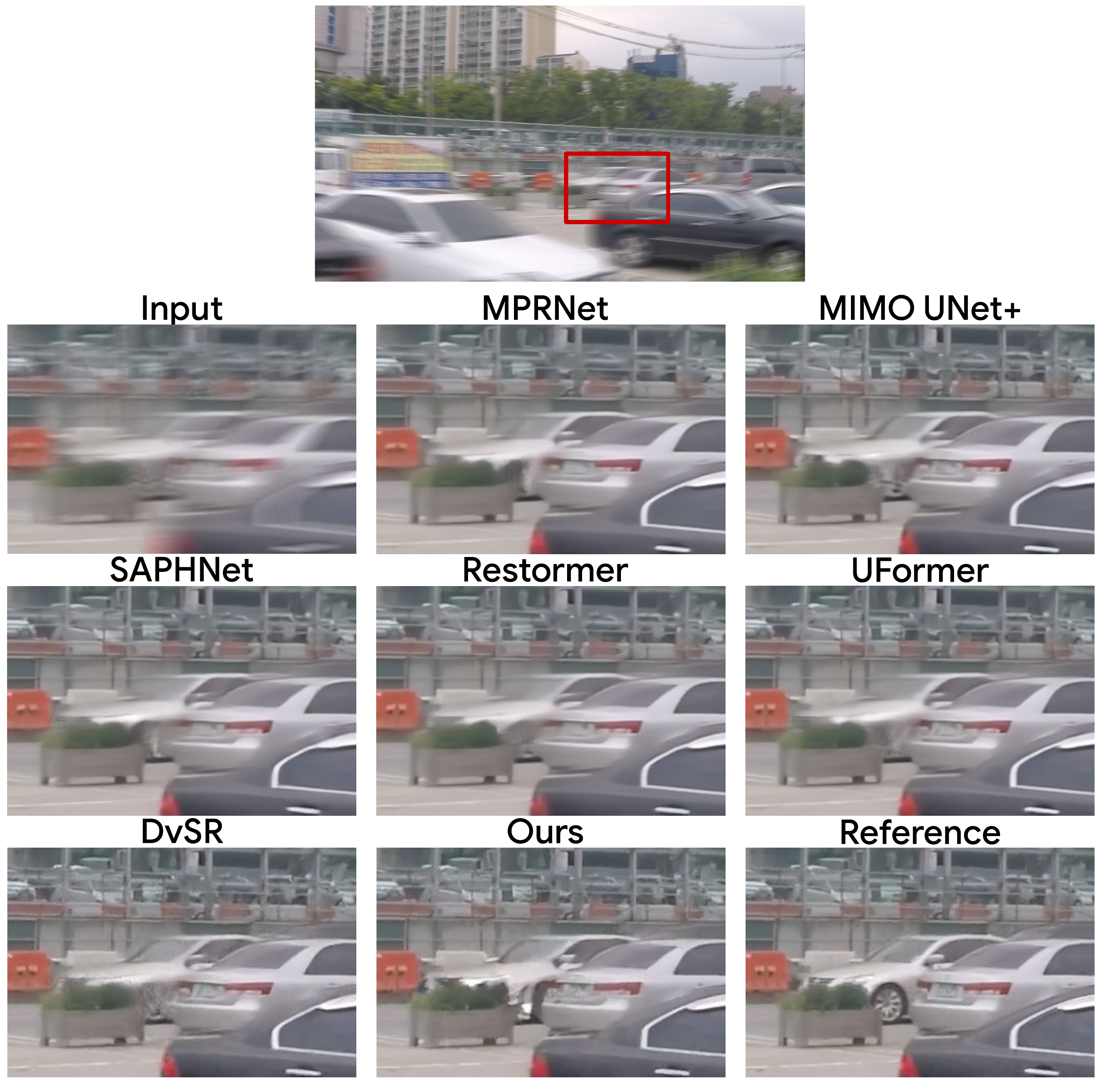}
    \caption{\textbf{GoPro}~\cite{nah2017deep} deblurring examples from MPRNet~\cite{Zamir2021MPRNet}, MIMO UNet+~\cite{cho2021rethinking_mimounet}, SAPHNet~\cite{suin2020spatially}, Restormer~\cite{Zamir2021Restormer}, UFormer~\cite{Wang_2022_CVPR_uformer}, DvSR~\cite{whang2022deblurring} and Ours.}
    \label{fig:gopro4}
\end{figure*}

\begin{figure*}[t]
    \centering
    \includegraphics[width=0.8\linewidth]{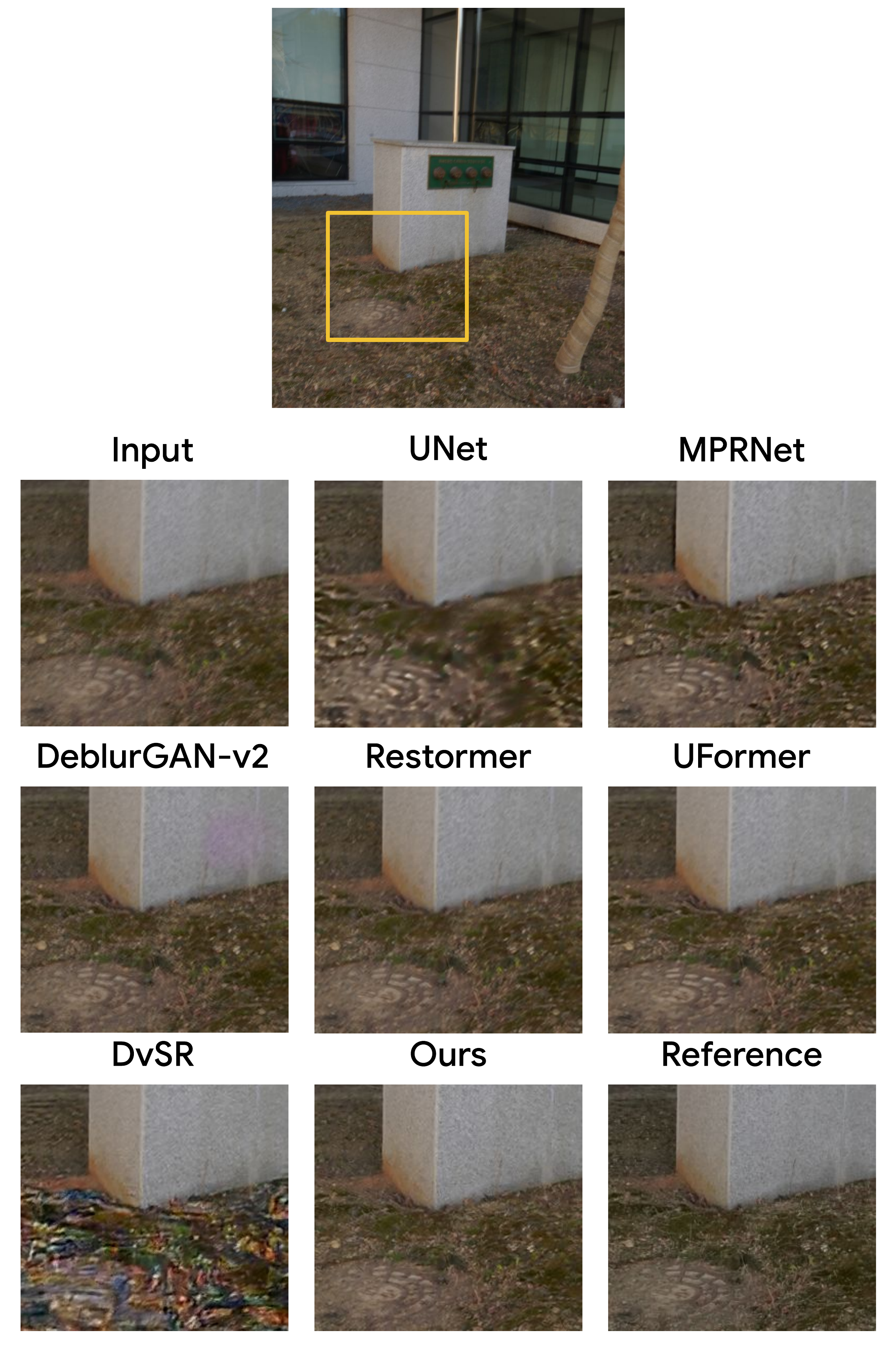}
    \caption{\textbf{Realblur-J}~\cite{rim2020real} deblurring examples from UNet~\cite{ronneberger2015u}, MPRNet~\cite{Zamir2021MPRNet}, DeblurGAN-v2~\cite{Kupyn_2019_ICCV_Deblurganv2}, Restormer~\cite{Zamir2021Restormer}, UFormer~\cite{Wang_2022_CVPR_uformer}, DvSR~\cite{whang2022deblurring} and Ours.    
    }
    \label{fig:real1}
\end{figure*}

\begin{figure*}[t]
    \centering
    \includegraphics[width=0.85\linewidth]{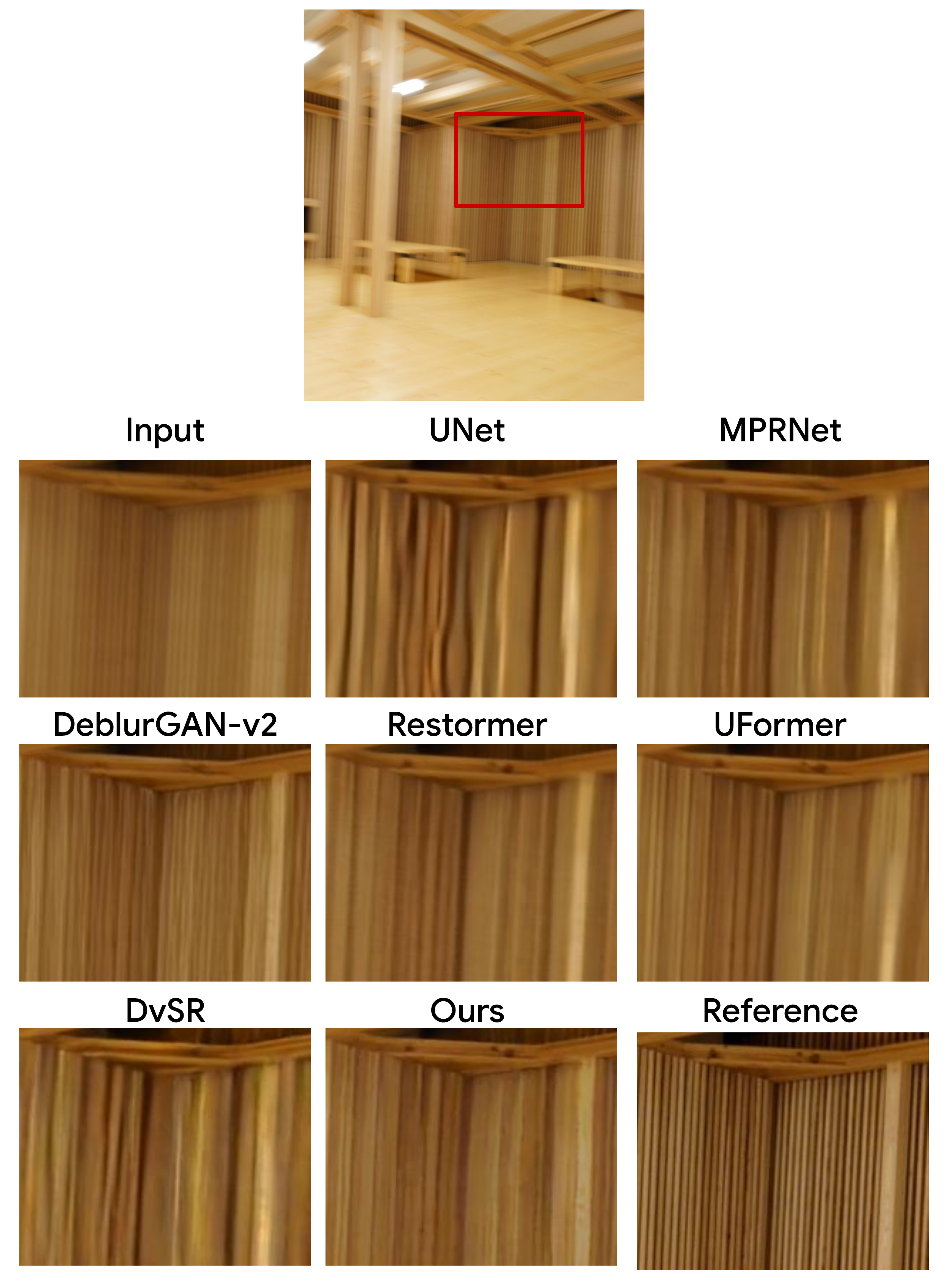}
    \caption{\textbf{Realblur-J}~\cite{rim2020real} deblurring examples from UNet~\cite{ronneberger2015u}, MPRNet~\cite{Zamir2021MPRNet}, DeblurGAN-v2~\cite{Kupyn_2019_ICCV_Deblurganv2}, Restormer~\cite{Zamir2021Restormer}, UFormer~\cite{Wang_2022_CVPR_uformer}, DvSR~\cite{whang2022deblurring} and Ours.    
    }
    \label{fig:real2}
\end{figure*}

\begin{figure*}[t]
    \centering
    \includegraphics[width=0.76\linewidth]{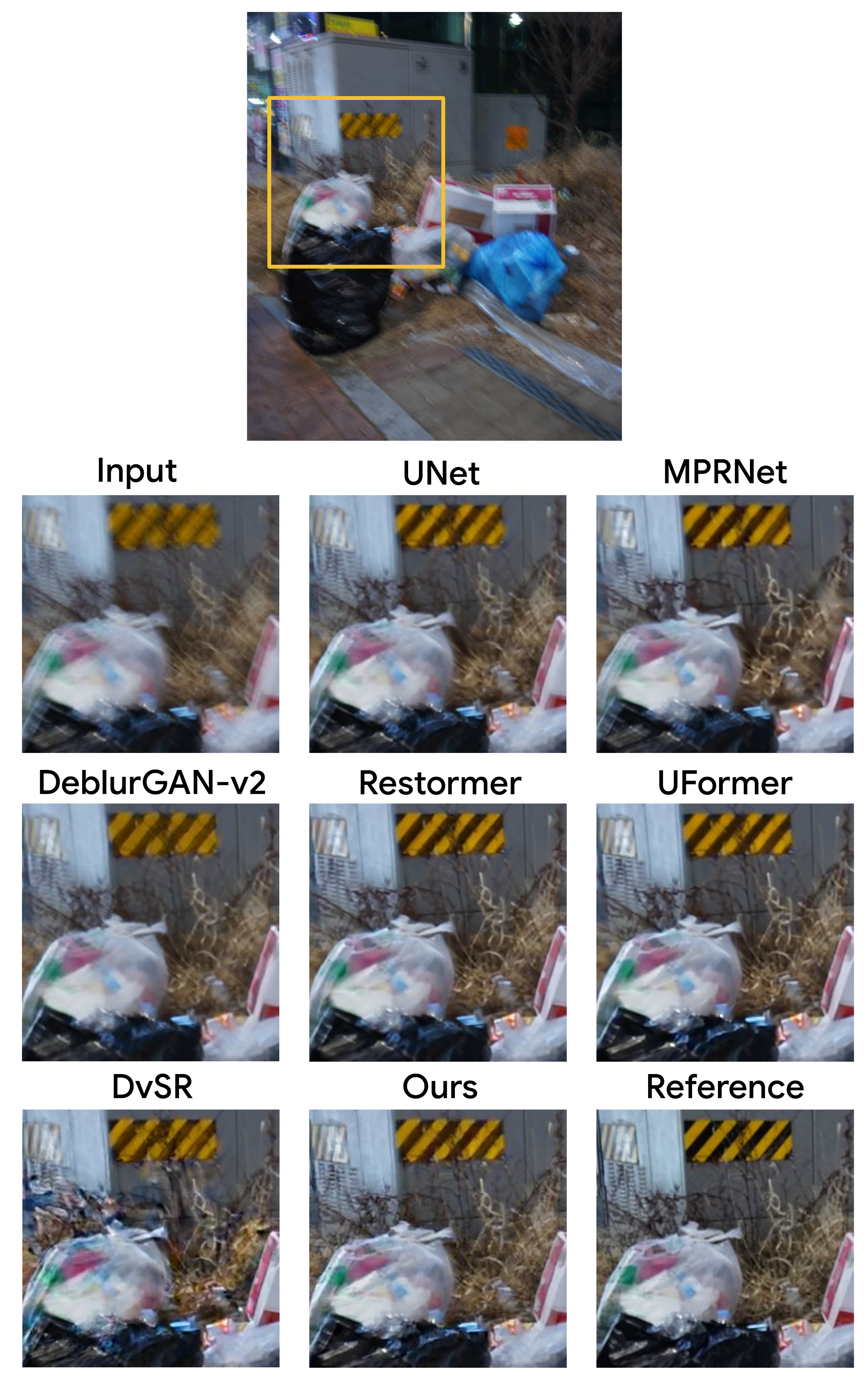}
    \caption{\textbf{Realblur-J}~\cite{rim2020real} deblurring examples from UNet~\cite{ronneberger2015u}, MPRNet~\cite{Zamir2021MPRNet}, DeblurGAN-v2~\cite{Kupyn_2019_ICCV_Deblurganv2}, Restormer~\cite{Zamir2021Restormer}, UFormer~\cite{Wang_2022_CVPR_uformer}, DvSR~\cite{whang2022deblurring} and Ours.    
    }
    \label{fig:real3}
\end{figure*}

\begin{figure*}[t]
    \centering
    \includegraphics[width=0.76\linewidth]{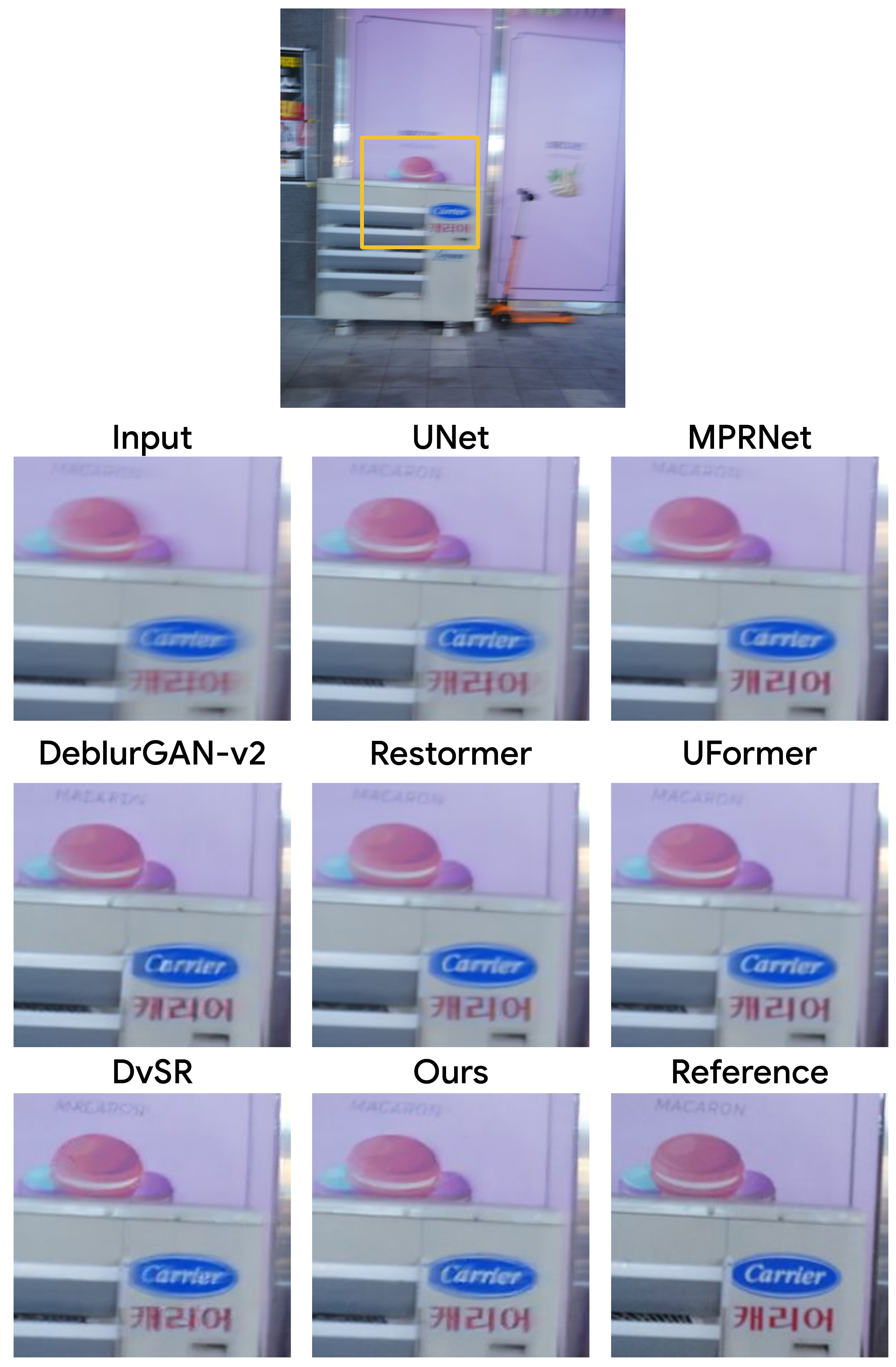}
    \caption{\textbf{Realblur-J}~\cite{rim2020real} deblurring examples from UNet~\cite{ronneberger2015u}, MPRNet~\cite{Zamir2021MPRNet}, DeblurGAN-v2~\cite{Kupyn_2019_ICCV_Deblurganv2}, Restormer~\cite{Zamir2021Restormer}, UFormer~\cite{Wang_2022_CVPR_uformer}, DvSR~\cite{whang2022deblurring} and Ours.    
    }
    \label{fig:real4}
\end{figure*}

\begin{figure*}[!ht]
    \centering
    \includegraphics[width=0.9\linewidth]{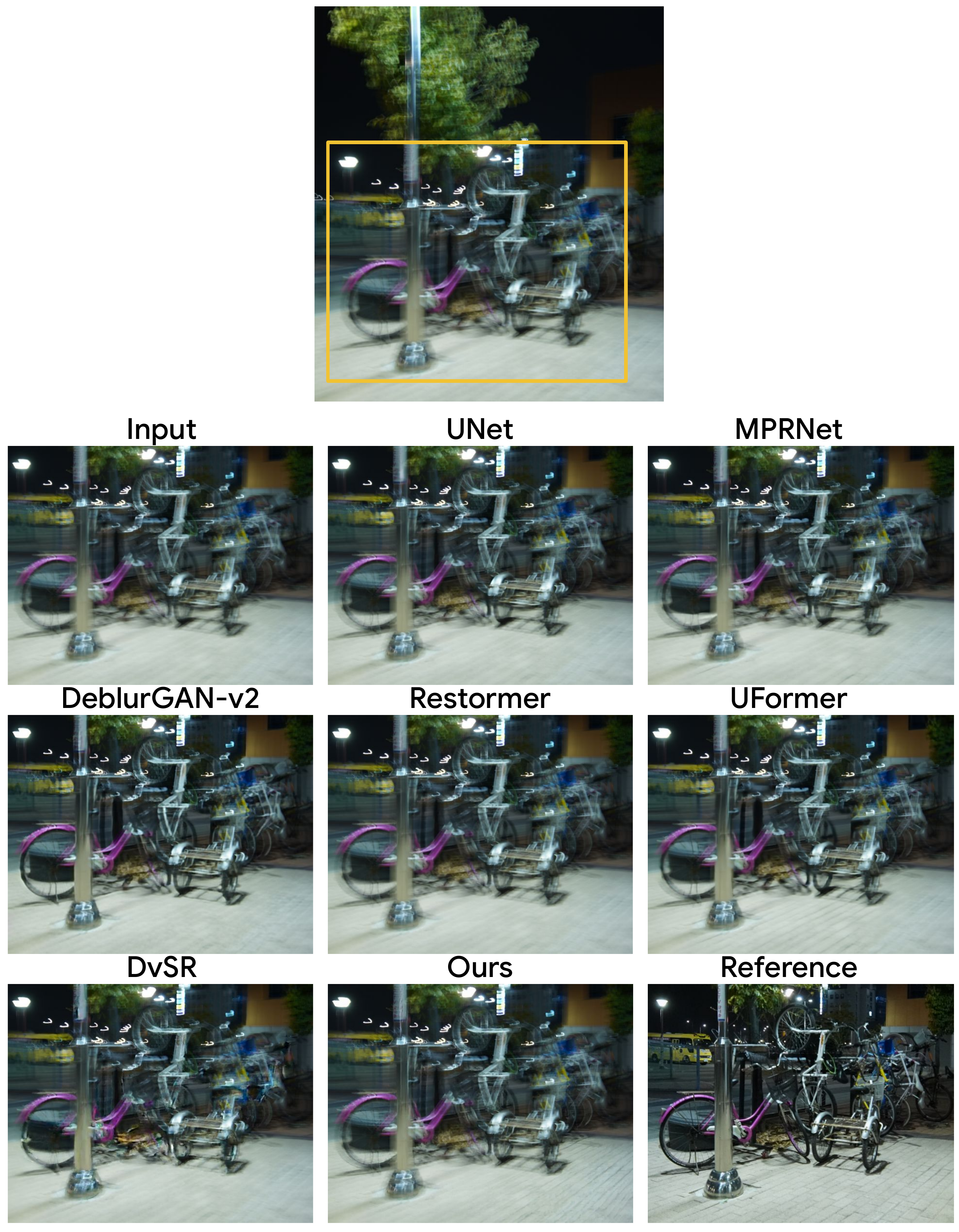}
    \caption{Failure case from \textbf{Realblur-J}~\cite{rim2020real} in low-light scenes.}
    \label{fig:failure1}
\end{figure*}

\begin{figure*}[!ht]
    \centering
    \includegraphics[width=0.9\linewidth]{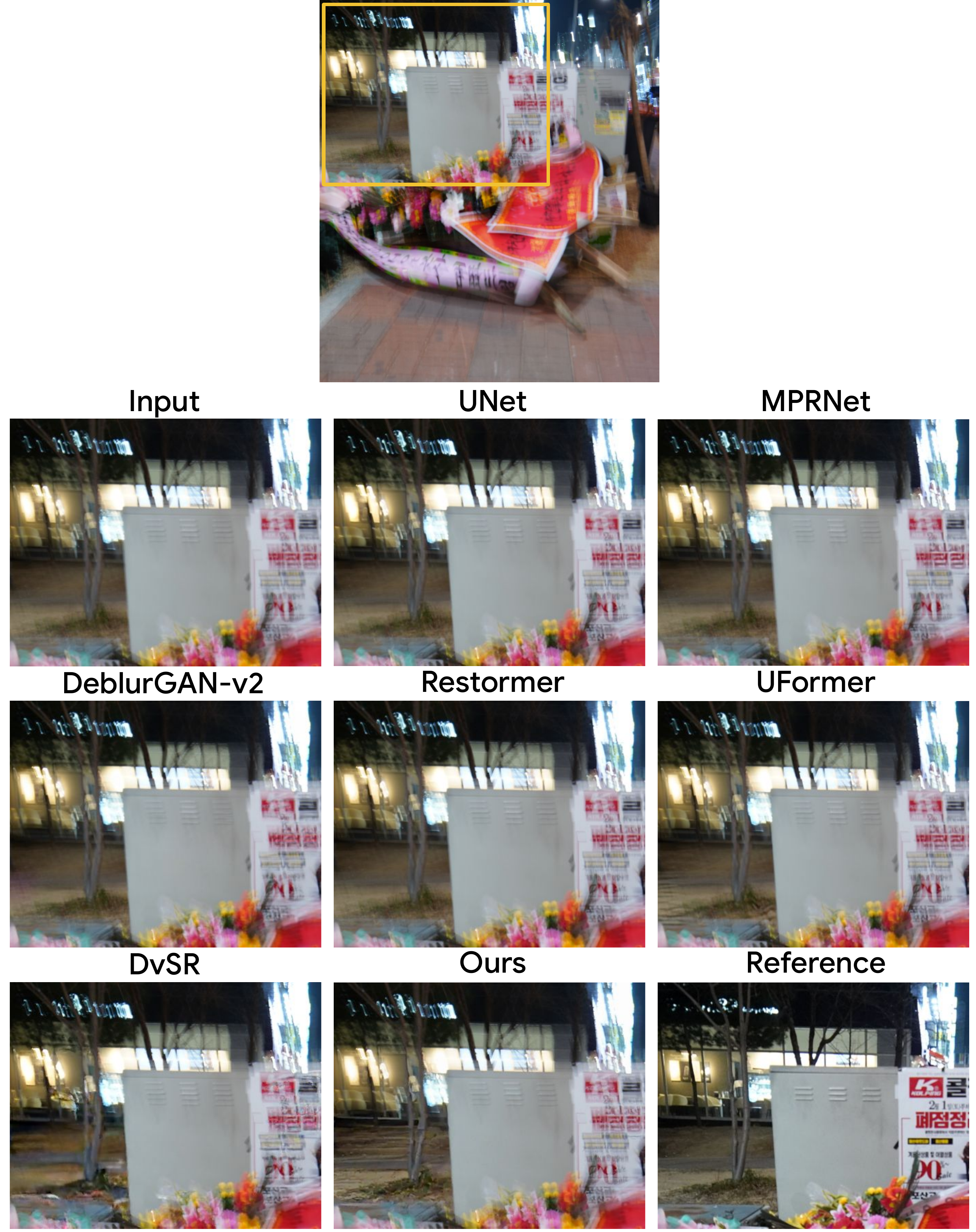}
    \caption{Failure case from \textbf{Realblur-J}~\cite{rim2020real} with strong light streaks.}
    \label{fig:failure2}
\end{figure*}

\begin{figure*}[!ht]
    \centering
    \includegraphics[width=0.95\linewidth]{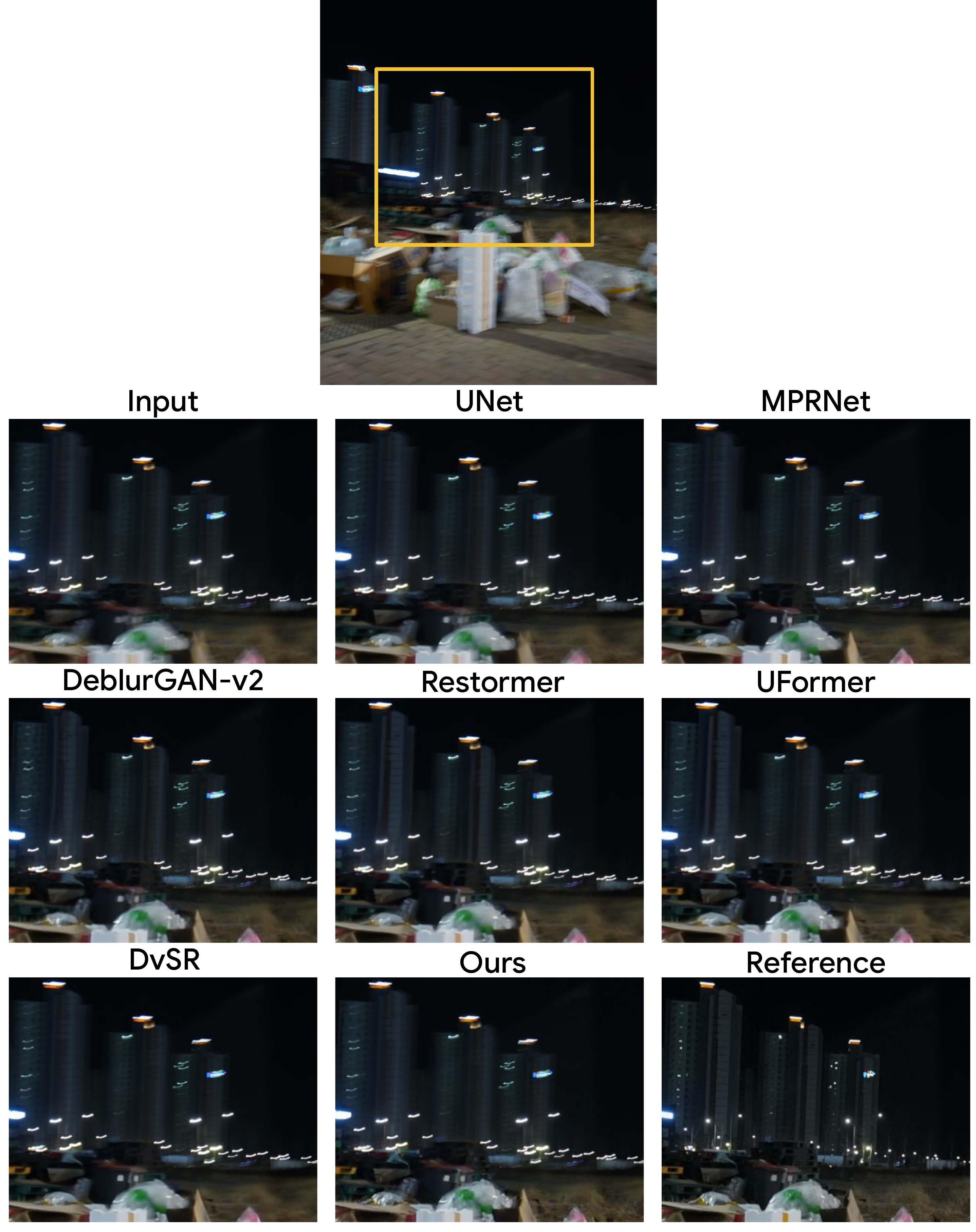}
    \caption{Failure case from \textbf{Realblur-J}~\cite{rim2020real} in night scenes.}
    \label{fig:failure3}

\end{figure*}

\begin{figure*}[!ht]
    \centering
    \includegraphics[width=0.95\linewidth]{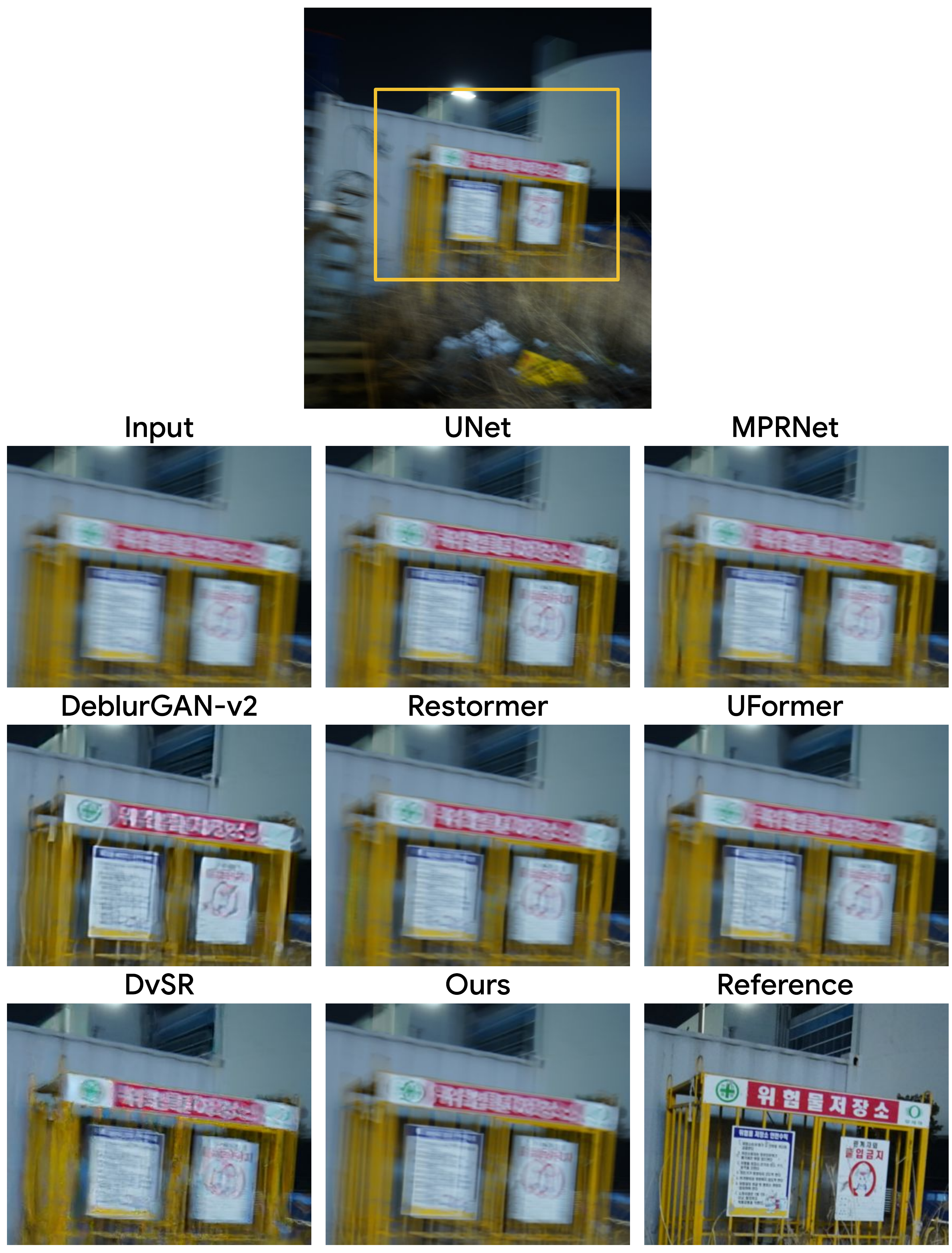}
    \caption{Failure case from \textbf{Realblur-J}~\cite{rim2020real} in low-light condition.}
    \label{fig:failure4}

\end{figure*}

The architectural details for the diffusion network and the guidance network are illustrated in Fig.~\ref{fig:detailed_arch}.

\clearpage
\subsection{Inference}
As we use continuous noise level sampling during training, it enables the use of different noise schedulers during the inference to potentially obtain samples with different distortion-perception trade-off. We therefore perform a grid search over a set of different diffusion steps $T$, as well as the upper bound of the noise variance $1-\alpha_T$. For efficiency, we also exclude certain combinations that do not produce reasonable sampling (i.e., sampling results are pure noise or blank image), and the final combinations are indicated in Table~\ref{tab:sampler}.
\begin{table}[!h]
    \centering
    \caption{The sampling parameters for inference.}
    \begin{tabular}{@{}cc|c*{5}{c}}
    \multicolumn{1}{c}{}  &   &\multicolumn{6}{c}{Maximum noise variance $1-\alpha_T$}\\
    \multicolumn{1}{c}{} & &0.01    & 0.02    & 0.05    & 0.1    & 0.2    & 0.5 \\ \hline
    \multirow{7}*{\rotatebox{90}{Steps ($T$)}}
   & 20 &     &      &    &     &     & $\checkmark$    \\ 
   & 30 &     &      &     &     &     & $\checkmark$    \\  
   & 50 &     &      &     &     & $\checkmark$ & $\checkmark$       \\ 
   & 100 &     &     &     & $\checkmark$ & $\checkmark$ &$\checkmark$      \\ 
   & 200 &     &     & $\checkmark$   &$\checkmark$ &$\checkmark$  &$\checkmark$ \\ 
   & 500 &    & $\checkmark$    &$\checkmark$    & $\checkmark$    &$\checkmark$    &    \\  
   & 1000 & $\checkmark$    &$\checkmark$ &$\checkmark$    &$\checkmark$  &   &        \\  \bottomrule
    \end{tabular}
    \label{tab:sampler}
\end{table}

\vspace{2em}
\subsection{Computational cost}
In Table~\ref{table:flops}, we report floating point operations per second (FLOPs) under different model configurations, calculated based on an input image of $720 \times 1280 \times 3$. For diffusion networks (c)-(d), the FLOPS are calculated based on a single diffusion step. 
While optimizing sampling speed is out-of-scope of this work, we believe recent advance in speeding up DPM sampling~\cite{ddim,chung2022come,xiao2022DDGAN,liu2021pseudo,lu2022dpm,ma2022accelerating,dockhorn2022genie,meng2022distillation,li2022efficient} could be further incorporated into our framework.
\begin{table*}[!h]
\small
\centering
\caption{FLOPs under different model configurations, calculated based on a full-size input image of $720 \times 1280 \times 3$. For diffusion networks (c)-(d), the FLOPs are calculated based on a single diffusion step.
}
\label{table:flops}
\begin{tabular}{lcccc}
\toprule
 & Guidance network  &  Diffusion network & \# Params & FLOPs  \\ \midrule
(a) icDPM-S             & -                             & ch=32                          & 6M                        &    1200B         \\
(b) icDPM-L             & -                             & ch=64                          & 27M                       &    4800B      \\
\midrule
(c) icDPM-S w/ Guide-S     & ch=32                         & ch=32                          & 10M                       & 2500B            \\
(d) icDPM-L w/ Guide-S    & ch=32                         & ch=64                          & 30M                       & 6100B          \\
(e) icDPM-L w/ Guide-L             & ch=64                         & ch=64                          & 52M              & 10000B         \\ \bottomrule
\end{tabular}
\end{table*}

\vspace{2em}
\subsection{Benchmark results}
We performed a consistent computation over all benchmarks for fair comparisons. To acquire the benchmark results, we use the author provided results whenever possible. On the cross-domain set up of Realblur-J with GoPro trained only models, we use author provided results of DvSR~\cite{whang2022deblurring}, UFormer~\cite{Wang_2022_CVPR_uformer}, Restormer~\cite{Zamir2021Restormer}. For DeblurGAN-v2~\cite{Kupyn_2019_ICCV_Deblurganv2} and MPRNet~\cite{Zamir2021MPRNet}, we use official code repository along with the provided GoPro checkpoints for inference. 
On REDS~\cite{Nah_2019_CVPR_Workshops_REDS}, all results are obtained by running their official models with the GoPro checkpoints.

\clearpage





\end{document}